%% file: paper.tex
\documentclass{article}

\usepackage[preprint]{neurips_2020}

\usepackage[utf8]{inputenc} % allow utf-8 input
\usepackage[T1]{fontenc}    % use 8-bit T1 fonts
\usepackage{hyperref}       % hyperlinks
\usepackage{url}            % simple URL typesetting
\usepackage{booktabs}       % professional-quality tables
\usepackage{amsfonts}       % blackboard math symbols
\usepackage{nicefrac}       % compact symbols for 1/2, etc.
\usepackage{microtype}      % microtypography

\usepackage{amsmath}
\usepackage{amsfonts}
\usepackage{amsthm}
\usepackage{amssymb}
\usepackage{subcaption}
\usepackage{natbib}
\usepackage{wrapfig}
\usepackage{xfrac}
\usepackage{nicefrac}

\usepackage{tikz}
\usetikzlibrary{positioning}
\usetikzlibrary{tikzmark} % arrows in tex
\usetikzlibrary{arrows}   % arrows in tex
\usetikzlibrary{calc}     % (node)+(3cm,2cm)
\tikzstyle{every picture}+=[remember picture]
\usetikzlibrary{shapes}

\definecolor{my_red}{RGB}{196,78,82}
\definecolor{my_green}{RGB}{85, 168, 104}
\definecolor{my_blue}{RGB}{76,114,176}
\definecolor{my_purple}{RGB}{129,114,178}
\definecolor{my_yellow}{RGB}{204,185,116}

\newtheorem{definition}{Definition}

\title{Geometrically Enriched Latent Spaces}

\author{
	Georgios Arvanitidis\textsuperscript{$\dagger$}
	\quad
	S{\o}ren Hauberg\textsuperscript{$\ddagger$}
	\quad
	Bernhard Sch{\"o}lkopf\textsuperscript{$\dagger$}\\
	\textsuperscript{$\dagger$}MPI for Intelligent Systems, T{\"u}bingen, Germany\\
	\textsuperscript{$\ddagger$}Technical University of Denmark, Lyngby, Denmark
}

\usepackage{smartdiagram}
\hypersetup{
	colorlinks,
	linkcolor={red!50!black},
	citecolor={blue!50!black},
	urlcolor={blue!80!black}
}

\input{notation}

\begin{document}

\maketitle

\begin{abstract}
%	A common assumption in generative models is that the generator immerses the latent space into an Euclidean ambient space. Instead, we consider the ambient space to be a Riemannian manifold, which provides additional information regarding the problem of interest. Thus, in the latent space the shortest paths respect both the geometry of the generated surface and the geometry of the ambient space. In addition, we propose a way to extrapolate meaningfully in deterministic generators. In the experiments we show the effect of our approach in representation learning and interpretable interpolations testing both stochastic and deterministic generators.
	
A common assumption in generative models is that the generator immerses the latent space into a Euclidean ambient space. Instead, we consider the ambient space to be a Riemannian manifold, which allows for encoding domain knowledge through the associated Riemannian metric. Shortest paths can then be defined accordingly in the latent space to both follow the learned manifold and respect the ambient geometry. Through careful design of the ambient metric we can ensure that shortest paths are well-behaved even for deterministic generators that otherwise would exhibit a misleading bias. Experimentally we show that our approach improves interpretability of learned representations both using stochastic and deterministic generators.
	
%	In the latent space of a generative model we can capture the geometry of the embedded high dimensional data manifold, by considering it as a Riemannian manifold. The common assumption is that the generated manifold is immersed into an Euclidean ambient space. Instead, in this paper we show how we can consider the ambient space to be another Riemannian manifold, which provides additional information to the latent geometry. In principle, we show how information about the problem can be formulated into a geometrical quantity, such that to move optimally in the latent space while we respect the geometry of the high dimensional manifold, and also, to follow the additional information of the problem. We demonstrate how the proposed can be useful in several cases using stochastic and deterministic generators. Also, this can be easily adapted to any problem of interest.
\end{abstract}

%%%%%%%%% MAIN PAPER %%%%%%%%%
\input{intro}

\input{geometry}
\input{methods}
\input{experiments}

\input{conclusion}

\section*{Broader Impact}

We have proposed a framework that allows for solving the identifiability
problem associated with latent variable models, while retaining flexibility with regards to the metric behavior of the latent space.
This is valuable wherever a \emph{faithful representation} is of use,
such as to ensure a \emph{fair} and \emph{interpretable} model. The approach also carries potential value for \emph{causal inference} where the identifiabiity issue is a paramount concern.

The model does carry a risk of inappropriate usage, as an end-user (most likely a data scientist) can easily manipulate empirical findings by changing the ambient metric. Misleading results can, thus, be presented by a manipulation of an ambient metric, which may not be transparent to recipients of the data analysis. Conceptually, this is the ``same old'' issue that occurs when empirical findings are overly sensitive to data pre-processing.

%In general, the framework that we proposed has a broad usability and the behavior is easy to interpret. In particular, users, researchers or domain experts can easily construct Riemannian metrics in the ambient space that encode knowledge about the problem of interest. We showed several examples where we controlled the shortest paths to prefer regions with selected data, to avoid some other regions with high cost or even to artificially bring closer disconnected data manifolds. Therefore, parts of the ambient space with some particular attributes can be considered differently in downstream tasks, which are solved in the geometrically enriched and more informed latent space. We believe that the proposed approach can be useful in a wide range of problems as in fairness, medical data, safety and control applications, etc. 
%
%Of course, it is worth mentioning the direct disadvantage of our method, which is the exact same argumentation of the previous paragraph but from the negative side this time. For instance, easily someone is able to bias the shortest paths to avoid or prefer specific regions in the data space. However, we believe that the model itself does not have a direct negative impact, but rather the purpose for which the ambient metrics will be constructed.

\section*{Acknowledgments}

SH was supported by a research grant (15334) from
VILLUM FONDEN. This project has received funding from the European Research Council (ERC) under the European Union’s Horizon 2020 research and innovation programme (grant agreement n\textsuperscript{o} 757360).

\small

\bibliographystyle{abbrvnat}
\bibliography{bib-paper.bib}

\newpage
\appendix
\input{appendix}

\end{document}

%% file: notation.tex
\DeclareMathOperator*{\argmin}{\arg\!\min}
\DeclareMathOperator*{\argmax}{\arg\!\max}
\newcommand{\length}[1]{\text{\emph{length}}[#1]} % FOR BOLD

\newcommand*{\M}{\mathcal{M}} % The manifold
\newcommand*{\U}{\mathcal{U}} % The manifold
\newcommand*{\X}{\mathcal{X}} % The ambient space
\newcommand*{\Z}{\mathcal{Z}} % The latent space
\newcommand*{\Id}{\mathbb{I}} % The identity matrix
\newcommand*{\R}{\mathbb{R}} % The real space

\newcommand{\bs}[1]{\boldsymbol{#1}} % FOR BOLD SYMBOLS
\renewcommand{\b}[1]{\mathbf{#1}} % FOR BOLD MATH
\newcommand{\vectorize}[1]{\text{vec}\!\left[ #1\right]} % THE VECTORIZE OPERATOR
\newcommand{\parder}[2]{\frac{\partial#1}{\partial#2}} % THE PARTIAL DERIVATIVE \parder{f}{x}
\newcommand{\inner}[2]{\langle #1, #2 \rangle} % The inner product \inner{v}{x}
\newcommand{\abs}[1]{\left\vert #1\right\vert} %  Absolute value
\newcommand{\Logmap}[2]{\text{Log}_{#1}(#2)}
\newcommand{\Expmap}[2]{\text{Exp}_{#1}(#2)}
\newcommand{\tangent}[2]{\mathcal{T}_{\boldsymbol{#1}}{#2}} % the tangent space symbol
\newcommand{\norm}[1]{\left\lVert#1\right\rVert} % THE NORM

\newcommand*{\T}{\intercal} % The Gaussian process
\newcommand*{\E}{\mathbb{E}}  % The expectation
\newcommand*{\N}{\mathcal{N}} % the Normal distribution symbol

%% file: intro.tex
%!TEX root = ./paper.tex

\section{Introduction}

%\begin{itemize}
%	\item Learn a generative model.
%	\item Representations of data manifold in a latent space.
%	\item It has been shown how we can capture properly the geometry.
%	\item This needs stochastic generators which is not always the case.
%	\item Also, learning representations for GANs is not easy.
%	\item We have additional problem information that we could use in the latent space.
%\end{itemize}

%\begin{wrapfigure}{r}{0.3\textwidth}
%	\vspace{-20pt}
%	\begin{center}
%		\begin{subfigure}[b]{0.3\textwidth}
%			\includegraphics[width=\textwidth]{imgs/intro/teaser_2_v3.png}
%%		\end{subfigure}
%%		\begin{subfigure}[b]{0.3\textwidth}
%			\includegraphics[width=\textwidth]{imgs/intro/teaser_1_v3.png}
%		\end{subfigure}
%		\caption[none]{Comparing ours ({\protect \begin{tikzpicture}[baseline={([yshift=-3pt]current bounding box.center)}]
%				{\draw [color={rgb,255:red,220; green,79; blue,53}, line width=1.5, yshift=0cm] (0,0) -- (.3,0);}
%				\end{tikzpicture}}) and standard ({\protect \begin{tikzpicture}[baseline={([yshift=-3pt]current bounding box.center)}]
%				{\draw [color={rgb,255:red,82; green,173; blue,58}, line width=1.5, yshift=0cm] (0,0) -- (.3,0);}
%				\end{tikzpicture}}) shortest paths, on a GAN (\emph{top}) and a VAE (\emph{bottom}) case.}
%		\label{intro:teaser_fig}
%	\end{center}
%	\vspace{-20pt}
%\end{wrapfigure}%
%

\begin{wrapfigure}{r}{0.40\textwidth}
	\vspace{-10pt}
	\begin{center}
		\begin{subfigure}[b]{0.40\textwidth}
			\includegraphics[width=\textwidth]{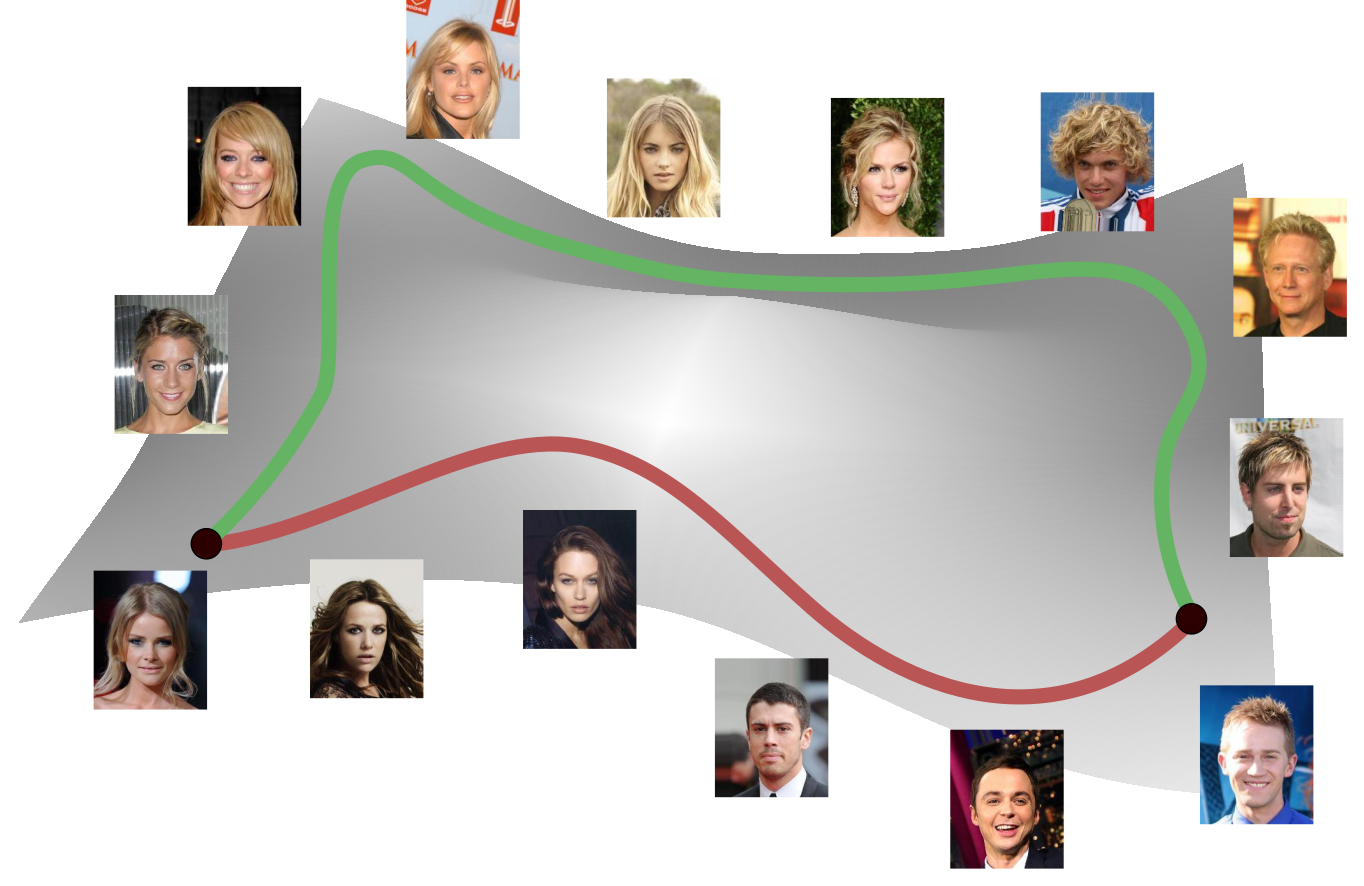}
		\end{subfigure}
	\caption[none]{The proposed shortest path 
		({\protect 
			\begin{tikzpicture}[baseline={([yshift=-3pt]current bounding box.center)}] 	{\draw [color={rgb,255:red,82; green,173; blue,58}, line width=1.5, yshift=0cm] (0,0) -- (.3,0);}
			\end{tikzpicture}}) favors the \emph{blond} class, while the standard shortest path 
		({\protect
			\begin{tikzpicture}[baseline={([yshift=-3pt]current bounding box.center)}]
				{\draw [color={rgb,255:red,220; green,79; blue,53}, line width=1.5, yshift=0cm] (0,0) -- (.3,0);}
			\end{tikzpicture}}) merely minimizes the distance on the manifold.}
		\label{sec:intro:fig:teaser}
	\end{center}
\vspace{-20pt}
\end{wrapfigure}\emph{Unsupervised representation learning} has made tremendous gains with generative
models such as \emph{variational autoencoders (VAEs)} \citep{kingma:iclr:2014, rezende:icml:2014}
and \emph{generative adversarial networks (GANs)} \citep{goodfellow:neurips:2014}.
These, and similar, models provide a flexible and efficient parametrization
of the density of observations in an ambient space $\X$ through a typically lower dimensional latent space $\Z$.

While the latent space $\Z$ constitutes a compressed representation of the data,
it is by no means unique. Like most other latent variable models, these generative
models are subject to \emph{identifiability problems}, such that different representations
can give rise to identical densities \citep{Bishop:2006:PRM}. This implies that
straight lines in $\Z$ are not shortest paths in any meaningful sense, and therefore
do not constitute natural interpolants. To overcome this issue, it has been 
proposed to endow the latent space with a \emph{Riemannian metric} such that
curve lengths are measured in the ambient observation space $\X$
\citep{tosi:uai:2014, arvanitidis:iclr:2018}. This approach immediately solves the identifiability problem. In other words, this ensures that any smooth invertible transformation of $\Z$ does not change the distance between a pair of points, as long as the ambient path in $\X$ remains the same.

While distances in $\X$ are well-defined and give rise to an identfiable latent
representation, they need not be particularly useful.
We take inspiration from \emph{metric learning} \citep{weinberger:2006:neurips, arvanitidis:nips:2016}
and propose to equip the ambient observation space $\X$ with a Riemannian metric
and measure curve lengths in latent space accordingly. With this approach it is
straight-forward to steer shortest paths in latent space to avoid low-density
regions, but also to incorporate higher level information. For instance,
Fig.~\ref{sec:intro:fig:teaser} shows a shortest path interpolant under an ambient metric
that favors images of \emph{blond} people. Hence, we get both identifiable
and useful latent representations.

%% file: geometry.tex
\section{A compact introduction to applied Riemannian geometry}
\label{sec:geometry}

We are interested in Riemannian manifolds \citep{docarmo:1992}, which constitute well-defined metric spaces, where the inner product is defined only locally and changes smoothly throughout space. In a nutshell, these are smooth spaces where we can compute shortest paths, which prefer to cross regions where the magnitude of the inner product is small. In this work, we show how to use such structures in machine learning, where is commonly assumed that data lie near a low dimensional manifold.

\begin{definition}
A Riemannian manifold is a smooth manifold $\mathcal{M}$, equipped with a positive definite Riemannian metric $\b{M}(\b{x})~\forall~\b{x}\in\M$, which changes smoothly and defines a local inner product on the tangent space $\tangent{\b{x}}{\M}$ at each point $\b{x}\in\M$ as  $\inner{\b{v}}{\b{u}}_\b{x} = \inner{\b{v}}{\b{M(\b{x})}\b{u}}$ with $\b{v},\b{u}\in\tangent{\b{x}}{\M}$. 
\end{definition}

\begin{wrapfigure}{r}{0.25\textwidth}
	\vspace{-20pt}
	\begin{center}
		\begin{subfigure}[b]{0.25\textwidth}
			\begin{tikzpicture}
				\node[anchor=south west,inner sep=0] (image) at (0,0) {\includegraphics[width=\textwidth]{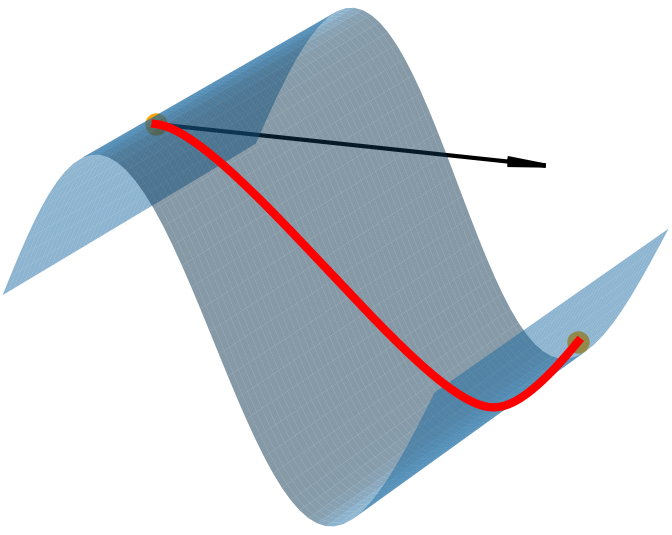}};
				\begin{scope}[x={(image.south east)},y={(image.north west)}]
				\node at (0.7, 0.8) {\small $\b{v}$} ;
				\node at (0.4, 0.4) {\small $\gamma(t)$} ;
				\node at (0.2, 0.9) {\small $\b{x}$} ;
				\node at (0.9, 0.25) {\small $\b{y}$} ;
					
%				{\draw [color={rgb,255:red,241; green,196; blue,67}, line width=1.5, yshift=0cm] (-0.05, 0.11) -- (0, 0.11);}
%							\draw[help lines,xstep=.1,ystep=.1] (0,0) grid (1,1);
%							\foreach \x in {0,1,...,9} { \node [anchor=north] at (\x/10,0) {0.\x}; }
%							\foreach \y in {0,1,...,9} { \node [anchor=east] at (0,\y/10) {0.\y}; }
				\end{scope}
			\end{tikzpicture}
		\end{subfigure}
		~
		\begin{subfigure}[b]{0.25\textwidth}
			\begin{tikzpicture}
				\node[anchor=south west,inner sep=0] (image) at (0,0) {\includegraphics[width=\textwidth]{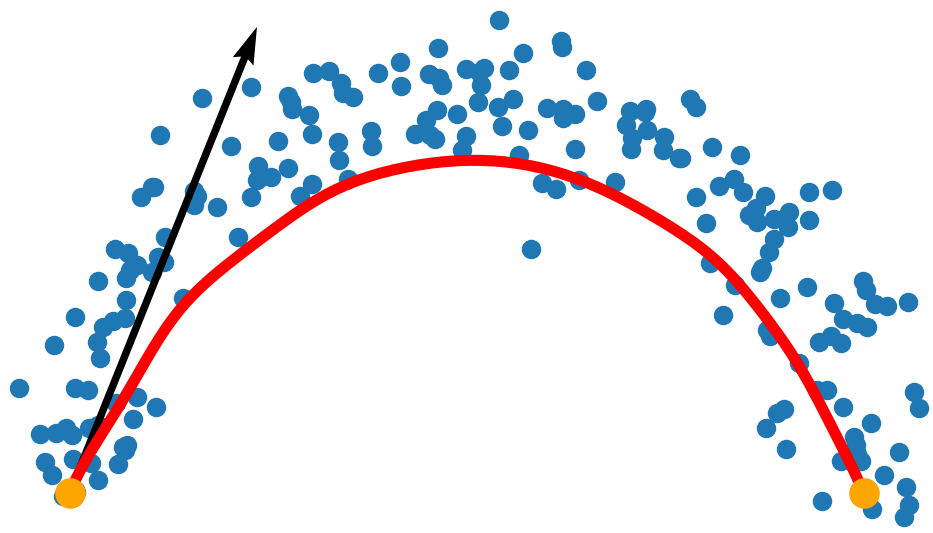}};
				\begin{scope}[x={(image.south east)},y={(image.north west)}]
				\node at (0.2, 1) {\small $\b{v}$} ;
				\node at (0.4, 0.5) {\small $\gamma(t)$} ;
				\node at (0.2, 0.1) {\small $\b{x}$} ;
				\node at (0.8, 0.1) {\small $\b{y}$} ;
				
%				{\draw [color={rgb,255:red,241; green,196; blue,67}, line width=1.5, yshift=0cm] (-0.05, 0.11) -- (0, 0.11);}
%				\draw[help lines,xstep=.1,ystep=.1] (0,0) grid (1,1);
%				\foreach \x in {0,1,...,9} { \node [anchor=north] at (\x/10,0) {0.\x}; }
%				\foreach \y in {0,1,...,9} { \node [anchor=east] at (0,\y/10) {0.\y}; }
				\end{scope}
			\end{tikzpicture}

		\end{subfigure}
			\caption[none]{Examples of tangent vector ({\protect \begin{tikzpicture}[baseline={([yshift=-.5ex]current bounding box.center)}]
				{\draw [->,  centered, black, line width=1.5] (0,0) -- (.4,0);}
				\end{tikzpicture}}) and shortest path ({\protect \begin{tikzpicture}[baseline={([yshift=-3pt]current bounding box.center)}]
				{\draw [red, anchor = west, line width=1.5, yshift=0cm] (0,0) -- (.4,0);}
				%				{\draw [white, line width=1.5, yshift=0cm] (0,0) -- (.4,0);}
				\end{tikzpicture}}) on embedded $\M\subset\X$ (\emph{top}), ambient $\X$ (\emph{bottom}).}
		\label{intro:fig:manifold_examples}
	\end{center}
	\vspace{-30pt}
\end{wrapfigure}A smooth manifold is a topological space, which locally is homeomorphic to a Euclidean space. An intuitive way to think of a $d$-dimensional smooth manifold is as an embedded non-intersecting surface $\M$ in an ambient space $\X$ for example $\R^D$ with $D>d$ (see Fig.~\ref{intro:fig:manifold_examples} top). In this case, the tangent space $\tangent{\b{x}}{\M}$ is a $d$-dimensional vector space tangential to $\M$ at the point $\b{x}\in\M$. Hence, $\b{v}\in\tangent{\b{x}}{\M}$ is a vector $\b{v}\in\R^D$ and actually the Riemannian metric is $\b{M}_\X:\M\rightarrow\R^{D\times D}_{\succ 0}$. The simplest approach is to assume that $\X$ is equipped with the Euclidean metric $\b{M}_\X(\b{x}) = \Id_D$ and its restriction is utilized as the Riemannian metric on $\tangent{\b{x}}{\M}$. Since the choice of $\b{M}_\X(\cdot)$ has a direct impact on $\M$, we can utilize other metrics designed to encode high-level information (see Sec.~\ref{section:data_learned_riemannian_manifolds}).

Another view is to consider as smooth manifold the whole ambient space $\X=\R^D$. Hence, the $\tangent{\b{x}}{\X}=\R^D$ is centered at the point $\b{x}\in\R^D$ and the simplest Riemannian metric is the Euclidean $\b{M}_\X(\b{x})=\Id_D$.
%This implies that the Euclidean space can be seen as a Riemannian manifold.
However, we are able to use other suitable metrics that simply change the way we measure distances in $\X$ (see Sec.~\ref{section:data_learned_riemannian_manifolds}). For instance, given a set of points in $\X$ we can construct a metric with small magnitude near the data, 
such that to pull the shortest paths towards them (see Fig.~\ref{intro:fig:manifold_examples} bottom).

For a $d$-dimensional embedded manifold $\M\subset \X$, a collection of \emph{chart maps} $\phi_i: \U_i\subset \M \rightarrow \R^{d}$ is used to assign \emph{local intrinsic coordinates} to neighborhoods $\U_i\subset \M$, and for simplicity, we assume that a global chart map $\phi(\cdot)$ exists. By definition, when $\M$ is smooth the $\phi(\cdot)$ and its inverse $\phi^{-1}:\phi(\M)\subset \R^{d} \rightarrow \M\subset\X$ exist and are smooth maps. Thus, a $\b{v}_\b{x}\in\mathcal{T}_{\b{x}}\M$ can be expressed as $\b{v}_\b{x} = \b{J}_{\phi^{-1}}(\b{z})\b{v}_{\b{z}}$, where $\b{z} = \phi(\b{x}) \in\R^{d}$ and $\b{v}_{\b{z}}\in\R^{d}$ are the representations in the intrinsic coordinates. Also, the Jacobian $\b{J}_{\phi^{-1}}(\b{z}) \in\R^{D \times d}$ defines a basis that spans the $\mathcal{T}_\b{x}\M$, and thus, we can represent the metric $\b{M}_\X(\cdot)$ in the intrinsic coordinates as
\begin{align}
\label{eq:inner_product_ambient_chart}
\hspace*{-1.45mm}\inner{\b{v}_\b{x}}{\b{v}_\b{x}}_\b{x} = \inner{\b{v}_\b{z}}{\b{J}_{\phi^{-1}}(\b{z})^\T \b{M}_\X(\phi^{-1}(\b{z}))\b{J}_{\phi^{-1}}(\b{z})\b{v}_\b{z}} = \inner{\b{v}_\b{z}}{\b{M}(\b{z})\b{v}_\b{z}}=\inner{\b{v}_\b{z}}{\b{v}_\b{z}}_\b{z},
\end{align}
with smooth $\b{M}(\b{z}) =\b{J}_{\phi^{-1}}(\b{z})^\T \b{M}_\X(\phi^{-1}(\b{z}))\b{J}_{\phi^{-1}}(\b{z})\in\R^{d\times d}_{\succ 0}$. As we discuss below, we should be able to evaluate the intrinsic $\b{M}(\b{z})$ in order to find length minimizing curves on $\M$. But, when $\M$ is embedded the chart maps are usually unknown, as well as a global chart rarely exists. In contrast, for ambient like manifolds the global chart is $\phi(\b{x}) = \b{x}$, which is convenient to use in practice.

Generally, one of the main utilities of a Riemannian manifold $\M\subseteq\X$ is to enable us compute shortest paths therein. Intuitively, the norm $\sqrt{\inner{d\b{x}}{d\b{x}}_\b{x}}$ represents how the infinitesimal displacement vector $d\b{x} \approx \b{x}' - \b{x}$ on $\M$ is locally scaled. Thus, for a curve $\gamma:[0,1]\rightarrow \M$ that connects two points $\b{x}=
\gamma(0)$ and $\b{y}=\gamma(1)$, the length on $\M$ or equivalently in $\phi(\M)$ using Eq.~\ref{eq:inner_product_ambient_chart}, is measured as
\begin{align}
\length{\gamma(t)} = \int_0^1 \sqrt{\inner{\dot{\gamma}(t)}{\dot{\gamma}(t)}_{\b{\gamma}(t)}} dt = \int_0^1\sqrt{\inner{\dot{c}(t)}{\b{M}(c(t)) \dot{c}(t)} }dt = \length{c(t)}
\end{align}
where $\dot{\gamma}(t)~=~\partial_t \gamma(t) \in \mathcal{T}_{\gamma(t)}\M$ is the velocity of the curve and accordingly $\dot{c}(t) \in \mathcal{T}_{c(t)}\phi(\M)$. The minimizers of this functional are the \emph{shortest paths}, also known as \emph{geodesics}. We find them by solving a system of 2\textsuperscript{nd} order nonlinear ordinary differential equations (ODEs) defined in the intrinsic coordinates. Notably, for ambient like manifolds the trivial chart map enables us to compute the shortest paths in practice by solving the ODEs system. In some sense, the behavior of the shortest paths is to avoid high magnitude $\sqrt{\abs{\b{M}(c(t))}}$ regions.  In general, the analytical solution is intractable, so we rely on approximate solutions \citep{ HennigAISTATS2014, yang:arxiv:2018,arvanitidis:aistats:2019}.  For further information on Riemannian geometry and the ODEs system see Appendix~\ref{appendix:sec:geometry_theory}.

\subsection{Unifying the two manifold perspectives}
\label{section:unifying_manifolds}

In all related works, the ambient space $\X$ is considered as a Euclidean space. Instead, we propose to consider $\X$ as a Riemannian manifold. This allows us to encode high-level information through the associated metric, which constitutes an interpretable way to control the shortest paths. In order to find such a path on an embedded $\M\subset \X$, we have to solve the system of ODEs defined in the intrinsic coordinates. But, when $\M$ is embedded the chart maps are mostly unknown. Hence, the usual trick is to utilize another manifold $\Z$ having a trivial chart map and to represent the geometry of $\M$ therein. Thus, we find the curve in $\Z$, which corresponds to the actual shortest path on $\M$.

%Interestingly, we combine the two views of Riemannian manifolds in order to compute shortest paths on embedded $\M\subset\X$. A shortest path is the solution to an ODEs system defined in the intrinsic coordinates, but for embedded $\M$ the chart maps are usually unknown. Instead, we utilize another manifold $\Z$ for which we know the chart map and we represent the geometry of $\M$ therein. Thus, we compute the shortest path in $\Z$ while respecting the intrinsic geometry of $\M$. Our main idea is to consider the ambient space as a Riemannian manifold, which allows to encode high-level information and control the shortest paths, while in previous works the $\X$ is considered as a Euclidean space.

%  which correspond to shortest paths on 

%This is the first time that this mathematical theory is applied rigorously to data learned manifolds

%This constitutes the mathematical theory over which we base our main contribution, and in particular, we show how we apply this theory on general data driven manifolds. 

In particular, assume an embedded $d$-dimensional manifold $\M\subset\X$ within a Riemannian manifold $\X=\R^D$ with metric $\b{M}_\X(\cdot)$, a Euclidean space $\Z=\R^{d_\Z}$ called as \emph{latent space} and a smooth function $g:\Z \rightarrow \X$ called as \emph{generator}. Since $g(\cdot)$ is smooth, $\M_\Z = g(\Z)\subset \X$ is an immersed $d_\Z$-dimensional smooth (sub)manifold\footnote{A function $g:\Z\rightarrow \M\subset\X$ is an immersion if the $\b{J}_g:\mathcal{T}_\b{z}\Z \rightarrow \mathcal{T}_{g(\b{z})}\M$ is injective (full rank) $\forall\b{z}\in\Z$. Intuitively, $\M$ is a $\text{dim}(\Z)$-dimensional surface and intersections are allowed. While $g(\cdot)$ is an embedding if it is an injective function, which means that interections are not allowed. An immersion locally is an embedding.}. In general, we assume that $d_\Z =d$ and also that $g(\cdot)$ approximates closely the true embedded $\M_\Z \approx \M$, while if $d_\Z < d$ then $g(\cdot)$ can only approximate a submanifold on $\M$. Consequently, the corresponding Jacobian matrix $\b{J}_g(\b{z}) \in \R^{D \times d_\Z}$ is a basis that spans the $\mathcal{T}_{\b{x}=g(\b{z})}{\M_\Z}$, and maps a tangent vector $\b{v}_\b{z} \in \tangent{\b{z}}{\Z}$ to a tangent vector $\b{v}_\b{x} \in \mathcal{T}_{\b{x}=g(\b{z})}\M_\Z$. Thus, as before the restriction of $\b{M}_\X(\cdot)$ on $\tangent{\b{x}}{\M_\Z}$ induces a metric in the latent space $\Z$ as
\begin{align}
\label{eq:new_pull_back_metric}
\inner{\b{v}_\b{x}}{\b{v}_\b{x}}_\b{x} = \inner{\b{v}_\b{z}}{\b{J}_g(\b{z})^\T \b{M}_\X(g(\b{z}))\b{J}_g(\b{z})\b{v}_\b{z}} = \inner{\b{v}_\b{z}}{\b{M}(\b{z})\b{v}_\b{z}}.
\end{align}
This Riemannian metric $\b{M}(\b{z})$ is known as the \emph{pull-back} metric, and essentially, captures the intrinsic geometry of the immersed $\M_\Z$, while taking into account the geometry of $\X$. The space $\Z$ together with $\b{M}(\b{z})$ constitutes a Riemannian manifold, but since  $\Z=\R^{d_\Z}$ the chart map and $\tangent{\b{z}}{\Z}$ are trivial. Therefore, we can evaluate the metric $\b{M}(\b{z})$ in intrinsic coordinates, which enables us to compute shortest paths on $\Z$ by solving the ODEs system. Intuitively, these paths in $\Z$ move optimally on $\M_\Z$, while simultaneously respecting the geometry of the ambient space $\X$. Also, note that $g(\cdot)$ is not a chart map, and hence, it is easier to learn. For further discussion see Appendix~\ref{appendix:manifold_theory}.

%The resulting paths respect the intrinsic geometry of the immersed $\M_\Z$ that lies into a Riemannian manifold $\X$. Intuitively, the shortest path in $\Z$ will move optimally on $\M_\Z$, while respecting the geometry of the ambient space through the ambient metric $\b{M}_\X(\cdot)$.

%% file: methods.tex
%!TEX root = ./paper.tex

\section{Data learned Riemannian manifolds}
\label{section:data_learned_riemannian_manifolds}

We discuss some usages of differential geometry in machine learning, which largely inspire our work. Briefly, we present previous Riemannian metric learning methods and we propose a simple technique to construct such metrics in the ambient space $\X$. This is a principled and interpretable way to encode domain knowledge in our models. Also, we present the related work where the structure of an embedded data manifold is properly captured in the latent space of stochastic generators.

%In this section we show the ways differential geometry theory presented in Sec.~\ref{sec:geometry} in a machine learning context. Briefly, we learn various Riemannian metrics in the ambient space $\X$ using the given data, which encode high-level information. These metrics provide a principled and interpretable way to include easily domain knowledge in our model.  Then, we pull-back properly the ambient Riemannian metric into the latent space $\Z$ of both stochastic and deterministic deep generative models.

\subsection{Learning Riemannian Metrics in the Ambient Space}
\label{section:methods_ambient_metric}

Assume that a set of points $\{\b{x}_n\}_{n=1}^N$ in $\X=\R^D$ is given. The Riemannian metric learning task is to learn a positive definite metric tensor $\b{M}_\X:\X \rightarrow \R^{D\times D}_{\succ 0}$ that changes smoothly across the space. The actual behavior of the metric depends on the problem we want to model. For example, when we want the shortest paths to stay on the data manifold (see Fig.~\ref{intro:fig:manifold_examples} bottom) the meaningful behavior for the metric is that the magnitude $\sqrt{|\b{M}_\X(\cdot)|}$ should be small near the data manifold and large as we move away. Similarly, in Fig.~\ref{sec:intro:fig:teaser} the ambient metric is designed such that its magnitude is small near the data points with blond hair, and thus, the shortest paths tend to follow this semantic constraint.

One of the first approaches to learn such a Riemannian metric was presented by \citet{hauberg:nips:2012}, where $\b{M}_\X(\b{x})$ is the convex combination of a predefined set of metrics, using a smooth weighting function. In particular, at first $K$ metrics are estimated $\{\b{M}_k\}_{k=1}^K \in \R^{D\times D}_{\succ 0}$ centered at the locations $\{\b{c}_k\}_{k=1}^K\in\R^D$. Then, we can evaluate the metric at a new point as
\begin{equation}
\label{eq:convex_combination_metric}
\b{M}_\X(\b{x}) = \sum_{k=1}^K w_k(\b{x}) \b{M}_k,~ \text{with}~ w_k(\b{x}) = \frac{\tilde{w}_k(\b{x})}{\sum_{j=1}^K \tilde{w}_j(\b{x})}~\text{ and }~\tilde{w}_k(\b{x}) = \exp\left( -\frac{\norm{\b{x} - \b{c}_k}_2^2}{2 \sigma^2}\right),
\end{equation}
where the kernel $\tilde{w}_k(\b{x})$ with bandwidth $\sigma\in\R_{>0}$ is a smooth function, and thus, the metric is smooth as a linear combination of smooth functions. A practical example for the base metrics $\b{M}_k$ is the local Linear Discriminant Analysis (LDA), where local metrics are learned using labeled data such that to separate well the classes locally \citep{Hastie94discriminantadaptive}. Also, a related approach is the Large Margin Nearest Neighbor (LMNN) classifier \citep{weinberger:2006:neurips}. Note that the domain of metric learning provides a huge list of options that  can be considered \citep{surez2018tutorial}.

Similarly, in an unsupervised setting \citet{arvanitidis:nips:2016} proposed to construct the Riemannian metric in a non-parametric fashion as the the inverse of the local diagonal covariance. In particular, for a given point set $\{\b{x}_n\}_{n=1}^N$ at a point $\b{x}$ the diagonal elements of the metric $\b{M}_\X(\cdot)$ are equal to 
\begin{align}
\label{eq:local_pca_metric}
M_{\X_{dd}}(\b{x}) = \left( \sum_{n=1}^N w_n(\b{x})(x_{nd}- x_d)^2 +\varepsilon\right)^{-1} \quad\text{ with }\quad  w_n(\b{x}) = \exp\left(-\frac{ \norm{\b{x}_n - \b{x}}_2^2}{2\sigma^2}\right),
\end{align}
where $\sigma\in\R_{>0}$ controls the curvature of the Riemannian manifold i.e., how fast the metric changes, and $\varepsilon>0$ is a small scalar to upper bound the metric. Although these are quite flexible and intuitive metrics, selecting the parameter $\sigma$ is a challenging task \citep{arvanitidis:gsi:2017}, especially due to the curse of dimensionality \citep{Bishop:2006:PRM} and the sample size $N$ due to the non-parametric regime.

\paragraph{The proposed Riemannian metrics.} Inspired by the approaches described above and \citet{peyre:cg:2010}, we propose a general and simple technique to easily construct metrics in $\X$, which allows to encode information depending on the problem. An unsupervised diagonal metric can be defined as
\begin{align}
\label{eq:ambient_metric_density}
\b{M}_\X(\b{x})=(\alpha \cdot h(\b{x}) + \varepsilon)^{-1} \cdot \Id_D
\end{align}
where $h(\b{x}):\R^D\rightarrow \R_{>0}$ with behavior $h(\b{x}) \rightarrow 1$ when $\b{x}$ is near the data manifold, otherwise  $h(\b{x}) \rightarrow 0$, and $\alpha, \epsilon >0$  are scaling factors to lower and upper bound the metric, respectively. One simple but very effective approach is to use a positive Radial Basis Function (RBF) network  \citep{que:aistats:2016} as $h(\b{x}) = \b{w}^\T \bs{\phi}(\b{x})$ where $\b{w} \in\R^{K}_{>0}$ and $\phi_k(\b{x}) = \exp(-0.5 \cdot \lambda_k \cdot \norm{\b{x} - \b{x}_k}_2^2)$ with bandwidth $\lambda_k >0$. Similarly, $h(\b{x})$ can be the probability density function of the given data. Usually, the true density function is unknown and difficult to learn, but we can approximate it roughly by utilizing a simple model as the Gaussian Mixture Model (GMM) \citep{Bishop:2006:PRM}. Such a Riemannian metric pulls the shortest paths towards areas of $\X$ with high $h(\b{x})$ (see Fig.~\ref{intro:fig:manifold_examples} bottom). 
%where $p(\b{x})$ is a probability density function in $\X$, the $\varepsilon >0$ is a small scalar to upper bound the metric when $p(\b{x})\rightarrow 0$, and accordingly, $\alpha>0$ is a scaling factor that can be used to set a lower bound to the metric when $p(\b{x})\rightarrow \max_\b{x}p(\b{x})$ that is especially useful when the $p(\b{x})$ is unnormalized (for details see Appendix~\ref{appendix:metric_construction}). Obviously, learning the density $p(\b{x})$ is a rather difficult task, however, we can rely on simple density estimation models as the Gaussian Mixture Model (GMM) \citep{Bishop:2006:PRM} or in  some cases the $p(\b{x})$ can be given by the problem. Such a metric will pull the shortest paths towards areas of $\X$ with high density and for an example see the top panel of Fig.~\ref{intro:teaser_fig}.

In a similar context, a supervised version can be defined where the function $h(\b{x})$ represents cost, while in Eq.~\ref{eq:ambient_metric_density} we do not use the inversion. In this way, shortest paths will tend to avoid regions of the ambient space $\X$ where the cost function is high. For instance, in Fig.~\ref{sec:intro:fig:teaser} we can think of a cost function that is high over all the non-blonde data points.  Of course, such a cost function can be learned as an independent regression or classification problem, while in other cases can even be given by the problem or a domain expert. For further details about all the metrics above see Appendix~\ref{appendix:metric_construction}.

\subsection{Learning Riemannian Metrics in the Latent Space}
\label{section:methods_latent_metric}

As discussed in Sec.~\ref{section:unifying_manifolds}, we can capture the geometry of the given embedded data manifold $\M\subset\X$ by learning a smooth generator $g:\Z\rightarrow\M_\Z\subset \X$ such that $\M_\Z\approx \M$. In previous works has been shown how to learn in practice such a function $g(\cdot)$, and also, the mild conditions it has to follow so that the induced Riemannian metric to capture properly the structure of $\M$. In the latent space $\Z=\R^d$ we call as latent codes or representations the points  $\{\b{z}_n\in\Z ~|~ \b{x}_n = g(\b{z}_n), ~n=1,\dots,N\}$.

First \citet{tosi:uai:2014} considered the Gaussian Process Latent Variable Model (GP-LVM) \citep{Lawrence:2005:jmlr}, where $g(\cdot)$ is a stochastic function defined as a multi-output Gaussian process $g \sim \mathcal{GP}(\b{m}(\b{z}), k(\b{z}, \b{z}'))$. Since the generator is stochastic, it induces a random Riemannian metric in $\Z$, and in practice, the expected metric is used for the computation of shortest paths. The advantage of such a stochastic generator is that the metric magnitude increases analogous to the uncertainty of $g(\cdot)$, which happens in regions of $\Z$ where there are no latent codes. Apart from this desired behavior, this metric is not very practical due to the  GPs computational cost.

Another set of approaches known as deep generative models, parameterizes $g(\cdot)$ as a deep neural network (DNN). On the one hand are the explicit density models, where the marginal likelihood can be computed, with main representatives the Variational Auto-Encoder (VAE) \citep{kingma:iclr:2014, rezende:icml:2014} and the normalizing flow models \citep{dinh:2016:arxiv, rezende:icml:2015}. On the other hand are the implicit density models for which the marginal likelihood is intractable, as is the Generative Adversarial Networks (GAN) \citep{goodfellow:neurips:2014}.

Recently, \citet{arvanitidis:iclr:2018} showed that we are able to properly capture the structure of the data manifold $\M\subset \X$ in the latent space $\Z$ of a VAE under the condition of having meaningful uncertainty quantification for the generative process. In particular, the standard VAE assumes a Gaussian likelihood $p(\b{x}~|~\b{z}) = \N(\b{x}~|~\mu(\b{z}),~\Id_D\cdot \sigma^2(\b{z}))$ with a prior $p(\b{z})$. Hence, the generator can be written as $g(\b{z}) = \mu(\b{z}) + \text{diag}(\varepsilon)\cdot\sigma(\b{z})$ where $\varepsilon \sim \N(\b{0}, \Id_D)$ and $\mu:\Z\rightarrow\X, ~\sigma:\Z\rightarrow \R^{D}_{>0}$ are usually parametrized with DNNs. However, parametrizing $\sigma(\cdot)$ with a DNN does not directly imply meaningful uncertainty quantification, because it extrapolates arbitrarily to regions of $\Z$ with no latent codes. Thus, the proposed solution in \citet{arvanitidis:iclr:2018} is to model the inverse variance $\beta(\b{z})  = (\sigma^2(\b{z}))^{-1}$ with a positive Radial Basis Function (RBF) network \citep{que:aistats:2016}, which implies that moving further from the latent codes increases the uncertainty. Under this stochastic generator, the expected Riemannian metric in $\Z$ is equal to
\begin{align}
\label{eq:latent_expected_metric}
\b{M}(\b{z}) = \E_{p(\varepsilon)}[\b{J}_{g_\varepsilon}(\b{z})^\T\b{J}_{g_\varepsilon}(\b{z})]=\b{J}_\mu(\b{z})^\T\b{J}_\mu(\b{z}) + \b{J}_\sigma(\b{z})^\T\b{J}_\sigma(\b{z}),
\end{align}
where $g_\varepsilon(\cdot)$ implies that $\varepsilon$ is kept fixed $\forall~\b{z}\in \Z$, such that to ensure a smooth mapping. Here, we observe that the metric increases when the generator becomes uncertain due to the second term. This constitutes a desired behavior, as the metric informs us to avoid regions of $\Z$ where there are no latent codes, which directly implies that these regions do not correspond to parts of the data manifold in $\X$. In some sense, we can think of modeling the topology of $\M$ too \citep{hauberg:only:2018}.

Clearly, the deterministic generators like the Auto-Encoder (AE) and the GAN, capture poorly the structure of $\M$ in $\Z$ since the second term in Eq.~\ref{eq:latent_expected_metric} does not exist. The reason is that these models are trained based on the likelihood $p(\b{x}~|~\b{z})=\delta(\b{x} - g(\b{z}))$, and hence, the uncertainty is not quantified. Of course, for the AE one potential \emph{heuristic} solution is to use the latent codes of the training data to fit \emph{post-hoc} a meaningful variance estimator under the Gaussian likelihood and the maximum likelihood principle as $\theta^* = \argmax_\theta \prod_{n=1}^N \N(\b{x}_n ~|~g({e}(\b{x}_n)), ~\Id_D\cdot \sigma^2_\theta({e}(\b{x}_n)))$, with encoder ${e}:\X\rightarrow\Z$. In principle, we could follow the same procedure for the GAN by learning an encoder \citep{donahue2016adversarial, dumoulin2016adversarially}. However, it is still unclear if the encoder for a GAN learns meaningful representations or if the powerful generator ignores the inferred latent codes \citep{arora:iclr:2018}.

Therefore, in order to properly capture the structure of $\M$ in $\Z$ we mainly rely on stochastic generators with increasing uncertainty as we move further from the latent codes. Even if the RBF based approach is a meaningful way to get the desired behavior, in general, uncertainty quantification with parametric models is still considered as an open problem \citep{mackay:1992, Gal2015DropoutB, lakshminarayanan:neurips:2017, detlefsen2019reliable, arvanitidis:iclr:2018}. Nevertheless, \citet{eklund:arxiv:2019} showed that the expected Riemannian metric in Eq.~\ref{eq:latent_expected_metric} is a reasonable approximation to use in practice. Obviously, when $g(\cdot)$ is deterministic, like the GAN, the second term in Eq.~\ref{eq:latent_expected_metric} disappears, since these models do not quantify the uncertainty of the generative process. This directly means that deterministic generators are not able, by construction, to properly capture the geometric structure of $\M$ in the latent space, and hence, exhibit a misleading bias \citep{hauberg:only:2018}.

\section{Enriching the Latent Space with Geometric Information}

%Previously, we have seen how to construct a Riemannian metric $\b{M}_\X(\cdot)$ in the ambient space in Sec.~\ref{section:methods_ambient_metric} and in Sec.~\ref{section:methods_latent_metric} how to capture in the latent space $\Z$ the geometry of an embedded data manifold $\M\subset\X$ using a stochastic generator $g(\cdot)$. The next step is to combine these two approaches such that to use additionally the geometry of the ambient space that is modeled by $\b{M}_\X(\cdot)$. Intuitively, the new metric in $\Z$ will take into account both the distortions induced by $g(\cdot)$ that represents the geometry of the data manifold, together with the geometrical properties of the ambient space.

Here, we unify the approaches presented in Sec.~\ref{section:methods_ambient_metric} and  Sec.~\ref{section:methods_latent_metric}, in order to provide extra structure in the latent space of a generative model. This is the first time that these two fundamentally different Riemannian views are combined. Their difference is that the metric induced by $g(\cdot)$ merely tries to capture the intrinsic geometry of the given data manifold $\M$, while $\b{M}_\X(\cdot)$ allows to directly encode high-level information in $\X$ based on domain knowledge. Moreover, we provide in the stochastic case a relaxation for efficient computation of the expected metric. While in the deterministic case we combine a carefully designed ambient metric with a new architecture for $g(\cdot)$ to extrapolate meaningfully, which is one way to ensure well-behaved shortest paths that respect the structure of $\M$.

%which constitutes one way to properly capture the structure of the data manifold.

\paragraph{Stochastic generators.} Assuming that an ambient $\b{M}_\X(\cdot)$ is given (see Sec.~\ref{section:methods_ambient_metric}), we learn a VAE with Gaussian likelihood, so the stochastic mapping is $g(\b{z})= \mu(\b{z}) + \text{diag}(\varepsilon)\cdot\sigma(\b{z})$, while using a positive RBF for meaningful estimation of the uncertainty $\sigma(\cdot)$. As before, we assume that $\varepsilon$ is constant for each $\M_\varepsilon=g_\varepsilon(\Z)$ to ensure smoothness. Therefore, we can apply Eq.~\ref{eq:new_pull_back_metric} to derive the new stochastic more informative pull-back metric in $\Z$, which is equal to
\begin{align}
\label{eq:pull_back_with_generator_stochastic}
{\b{M}_\varepsilon}(\b{z} ) = \left[\b{J}_\mu(\b{z}) +  \b{J}_\sigma(\b{z})\varepsilon\right]^\T \b{M}_\X\left(\mu(\b{z}) + \text{diag}(\varepsilon)\cdot\sigma(\b{z}) \right) \left[\b{J}_\mu(\b{z}) + \b{J}_\sigma(\b{z})\varepsilon\right].
\end{align}
Since this is a random metric, in principle, we can compute the expectation simply by sampling $\varepsilon\sim\N(\b{0},\Id_D)$, as ${\b{M}}(\b{z}) = \E_{\varepsilon \sim p(\varepsilon)}[{\b{M}_\varepsilon}(\b{z})]$. 
%Intuitively, for each fixed $\varepsilon$ the $g_\varepsilon(\Z)$ immerses a smooth $\M_\varepsilon$ in $\X$. The geometry of $\M_\varepsilon$  is captured by the Jacobian of the generator. On the same time, for the inner product on $\mathcal{T}_{g_\varepsilon(\b{z})}{\M_\varepsilon}$ we use the restriction of the ambient metric $\b{M}_\X(g_\varepsilon(\b{z}))$. Since the ambient metric encodes information about the problem of interest, the geometry that we capture in the latent space will be enriched accordingly. Essentially, $\b{M}_\X(\cdot)$ provides additional information such that to move optimally on $\M_\varepsilon$, while respecting the geometry of the ambient space. Therefore, for each point $\b{z}$ in the latent space the expected Riemannian metric represents the average behavior over all of the sampled $\M_\varepsilon$ immersed in $\X$ and regulated by the $\b{M}_\X(\cdot)$.
However, in practice this expectation will increase dramatically the cost, especially, since we need to evaluate the expected metric many times when computing a shortest path. Hence, we consider only the $\E_{\varepsilon \sim p(\varepsilon)}[\M_\varepsilon] = \mu(\b{z})$ for the evaluation of the ambient metric $\b{M}_\X(\cdot)$ in Eq.~\ref{eq:pull_back_with_generator_stochastic}, which simplifies the expected Riemannian metric to
\begin{align}
\label{eq:pull_back_with_generator_stochastic_approximated}
{\b{M}}(\b{z} ) \triangleq \b{J}_\mu(\b{z})^\T \b{M}_\X\left(\mu(\b{z})\right) \b{J}_\mu(\b{z}) + \b{J}_\sigma(\b{z})^\T \b{M}_\X\left(\mu(\b{z})\right) \b{J}_\sigma(\b{z}).
\end{align}
Essentially, the realistic underlying assumption is that near the latent codes $\sigma(\b{z})\rightarrow\b{0}$ so $g(\b{z}) \rightarrow \mu(\b{z})$ and the first term dominates. But, as we move in regions of $\Z$ with no codes the $\sigma(\b{z}) \gg 0$, and for this reason, we need the second term in the equation. In particular, $\b{J}_\sigma(\cdot)$ dominates when moving further from the latent codes, and hence, the behavior of $\b{M}_\X(\cdot)$ will be less important there. Thus, we are allowed to consider this relaxation, for which the meaningful uncertainty estimation is still necessary. We further analyze and check empirically this relaxation in Appendix~\ref{appendix:section:stochastic_vs_deterministic}.

%\marginpar{\gear{This paragraph need to be seen again. Why a surrogate model is not good? Is simple an heuristic and does not maximize any principled quantity like maximum likelihood of VAE}}

%Consequently, when $g(\cdot)$ is deterministic the relaxation of Eq.~\ref{eq:pull_back_with_generator_stochastic_approximated} is not necessary. However, this comes with the price of underestimating the structure of $\M$.

\paragraph{Deterministic generators.}  As regards the deterministic generators, we propose a simple solution that ensures well-behaved shortest paths which respect the structure of the given data manifold  $\M$. The idea is to learn a Riemannian metric $\b{M}_\X(\cdot)$ in $\X$ that only roughly represents the structure of $\M$, for instance, by using an RBF or a GMM  (Eq.~\ref{eq:ambient_metric_density}). Essentially, this ambient metric informs us how close the generated $\M_\Z$ is to the given data. Hence, we additionally need the $g(\cdot)$ to extrapolate meaningfully. That means $g(\cdot)$ should learn to generate well the given data from a prior $p(\b{z})$, but as we move further from the support of $p(\b{z})$, the generated $\M_\Z$ should also move further from the given data in $\X$. Consequently, since $\b{M}_\X(\cdot)$ is designed to increase far from the given data, the induced Riemannian metric in $\Z$ properly captures the structure of the data manifold.

One of the simplest deterministic generators with this desirable behavior is the probabilistic Principal Component Analysis (pPCA) \citep{tipping:1999:ppca}. This is a very basic model with a Gaussian prior $p(\b{z}) = \mathcal{N}(\mathbf{0},~\Id_d)$ and a generator that is simply a linear map. The generator  is constructed by the top $d$ eigenvectors of the empirical covariance matrix, scaled by their eigenvalues. Inspired by this simple model, we propose for the deterministic generator the following architecture
\begin{align}
\label{eq:resnet_generator}
g(\b{z}) = f(\b{z}) + \b{U}\cdot \text{diag}([\sqrt{\lambda_1},\dots,\sqrt{\lambda_d}]) \cdot \b{z} + \b{b},
\end{align}
where $f:\Z\rightarrow\X$ is a deep neural network, $\b{U}\in\R^{D\times d}$ the top $d$ eigenvectors with their corresponding  eigenvalues $\lambda_d$ computed from the empirical data covariance, and $\b{b} \in\R^D$ is the data mean. This interpretable model can be seen as a residual network (ResNet) \citep{he:2015:arxiv}. In particular, the desired behavior is that as we move further from the $p(\b{z})$ the linear part of Eq.~\ref{eq:resnet_generator} becomes the dominant one, especially, when bounded activation functions are utilized for $f(\cdot)$. Hence, the $\M_\Z$ will extrapolate meaningfully as we move further from the support of the prior. However, we need again the generated $\M_\Z$ to be a valid immersion. For further discussion see Appendix~\ref{appendix:manifold_theory}.

%% file: experiments.tex
%!TEX root = ./paper.tex

\section{Experiments}
\label{section:experiments}

\subsection{Demonstrations with Deterministic Generators}

\begin{wrapfigure}{r}{0.3\textwidth}
	\vspace{-20pt}
	
	\begin{tikzpicture}
	\draw (0, 0) node[inner sep=0] {\includegraphics[width=\linewidth]{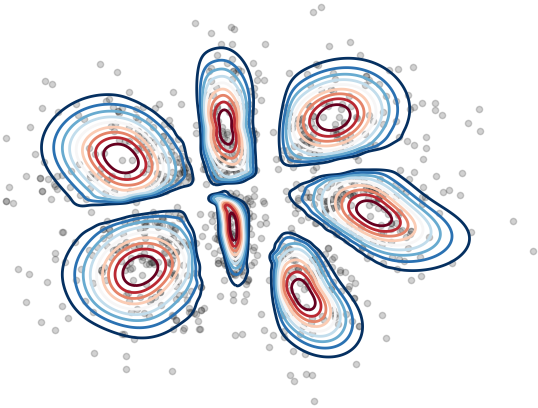}};
	\draw (-1.3, 1.2) node {\tiny LANDs};
	\end{tikzpicture}
	
	\vspace{5pt}
	
	\begin{tikzpicture}
	\draw (0, 0) node[inner sep=0] {\includegraphics[width=\linewidth]{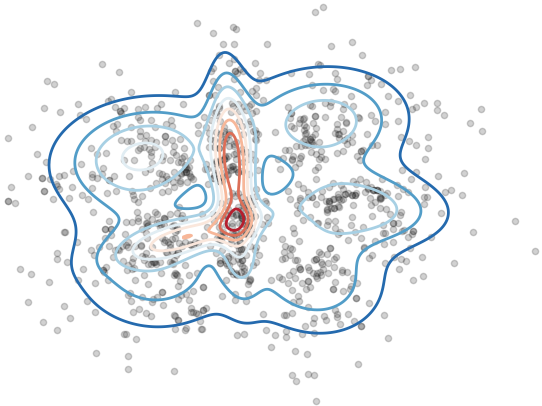}};
	\draw (-1.3, 1.2) node {\tiny GMM};
	\end{tikzpicture}
	
	\vspace{-5pt}	
	\caption[none]{}
	\label{experiments:land_vs_gmm}
	\vspace{-20pt}
\end{wrapfigure}
\begin{figure}[t]
	\centering
	\hspace{-25pt}
	\begin{subfigure}[b]{0.23\textwidth}
		{\includegraphics[scale=0.22]{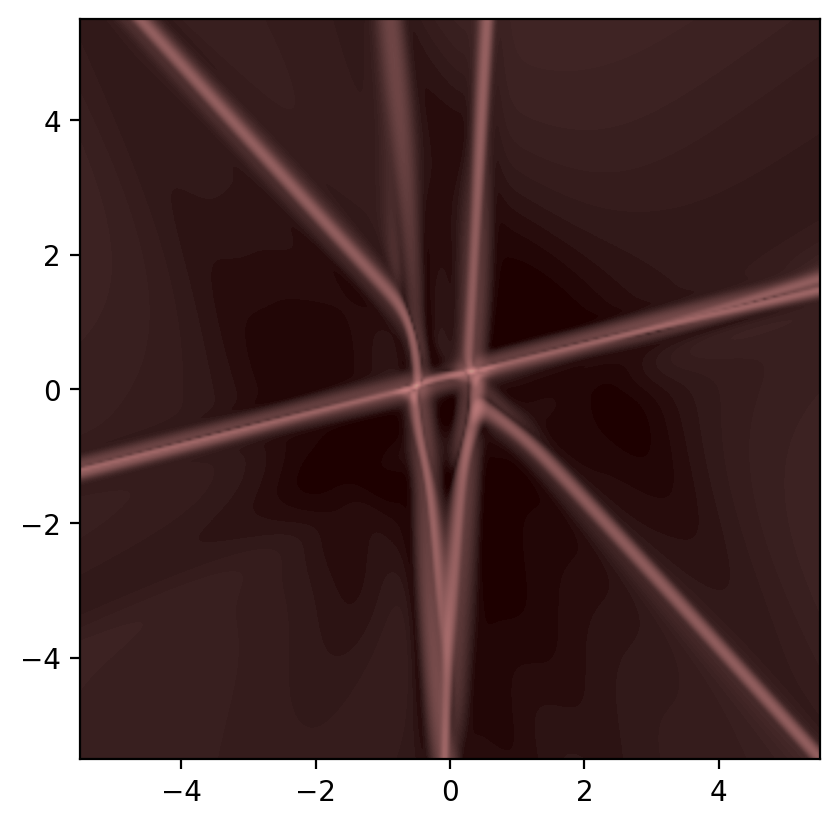}}
	\end{subfigure}
	\hspace{-24pt}
	\begin{subfigure}[b]{0.23\textwidth}
		{\includegraphics[scale=0.22]{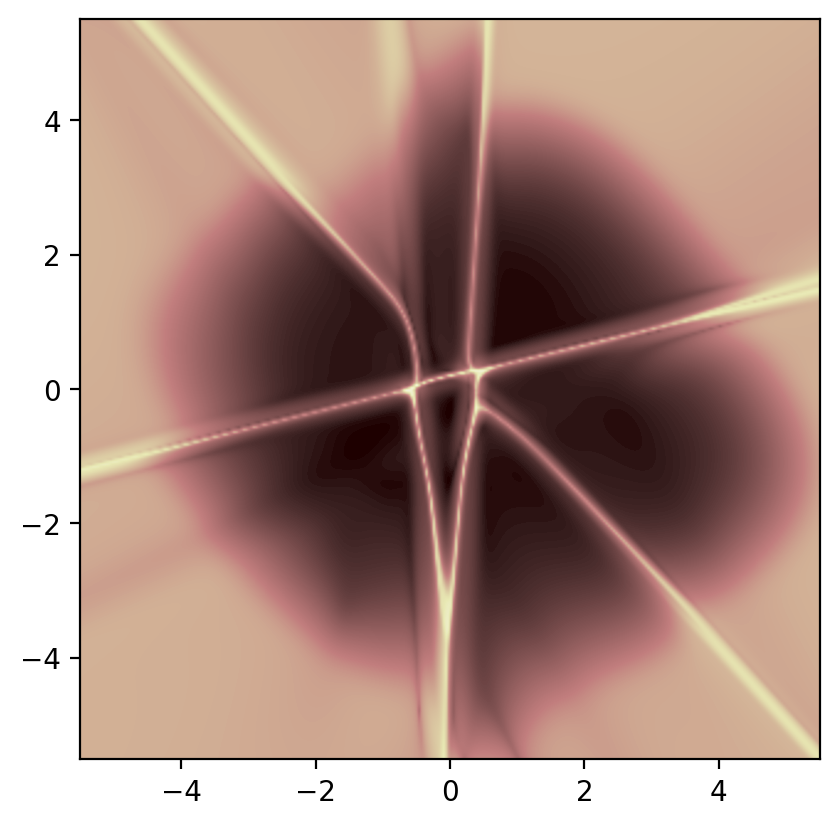}}
		%		\caption{AE}
		%		\label{fig:AE}
	\end{subfigure}
	\hspace{-24pt}
	\begin{subfigure}[b]{0.23\textwidth}
		{\includegraphics[scale=0.22]{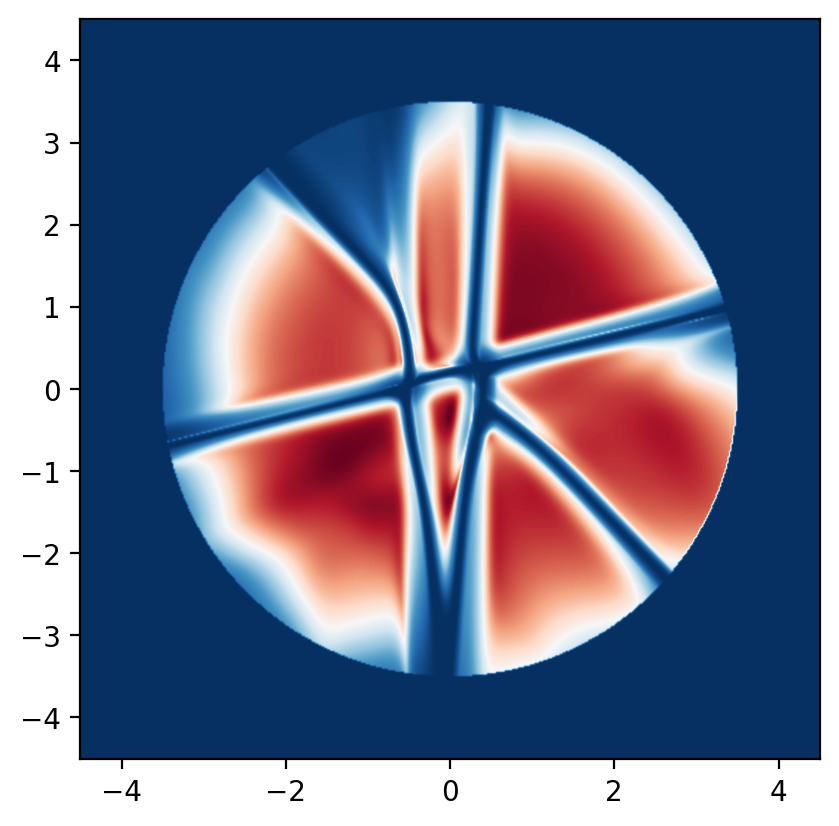}}
		%		\caption{AE}
		%		\label{fig:AE}
	\end{subfigure}
	\hspace{-25pt}
	\begin{subfigure}[b]{0.23\textwidth}
		{\includegraphics[width=\textwidth]{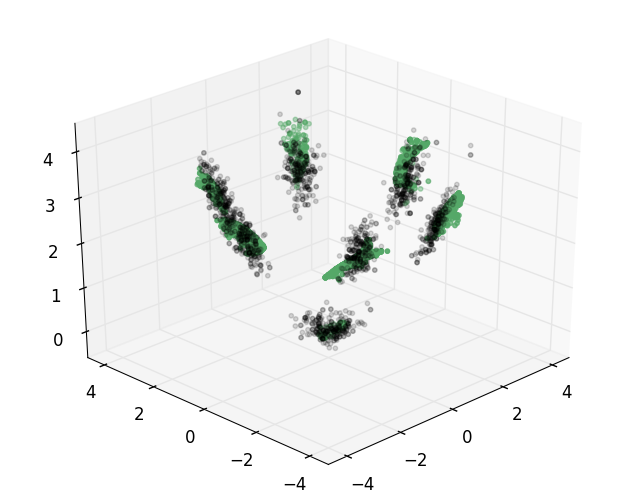}}
		%		\caption{AE}
		%		\label{fig:AE}
	\end{subfigure}
	\hspace{0pt}
	\begin{subfigure}[b]{0.23\textwidth}
		{\includegraphics[width=\textwidth]{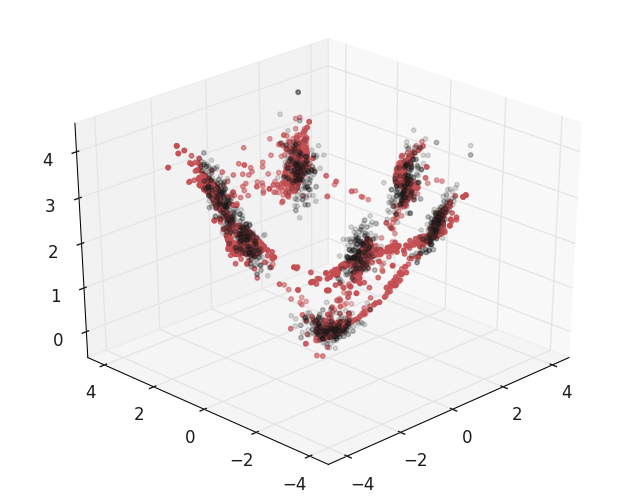}}
		%		\caption{AE}
		%		\label{fig:AE}
	\end{subfigure}
	\hspace{-30pt}
	\caption[none]{From \emph{left} to \emph{right}: The measure $\sqrt{\abs{\b{M}(\b{z})}}$ without and with the ambient Riemannian metric, the approximate aggregated posterior $q(\b{z})$, generated samples by the $q(\b{z})$ ({\protect \begin{tikzpicture}
			\draw  [my_green, fill=my_green,  radius=3pt] circle;
			\end{tikzpicture}}) and the prior $p(\b{z})$ ({\protect \begin{tikzpicture}
			\draw  [my_red, fill=my_red,  radius=3pt] circle;
			\end{tikzpicture}}). Comparing the first two panels, we see that the proposed model properly captures the structure of $\M$.}
	\label{experiments:gan_synthetic}
\end{figure}
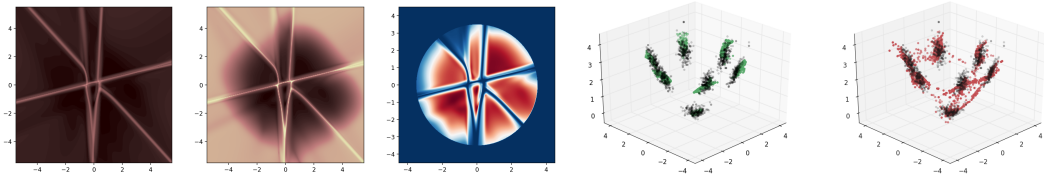

\paragraph{Synthetic experiment. }Usually, in GANs the $g(\cdot)$ is a continuous function, so we expect some generated points to fall off the given data support. Mainly, when the data lie near a disconnected $\M$ and irrespective of training optimallity. This is known as the distribution mismatch problem. We generate a synthetic dataset (Fig.~\ref{experiments:gan_synthetic}) and we train a Wasserstein GAN \citep{arjovsky17a:icml:2017} with a latent space $\Z=\R^2$ and $p(\b{z}) = \N(\b{0},~\Id_2)$. For the ambient metric we used a positive RBF (Eq.~\ref{eq:ambient_metric_density}). For implementation details see Appendix~\ref{appendix:section:experimental_details}.

%\begin{wrapfigure}{r}{0.225\textwidth}
%	\vspace{-18pt}
%	\begin{center}
%		\begin{subfigure}[b]{0.225\textwidth}
%			\includegraphics[width=\textwidth]{imgs/experiments/synthetic_gan/res_ours.png}
%		\end{subfigure}
%		\vspace{5pt}
%		\begin{subfigure}[b]{0.2\textwidth}
%			\includegraphics[width=\textwidth]{imgs/experiments/synthetic_gan/res_gmm.png}
%		\end{subfigure}
%		\vspace{-10pt}
%%		\caption[none]{\emph{Top}: The mixture of LANDs. \emph{Bottom}: The linear GMM.}
%	\caption[none]{}
%	\end{center}
%	\vspace{-15pt}
%\end{wrapfigure}

Then, we define a density function in $\Z$ using the learned Riemannian metric as $q(\b{z}) \propto \tilde{p}_r(\b{z})\cdot(1 + \sqrt{|{\b{M}}(\b{z})|})^{-1}$, where $\tilde{p}_r(\b{z})$ is a uniform density within a ball of radius $r$ \citep{lebanon:uai:2002, tam:icml:2015}. The behavior of $q(\b{z})$ is interpretable, since the density is high wherever  ${\b{M}}(\cdot)$ is small and that happens only in the regions of $\Z$ that $g(\cdot)$ learns to map near the given data manifold in $\X$. Also, the $\tilde{p}(\b{z})$ simply ensures that the support of $q(\b{z})$ is within the region where the $g(\cdot)$ is trained. Thus, we compare samples generated from $q(\b{z})$ using Markov Chain Monte Carlo (MCMC) and the prior $p(\b{z})$. We see in Fig.~\ref{experiments:gan_synthetic} that our samples align better with the data manifold, %Also, we approximated the negative log-likelihood using a surrogate GMM fitted on the given data and the results are 1.21 (data), 1.61 (ours), 3.35 (prior).
so we can think of $q(\b{z})$ as an approximate aggregated posterior without using an encoder. In a similar spirit, \cite{tanielian2020learning} proposed to reject samples from $p(\b{z})$ based only on the norm of generator’s Jacobian. Moreover, using the sampled latent points we fit a mixture of LANDs (Fig.~\ref{experiments:land_vs_gmm}), which are locally adaptive normal distributions on Riemannian manifolds \citep{arvanitidis:nips:2016}. Hence, we can sample from each component individually, without training a conditional GAN \citep{mirza:2014:arxiv}.

\vspace{-5pt}

\begin{figure}[h]
	\centering
	
	\begin{subfigure}[b]{0.32\textwidth}
		\includegraphics[width=\textwidth]{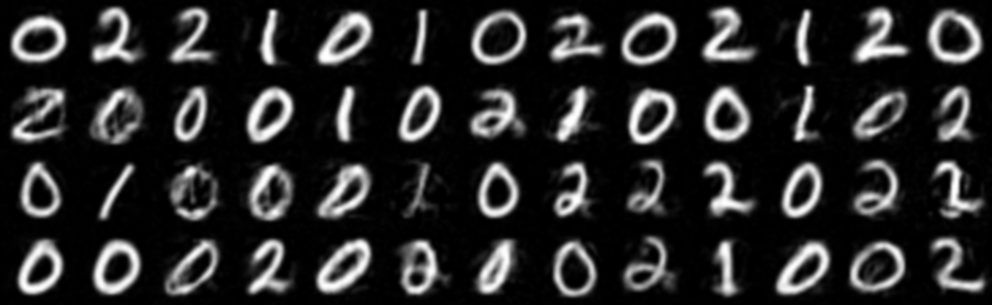}
	\end{subfigure}
	\begin{subfigure}[b]{0.32\textwidth}
		\includegraphics[width=\textwidth]{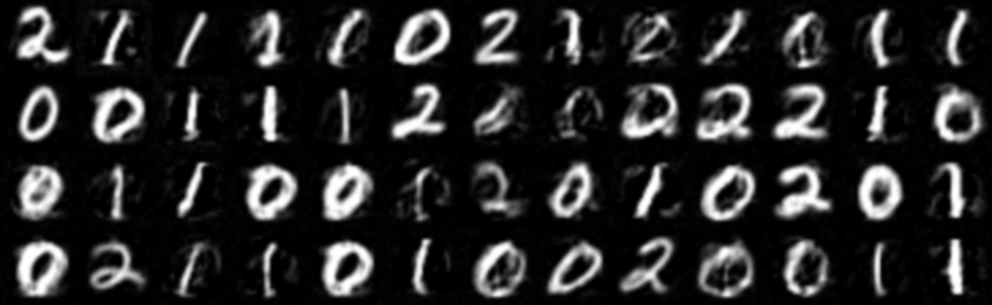}
	\end{subfigure}
	\begin{subfigure}[b]{0.32\textwidth}
		\includegraphics[width=\textwidth]{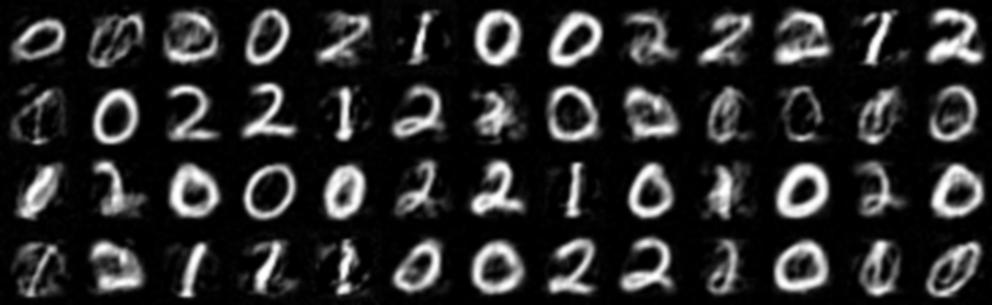}
	\end{subfigure}
	
	\vspace{5pt}
	
	\begin{subfigure}[b]{1\textwidth}
		\includegraphics[width=\textwidth]{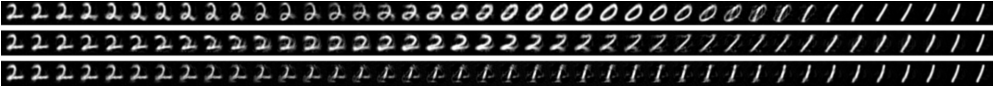}
	\end{subfigure}
	
	\caption{\emph{Top} panels, \emph{left} to \emph{right}: samples from $q(\b{z})$, from $q(\b{z})$ without the $\b{M}_{\X'}(\cdot)$ and from $p(\b{z})$. \emph{Bottom} panels, \emph{top} to \emph{bottom} row: interpolations under the proposed Riemannian metric, without the $\b{M}_{\X'}(\cdot)$ and the linear interpolation. We see that $\b{M}_{\X'}(\cdot)$ improves the sampling, and also, our path (top row) respects the manifold structure avoiding off-the-manifold ``shortcuts'' (middle row).}
	\label{experiments:gan_mnist_samples}
\end{figure}

\paragraph{MNIST data.} Similarly, we performed an experiment with the MNIST digits 0,1,2 and $\Z = \R^5$ and we show the results in Fig.~\ref{experiments:gan_mnist_samples}. Since this is a high dimensional dataset, the RBF used for the $\b{M}_\X(\cdot)$ will be a poor fit, and also, the Euclidean distance between images is not meaningful. Therefore, after the Wasserstein GAN training, we used PCA to project linearly the data in a $d'$-dimensional subspace $D>d' > d$ with $d'=10$, where we define the ambient $\b{M}_{\X'}(\cdot)$. This step removes the non informative dimensions from the data, while keeping the global structure of the data manifold unchanged. So the intrinsic geometry is approximately preserved. We discuss this linear projection step in Appendix~\ref{appendix:manifold_theory}, and provide further implementation details and results in Appendix~\ref{appendix:section:experimental_details}.

%%%%%%%%%%%%%%%%%%%%%%%%%
%%%%%%%% MNIST RESULT %%%%%%%% 
%%%%%%%%%%%%%%%%%%%%%%%%%

\vspace{-10pt}

\begin{wrapfigure}{r}{0.3\textwidth}
		\vspace{-10pt}
	\centering	
	
	\includegraphics[width=\linewidth]{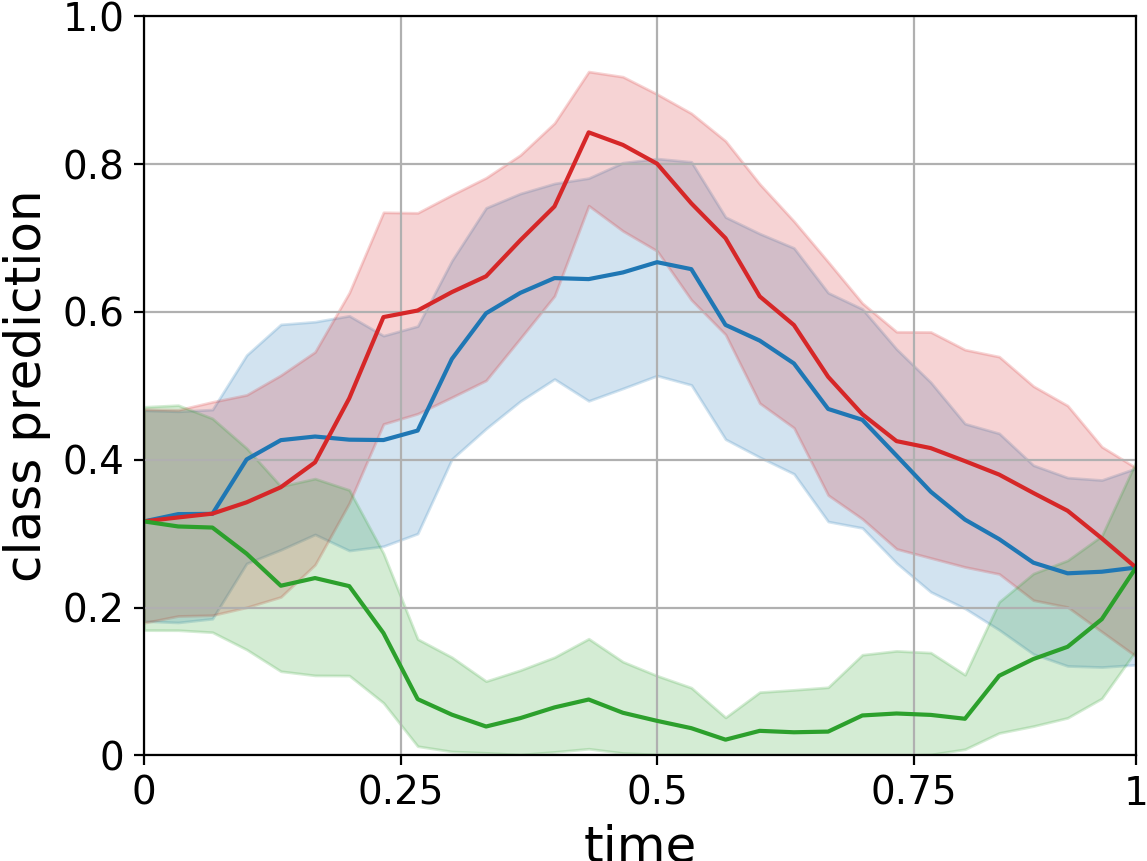}
	\vspace{-15pt}
	\caption[none]{}
	%	\vspace{-25pt}
	\label{experiments:pgan_classifier_result}
	\vspace{-10pt}
\end{wrapfigure}\paragraph{Pre-trained generator.} Finally, we use as $g(\cdot)$ a pre-trained Progressive GAN on the CelebA dataset \citep{karras2018progressive}, and we also train a classifier that distinguishes the blond people. As before, we use a linear projection to $d'=1000$, where we define a cost based ambient metric $\b{M}_{\X'}(\cdot)$ based on a positive RBF (Eq.~\ref{eq:ambient_metric_density}). This metric is designed to penalize regions in $\X$ that correspond to blonds and the goal is to avoid these regions when interpolating in $\Z$. As we discussed in Sec.~\ref{section:data_learned_riemannian_manifolds}, it is not guaranteed that this deterministic generator properly captures the structure of the data manifold in $\Z$, but even so, we test our ability to control the shortest paths.

\begin{figure}[t]
	\includegraphics[width=\textwidth]{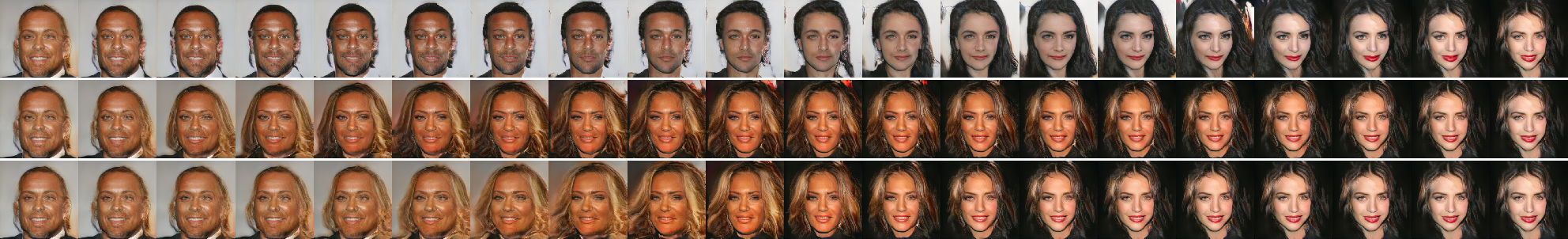}
	\caption[none]{\emph{Top} to \emph{bottom}: interpolations under the proposed Riemannian metric, without the $\b{M}_{\X'}(\cdot)$ and linear interpolation. Only our path successfully avoids the high cost regions (blond hair). This high-level semantic information is encoded into the ambient metric which only our path utilizes.}
	\label{experiments:pgan}
\end{figure}

As we observe in Fig.~\ref{experiments:pgan} only our path that utilizes the informative $\b{M}_{\X'}(\cdot)$ successfully avoids crossing regions with blond hair. In particular, it corresponds to the optimal path on the generated manifold, while taking into account the high-level semantic information. In contrast, the shortest path without the $\b{M}_{\X'}(\cdot)$ passes through the high cost region in $\X$, as it merely minimizes the distance on the manifold and does not utilize the additional information. Also, we show in Fig.~\ref{experiments:pgan_classifier_result} the classifier prediction along several interpolants. Clearly, we see that only our shortest paths ({\protect \begin{tikzpicture}[baseline={([yshift=-3pt]current bounding box.center)}]
{\draw [color={rgb,255:red,82; green,173; blue,58}, line width=1.5, yshift=0cm] (0,0) -- (.3,0);}
\end{tikzpicture}}) respect the ambient geometry, while both the straight line ({\protect \begin{tikzpicture}[baseline={([yshift=-3pt]current bounding box.center)}]
{\draw [color={rgb,255:red,69; green,152; blue,230}, line width=1.5, yshift=0cm] (0,0) -- (.3,0);}
\end{tikzpicture}}) and the naive shortest path ({\protect \begin{tikzpicture}[baseline={([yshift=-3pt]current bounding box.center)}]
{\draw [color={rgb,255:red,220; green,79; blue,53}, line width=1.5, yshift=0cm] (0,0) -- (.3,0);}
\end{tikzpicture}}), interpolate through regions classified as blond. For further details and interpolation results see Appendix~\ref{appendix:section:experimental_details}.

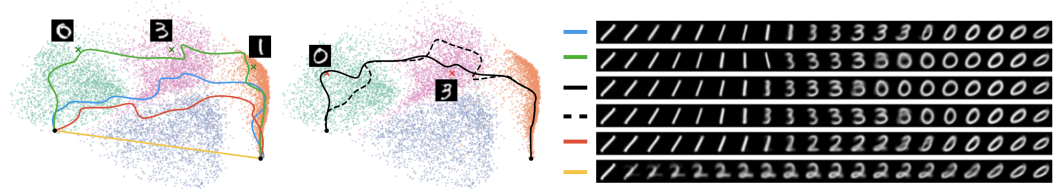
\begin{figure}[t]
	\centering
	\hspace{-25pt}
	\begin{subfigure}[b]{0.25\textwidth}
		{\includegraphics[width=\textwidth]{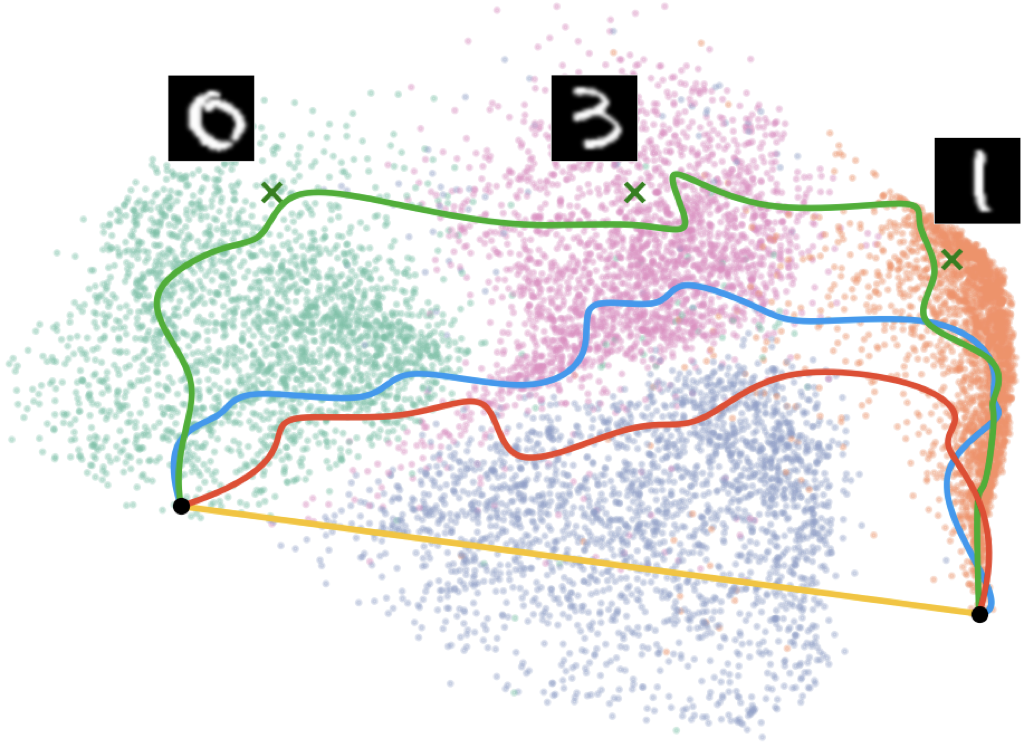}}
		%		\caption{AE}
		%		\label{fig:AE}
	\end{subfigure}
	~
	\hspace{-7pt}
	\begin{subfigure}[b]{0.25\textwidth}
		{\includegraphics[width=\textwidth]{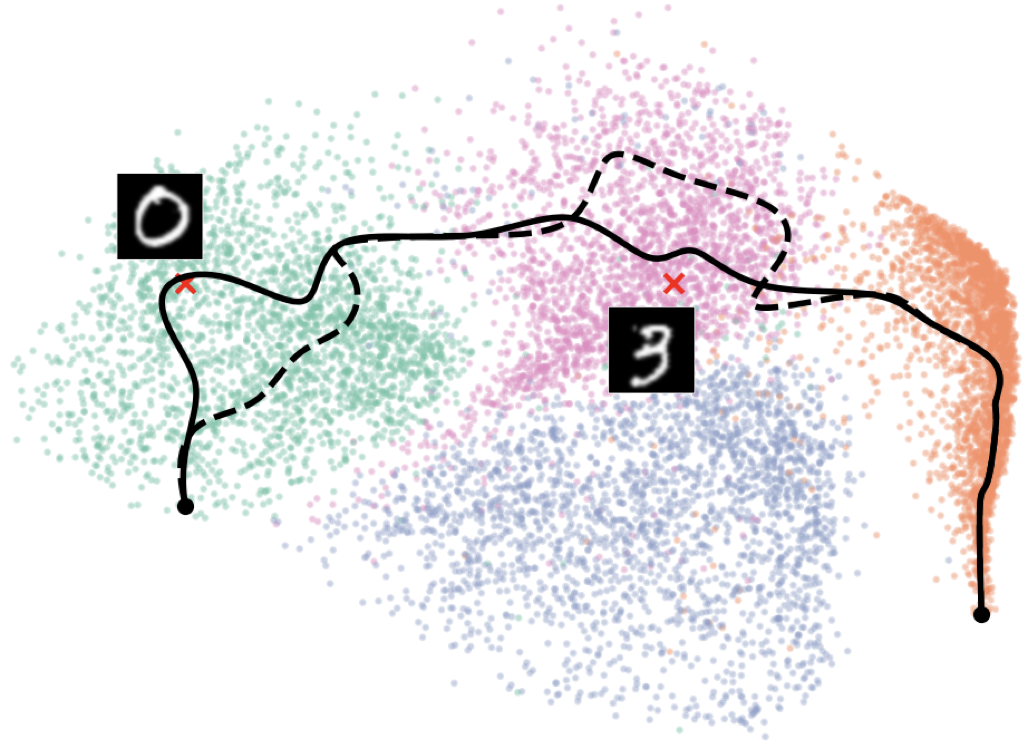}}
		%		\caption{AE}
		%		\label{fig:AE}
	\end{subfigure}
	~
	\begin{subfigure}[b]{0.45\textwidth}
		%		{\includegraphics[width=\textwidth]{imgs/experiments/mnist_vae/lda_pca_cost/temp_img_3.png}}
		%		\caption{AE}
		%		\label{fig:AE}
		\begin{tikzpicture}
		\node[anchor=south west,inner sep=0] (image) at (0,0) {\includegraphics[width=\textwidth]{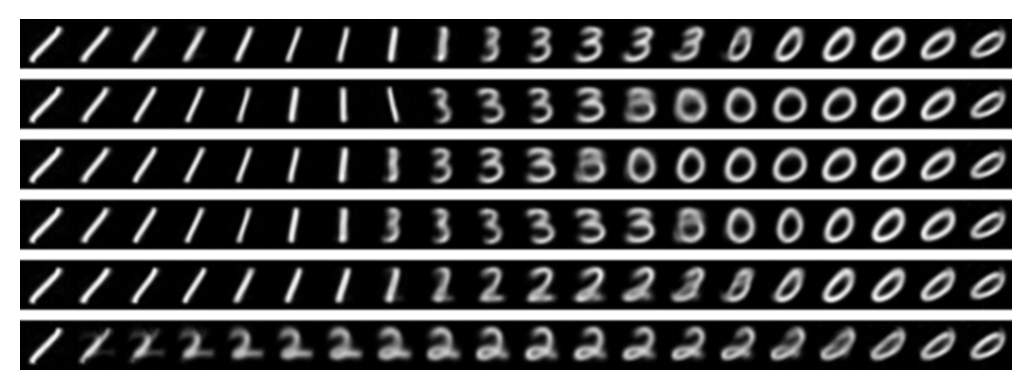}};
		\begin{scope}[x={(image.south east)},y={(image.north west)}]
		{\draw [color={rgb,255:red,69; green,152; blue,230}, line width=1.5, yshift=0cm] (-0.05, 0.89) -- (0, 0.89);}
		{\draw [color={rgb,255:red,82; green,173; blue,58}, line width=1.5, yshift=0cm] (-0.05, 0.75) -- (0, 0.75);}
		{\draw [black, line width=1.5, yshift=0cm] (-0.05, 0.58) -- (0, 0.58);}
		{\draw [black, dashed, line width=1.5, yshift=0cm] (-0.05, 0.42) -- (0, 0.42);}			
		{\draw [color={rgb,255:red,220; green,79; blue,53}, line width=1.5, yshift=0cm] (-0.05, 0.28) -- (0, 0.28);}
		{\draw [color={rgb,255:red,241; green,196; blue,67}, line width=1.5, yshift=0cm] (-0.05, 0.11) -- (0, 0.11);}
		%			\draw[help lines,xstep=.1,ystep=.1] (0,0) grid (1,1);
		%			\foreach \x in {0,1,...,9} { \node [anchor=north] at (\x/10,0) {0.\x}; }
		%			\foreach \y in {0,1,...,9} { \node [anchor=east] at (0,\y/10) {0.\y}; }
		\end{scope}
		\end{tikzpicture}
	\end{subfigure}
	\hspace{-20pt}
	
	\caption[none]{Comparing interpolants under different ambient metrics and their generated images. We can control effectively the shortest paths to follow high-level information incorporated in $\b{M}_\X(\cdot)$.}
	\label{experiments:vae_mnist_interpolations}
\end{figure}

\subsection{Demonstrations with Stochastic Generators}

\paragraph{Controlling Shortest Paths.} We compare in Fig.~\ref{experiments:vae_mnist_interpolations} the effect of several interpretable ambient metrics on the shortest paths in the latent space of a VAE trained with $\Z=\R^2$ on the MNIST digits 0,1,2,3. As before, we project linearly the data in $d'=10$ to construct there the ambient metrics.  At first, we observe that under the Euclidean metric in $\X$ the path ({\protect \begin{tikzpicture}[baseline={([yshift=-3pt]current bounding box.center)}]
{\draw [color={rgb,255:red,220; green,79; blue,53}, line width=1.5, yshift=0cm] (0,0) -- (.3,0);}
\end{tikzpicture}}) merely follows the structure of the generated $\M_\Z$ since it avoids regions with no data. Then, we construct an LDA metric in $\X'$ (Eq.~\ref{eq:convex_combination_metric}) by considering the digits 0,1,3 to be in the same class. Hence, the resulting path ({\protect \begin{tikzpicture}[baseline={([yshift=-3pt]current bounding box.center)}]
{\draw [color={rgb,255:red,69; green,152; blue,230}, line width=1.5, yshift=0cm] (0,0) -- (.3,0);}
\end{tikzpicture}}) avoids crossing the regions in $\Z$ that correspond to digit 2, while simultaneously respects the geometry of $\M_\Z$. Also, we select 3 data points ($\color[RGB]{82,173,58}\boldsymbol{\times}$) and using their 100 nearest neighbors in $\X'$ we construct the local covariance based metric (Eq.~\ref{eq:local_pca_metric}), such that the path ({\protect \begin{tikzpicture}[baseline={([yshift=-3pt]current bounding box.center)}]
{\draw [color={rgb,255:red,82; green,173; blue,58}, line width=1.5, yshift=0cm] (0,0) -- (.3,0);}
\end{tikzpicture}}) to move closer to the selected points.

\begin{wrapfigure}{r}{0.3\textwidth}
	\vspace{-20pt}
	
	\begin{tikzpicture}
	\draw (0, 0) node[inner sep=0] {\includegraphics[width=\linewidth]{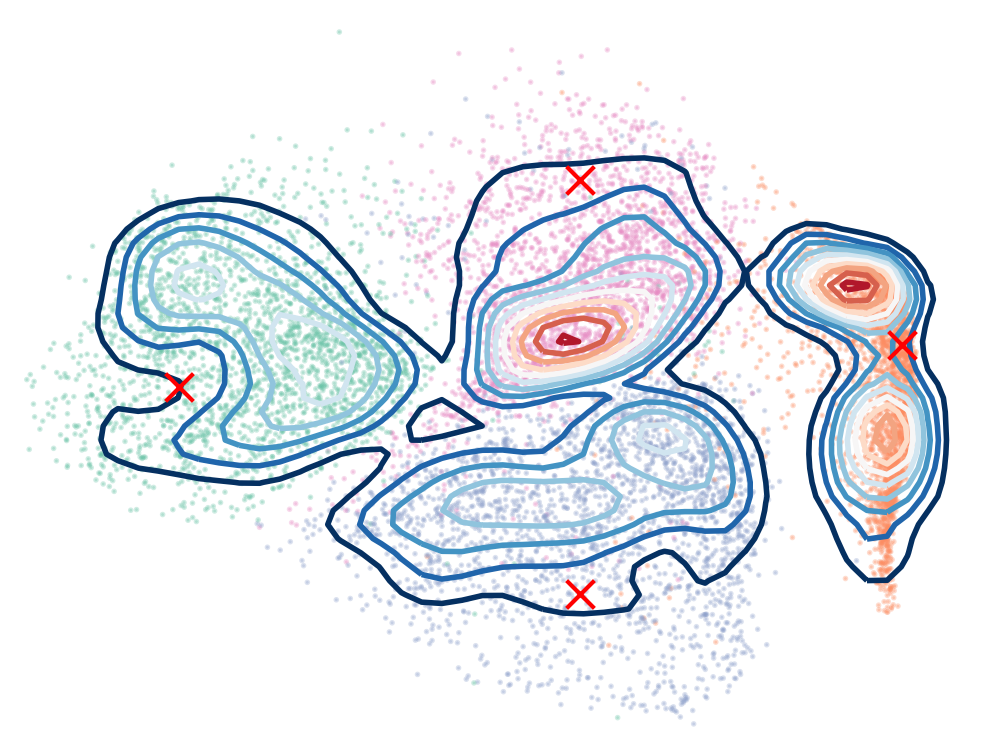}};
	\draw (-1.2, 1.1) node {\tiny Shortest path};
	\end{tikzpicture}
	\vspace{-5pt}
	\begin{tikzpicture}
	\draw (0, 0) node[inner sep=0] {\includegraphics[width=\linewidth]{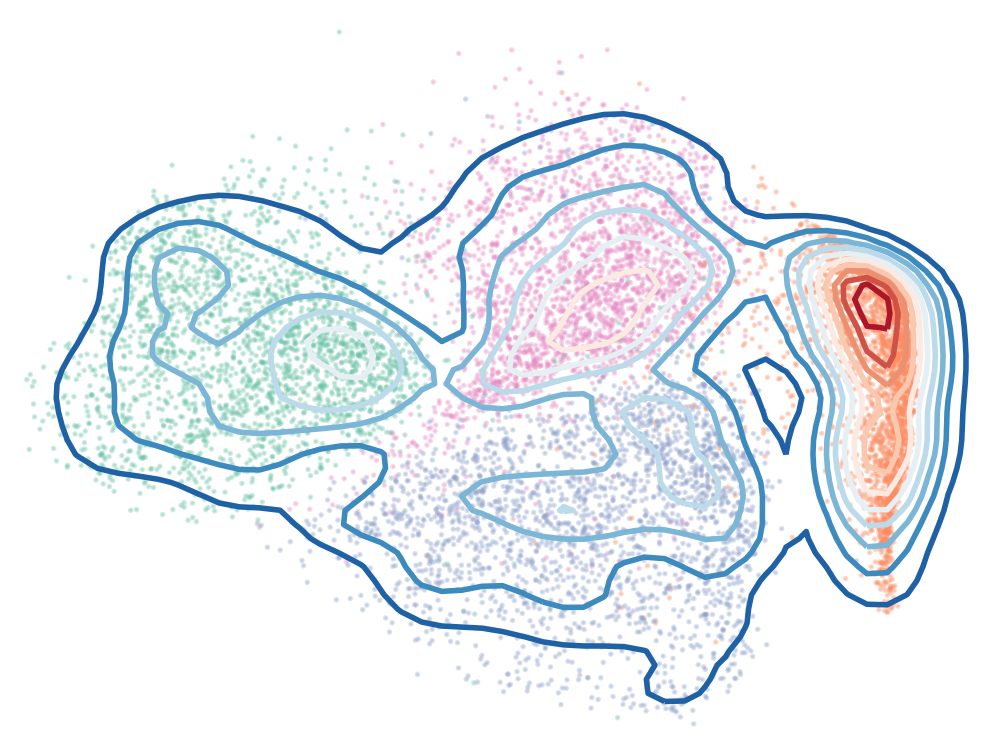}};
	\draw (-1.1, 1.1) node {\tiny Linear distance};
	\end{tikzpicture}
	
	\caption[none]{}	
	%	\caption[none]{\emph{Top}: Classifier result in PGAN. \emph{Bottom}: KDE plots in VAE.}
	\label{experiments:vae_mnist_kde}
		\vspace{-40pt}
\end{wrapfigure}

Moreover, we linearly combine these two metrics, such that to enforce the path ({\protect \begin{tikzpicture}[baseline={([yshift=-3pt]current bounding box.center)}]
{\draw [color=black, line width=1.5, yshift=0cm] (0,0) -- (.3,0);}
\end{tikzpicture}}) to pass through 0,1,3 while moving closer to the selected points ($\color[RGB]{82,173,58}\boldsymbol{\times}$). Finally, we include to this linear combination a cost related metric (Eq.~\ref{eq:ambient_metric_density}) based on a positive RBF that increases near the points ($\color[RGB]{220,79,53}\boldsymbol{\times}$) in $\X'$. Therefore, the resulting path ({\protect \begin{tikzpicture}[baseline={([yshift=-3pt]current bounding box.center)}]
{\draw [color=black, dashed, line width=1.5, yshift=0cm] (0,0) -- (.3,0);}
\end{tikzpicture}}) avoids these neighborhoods, while respecting the other ambient metrics and the geometry of $\M_\Z$. Hence, we can effectively control the shortest paths by designing and combining the ambient metrics accordingly.

\paragraph{Kernel Density estimation.} Finally, we show in Fig.~\ref{experiments:vae_mnist_kde} the kernel density estimation in $\Z$ comparing the straight line to the shortest path. For $\b{M}_{\X'}(\cdot)$ we linearly combine an LDA metric where each digit is a separate class and a cost based metric that increases near the points ($\color[RGB]{220,79,53}\boldsymbol{\times}$) in $\X'$. We see that $\b{M}(\cdot)$ helps to distinguish better the classes due to the LDA metric, while the density is reduced near the regions with high cost. For further discussion see Appendix~\ref{appendix:section:experimental_details}.

%% file: conclusion.tex
%!TEX root = ./paper.tex

\section{Conclusion}
\label{section:conclusion}

% THIS IS REMOVED
%In this work, we showed how the geometry of the ambient space can be captured into the latent space of a generative model. The geometry could potentially provide high level information regarding the problem of interest, which can be used in the representation space. As we discussed the learning of the Riemannian metric in the ambient space is totally problem related and we demonstrated some basic approaches. Moreover, we proposed a simple modeling for deterministic generators to extrapolate meaningfully outside of the trained regions of $\Z$. In the future, it will be interesting to see if and how we can use the ambient space geometry during the learning phase of the generative model.

%We showed how to construct Riemannian metrics in order to equip the ambient space of a generative model, which allows to encode high-level domain knowledge. Then, we analyzed the requirements to bring this information into the latent space. In particular, proper uncertainty estimation is essential in stochastic generators, while in the deterministic case one way is based on the proposed meaningful extrapolation. Thus, we get interpretable shortest paths that respect the ambient space geometry, while moving optimally on the learned manifold. In the future, it will be interesting to see if and how we can use the ambient space geometry during the learning phase of the generative model.

We considered the ambient space of generative models as a Riemannian manifold. This allows for encoding domain knowledge through the metric and we proposed an easy way to construct suitable metrics. In order to capture the geometry into the latent space, proper uncertainty estimation is essential in stochastic generators, while in the deterministic case one way is through the proposed meaningful extrapolation. Thus, we get interpretable shortest paths that respect the ambient space geometry, while moving optimally on the learned manifold. In the future, it will be interesting to see if and how we can use the ambient space geometry during the learning phase of the generative model.

%% file: appendix.tex
%!TEX root = ./paper.tex

\section*{Appendix}

\section{Further information on Riemannian geometry}
\label{appendix:sec:geometry_theory}

\setcounter{equation}{10}
\setcounter{figure}{10}

Let us assume a $d$-dimensional smooth manifold $\M$ embedded in an ambient space $\X=\R^D$ with $d<D$, where it is defined a Riemannian metric $\b{M}_{\X}: \X \rightarrow \R^{D\times D}_{\succ 0}$. Therefore, the space $\X$ is a Riemannian manifold, since $\X$ is a smooth manifold. This directly implies that the simple Euclidean space is a Riemannian manifold as well. Due to the embedding of $\M$ a Riemannian metric is induced in the tangent space $\tangent{\b{x}}{\M}$ by the restriction of the Riemannian ambient metric $\b{M}_{\X}(\cdot)$, even for the simple case $\b{M}_\X(\b{x}) = \Id_D$. For simplicity, we further assume that a global chart map $\phi(\M)$ exists.

Generally, one of the main utilities of a Riemannian manifold $\M\subseteq\X$ is to enable us compute shortest paths therein. Intuitively, the norm $\sqrt{\inner{d\b{x}}{d\b{x}}_\b{x}}$ represents how the infinitesimal displacement vector $d\b{x} \approx \b{x}' - \b{x}$ on $\M$ is locally scaled. Thus, for a curve $\gamma:[0,1]\rightarrow \M$ that connects two points $\b{x}=
\gamma(0)$ and $\b{y}=\gamma(1)$, the length on $\M$ or equivalently in $\phi(\M)$ using Eq.~2 is measured as
\begin{align}
\label{appendix:eq:curve_length}
\length{\gamma(t)} = \int_0^1 \sqrt{\inner{\dot{\gamma}(t)}{\dot{\gamma}(t)}_{\b{\gamma}(t)}} dt = \int_0^1\sqrt{\inner{\dot{c}(t)}{\b{M}(c(t)) \dot{c}(t)} }dt = \length{c(t)}
\end{align}
where $\dot{\gamma}(t)~=~\partial_t \gamma(t) \in \mathcal{T}_{\gamma(t)}\M$ is the velocity of the curve and accordingly $\dot{c}(t) \in \mathcal{T}_{c(t)}\phi(\M)$. The length is an invariant quantity under reparametrization i.e., for any continuous monotonic function $s:[a,b]\rightarrow[0,1]$ the curve $\tilde{\gamma}(t')=\gamma(s(t')),~t'\in[a,b]$  has the same length. Instead, in order to find the \emph{shortest path} we minimize the corresponding energy functional, which is a non-invariant quantity, \begin{align}
\gamma^*(t)~=~\argmin_{\gamma(t)}~ E[\gamma(t)]~=~\argmin_{\gamma(t)}~ \frac{1}{2}\int_0^1\inner{\dot{\gamma}(t)}{\dot{\gamma}(t)}_{\gamma(t)} dt,
\end{align} and similarly the energy can be written for the $c(t)\in\phi(\M)$ in the intrinsic coordinates as Eq.~\ref{appendix:eq:curve_length}. The minimizers of this energy have constant speed $\norm{\dot{\gamma}(t)}_{\gamma(t)}$and are known as \emph{geodesics}.

In theory, instead of solving the problem directly on $\M$, we utilize the intrinsic coordinates. Assuming the global chart $\phi(\cdot)$, we search for a curve $c(t)=\phi(\gamma(t)) \in \phi(\M)$ that minimizes the corresponding energy functional $E[c(t)]=\frac{1}{2}\int_0^1 \inner{\dot{c}(t)}{{\b{M}}(c(t))\dot{c}(t)} dt$. Here, we used the fact that $\gamma(t) = \phi^{-1}(c(t)) \Rightarrow \dot{\gamma}(t) = \b{J}_{\phi^{-1}}({c}(t)) \dot{c}(t)$, since by the definition of a smooth manifold $\phi(\cdot),~\phi^{-1}(\cdot)$ exist and are smooth maps.  Now, we can find the minimizers by directly applying the Euler-Lagrange equations to the energy $E[c(t)]$, which results to a system of 2\textsuperscript{nd} order non-linear ordinary differential equations (ODEs) written as in \cite{arvanitidis:iclr:2018}
\begin{align}
\label{eq:ode}
\ddot{c}(t) 
=-\frac{1}{2}\b{M}^{-1}(c(t))\Big[2 (\dot{c}(t)^\T \otimes \Id_d) \parder{\vectorize{\b{M}(c(t))}}{c(t)}\dot{c}(t) 
- \parder{\vectorize{\b{M}(c(t))}}{c(t)}^\T (\dot{c}(t) \otimes \dot{c}(t))\Big],
\end{align}
where $\vectorize{\cdot}$ stacks the columns of a matrix into a vector and
$\otimes$ is the Kronecker product. This is solved as a boundary value problem (BVP) with boundary conditions $c(0)=\b{x}$ and $c(1)=\b{y}$. Note that this ODEs system is a standard result in differential geometry, and intuitively, the resulting shortest paths tend to avoid areas where the metric magnitude $\sqrt{\abs{\b{M}(c(t))}}$ is high.

In order to perform computations on a Riemannian manifold we need to define two operations analogous to the ``plus'' and ``minus'' of the Euclidean space. First, the \emph{exponential map} is an operator $\Expmap{\b{x}}{\b{v}t}=\gamma(t)$ that takes two inputs, a point $\b{x}\in\M$ and a $\b{v}\in\tangent{\b{x}}{\M}$, and generates a geodesic with $\gamma(1)=\b{y}\in\M$ and initial velocity $\dot{\gamma}(0)=\b{v}$. The inverse operator is called the \emph{logarithmic map} $\Logmap{\b{x}}{\b{y}}=\b{v}$ that takes two inputs $\b{x},\b{y}\in\M$ to return the tangent vector $\b{v}\in\tangent{\b{x}}{\M}$. Note that these two operators are dual in a small neighborhood around $\b{x}\in\M$. Moreover, the logarithmic map provides \emph{coordinates} for the points in a neighborhood on $\M$ with respect to the base point $\b{x}$, but only the distances between the center $\b{x}$ and the points are meaningful and not the ones between the points. Also, by definition the $\inner{\Logmap{\b{x}}{\b{y}}}{\b{M}_{\X}(\b{x})\Logmap{\b{x}}{\b{y}}} = \text{dist}^2(\b{x},\b{y}) = \text{length}^2[\gamma(t)]$, but we can rescale the logarithmic map such that the $\text{dist}^2(\b{x},\b{y}) = \inner{\overline{\text{Log}}_{\b{x}}(\b{y})}{\overline{\text{Log}}_{\b{x}}(\b{y})}$. The rescaled coordinates $\overline{\text{Log}}_{\b{x}}(\b{y})$ are known as \emph{normal coordinates} and are the ones that we use in practice.

\clearpage

\section{Theoretical analysis of the generator}
\label{appendix:manifold_theory}

In this section we analyze the properties that the proposed generator $g(\cdot)$ should have. In particular, in order to have a theoretically sound model the generator has to be at least twice differentiable, and additionally, an immersion. Also, we need a specific behavior such that to properly capture  the structure of the data manifold in the latent space, both in the stochastic and deterministic generator case. Of course, the basic assumption is that the data lie \emph{near} an \emph{embedded} smooth manifold $\M$ in the ambient space $\X$. Intuitively, an embedded $d$-dimensional manifold can be considered as a surface that is everywhere homeomoerphich to a $d$-dimensional Euclidean space, which implies that contractions and intersections are now allowed. In contrast, an \emph{immersion} is a relative simpler condition, since intersections are allowed but again no contractions. In theory, the generator has to be at least an immersion such that to pull-back the Riemannian metric of the manifold.

\paragraph{Stochastic generator.} We consider as generator the function $g(\b{z}) = \mu(\b{z}) + \text{diag}(\varepsilon)\cdot \sigma(\b{z})$, where $\varepsilon\sim\N(\b{0},~\Id_D)$ and $\mu:\Z\rightarrow\X$ is a DNN and $\sigma:\Z\rightarrow \R^{\text{dim}(\X)}_{> 0}$ is based on a positive RBF. Note that in principle we model the precision $\beta(\b{z}) = (\sigma^2(\b{z}))^{-1}$ with the positive RBF, so the $\sigma(\b{z}) = \beta^{-\nicefrac{1}{2}}(\b{z})$.
From the theory we know that $g(\cdot)$ has to be smooth. At first, we can achieve smoothness easily for $\mu(\cdot)$ and $\sigma(\cdot)$. In particular, $\sigma(\cdot)$ is smooth as a linear combination of smooth functions. For the DNN $\mu(\cdot)$ we can use smooth activation functions as the $\texttt{tanh}(\cdot),~\texttt{softplus}(\cdot),~\text{etc}$. But the stochasticity of $\varepsilon$ makes $g(\cdot)$ non-smooth, and hence, non differentiable with respect to $\b{z}$. Instead, if $\varepsilon$ is fixed $\forall \b{z}\in\Z$ denoted as $g_\varepsilon(\cdot)$, then this is a smooth nonlinear map, and consequently, differentiable. A different perspective on the smoothness of $g(\cdot)$ has been given by \cite{eklund:arxiv:2019}. There it is shown that $g_\varepsilon(\cdot)$ is actually the random projection of the deterministic smooth nonlinear map $\b{z} \mapsto [\mu(\b{z}),~\sigma(\b{z}) ]$ under the random projection matrix $\b{P}_\varepsilon=[\Id_D,~\text{diag}(\varepsilon)]$. In both views, fixing $\varepsilon$ implies that the \emph{sampled} $\M_\varepsilon = g_\varepsilon(\Z)$ is a smooth immersed manifold in $\X$. Obviously, the $\E_\varepsilon[\M_\varepsilon] = \E_\varepsilon[g_\varepsilon(\Z)] = \E_{\varepsilon}[g(\Z)] = \mu(\Z)$, which shows that the expected manifold, as well as the likelihood of the individual points $p(\b{x}|\b{z})$ do not change.

The $g_\varepsilon(\cdot)$ is an immersion if  $\b{J}_{g_\varepsilon}(\b{z}) = \b{J}_\mu(\b{z}) + \b{J}_\sigma(\b{z}) \varepsilon$ has full rank $\forall~\b{z}\in\Z$. For $\mu(\cdot)$ (DNN) this can be true within the support of $p(\b{z})$ where the activation functions usually do not reach their limit behavior. For instance, with $\texttt{tanh}(\cdot)$ as activation, we expect within the support of $p(\b{z})$ the hidden units output to not be constant $\pm 1$. In addition, we need each hidden layer to have greater or equal number of units to the previous layer while all the weight matrices are full rank. While for $\sigma(\cdot)$ (inverse positive RBF) at least $\text{dim}(\Z)$ basis functions has to be active and the weight matrix has to be full rank. The conventions above define an immersed manifold $\M_\varepsilon = g_\varepsilon(\Z)$ in $\X$, since we avoid \emph{contractions}. Of course, the two matrices $\b{J}_\mu(\b{z}), \b{J}_\sigma(\b{z})$ should not cancel any of their columns.

\paragraph{Generator with linear extrapolation.} We analyze the behavior of the proposed architecture $g(\b{z}) = f(\b{z}) + \b{U}\cdot \text{diag}([\sqrt{\lambda_1},\dots, \sqrt{\lambda_d}])\cdot \b{z} + \b{b}$ where $f:\Z \rightarrow \X$ is a nonlinear map and $\Z=\R^d$, $\X=\R^D$ with $D\gg d$. Note that for the stochastic generator case and for fixed $\varepsilon$ the function $f_\varepsilon(\b{z}) = \mu(\b{z}) + \text{diag}(\varepsilon)\cdot\sigma(\b{z})$ can be simply seen as the addition of two nonlinear functions (see above). The linear map is constructed using the top $d$-eigenvectors scaled by their eigenvalues, coming from the eigen-decomposition of the data empirical covariance matrix. More specifically, the empirical data covariance $\b{C} = \frac{1}{N-1}\sum_{n=1}^N (\b{x}_n - \b{b}) (\b{x}_n - \b{b})^\T$, where $\b{b} = \frac{1}{N} \sum_{n=1}^N \b{x}_n$, which can be decomposed as $\b{C} = \b{V}\bs{\Lambda}\b{V}^\T$. We use for $\b{U}$ the first $d$ columns of $\b{V}\in\R^{D\times D}$ and the corresponding eigenvalues.  In particular, we check if and when $g(\cdot)$ satisfies the properties:

\paragraph{Smoothness.} We need the generator to be sufficiently smooth, which means in our case at least twice differentiable. This condition can be easily satisfied by selecting the activation functions accordingly as $\texttt{tanh}(\cdot), \texttt{softplus}(\cdot)$, etc. In practice, this is necessary since in the geodesic ODEs system we need to compute the derivative of the metric tensor, which in our case is implemented by first  taking the derivative of the Jacobian $\b{J}_{g}(\cdot)$. Obviously, by including in $g(\cdot)$ the linear map $\b{A}= \b{U}\cdot \text{diag}([\sqrt{\lambda_1},\dots, \sqrt{\lambda_d}])$ and $\b{b}$, the smoothness property will not change.
	
\paragraph{Immersion.} In theory a mapping  $g:\R^d \rightarrow \R^D$ with $D\gg d$ is an immersion if the corresponding $\b{J}_g(\b{z})\in\R^{D\times d}$ is everywhere injective or in other words full rank. In our case, the Jacobian includes a neural network and for an example we consider the simple function $f(\b{z}) = \b{W}_1\cdot \psi(\b{W}_0 \b{z} + \b{b}_0) +\b{b}_1$ with $\psi(\cdot)$ the activation function, and thus, the Jacobian is $\b{J}_f(\b{z}) = \b{W}_1\cdot {\psi'}(\b{W}_0 \b{z} + \b{b}_0)\odot \b{W}_0$. In order to be this quantity an immersion, first we need each hidden layer to have more or equal number of hidden units from the previous layer and the weight matrices to have full rank. Additionally, since the derivative of the activation functions $\psi'(\cdot)$ appears, we need this to be non-zero. Otherwise, this will directly affect the total rank of the Jacobian, because it will reduce the rank of the corresponding weight matrix. We \emph{conjecture} that for generative models which are trained using a compact support prior $p(\b{z})$ like the Gaussian, the trained model uses the activation functions $\psi(\cdot)$ closer to the center of their domain, where their corresponding derivative $\psi'(\b{z})$ is not zero, and not towards the domain limits. This basically implies that the corresponding hidden unit is active and is used by the model.
	
However, the Jacobian in our case is $\b{J}_g(\b{z}) = \b{J}_f(\b{z}) + \b{A}$, which means that in theory there are cases where the two matrices could cancel some of their columns. This will directly break the full rank condition, and thus, the mapping at this point will not be an immersion. Practically, this means that the corresponding Riemannian metric tensor in $\Z$, computed as the $\b{J}_g(\b{z})^\T  \b{J}_g(\b{z})$, will be degenerate since it will not have full rank. However, even if in theory this is a case that could happen, in practice, we \emph{conjecture} that this is a relatively unrealistic scenario. Instead, if the $\b{J}_f(\b{z})$ has low rank the linear part $\b{A}$ could even fix the problem, of course, if any of the rest columns do not cancel each other.

\begin{figure}[t]
	\centering
	\begin{subfigure}[b]{0.245\textwidth}
		\includegraphics[width=\textwidth]{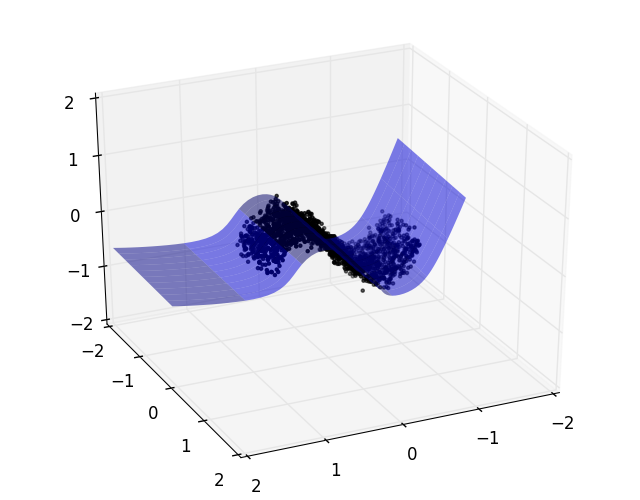}
	\end{subfigure}
	\begin{subfigure}[b]{0.245\textwidth}
		\includegraphics[width=\textwidth]{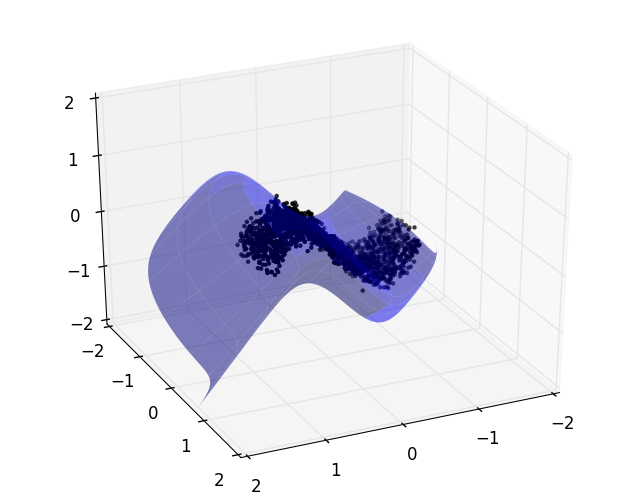}
	\end{subfigure}
	\begin{subfigure}[b]{0.245\textwidth}
		\includegraphics[width=\textwidth]{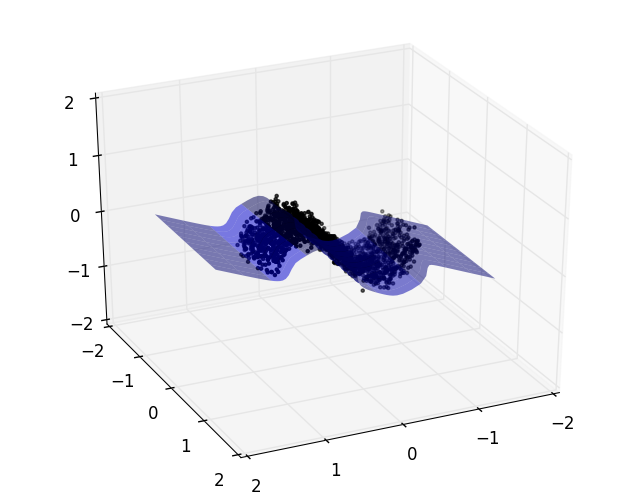}
	\end{subfigure}
	\begin{subfigure}[b]{0.245\textwidth}
		\includegraphics[width=\textwidth]{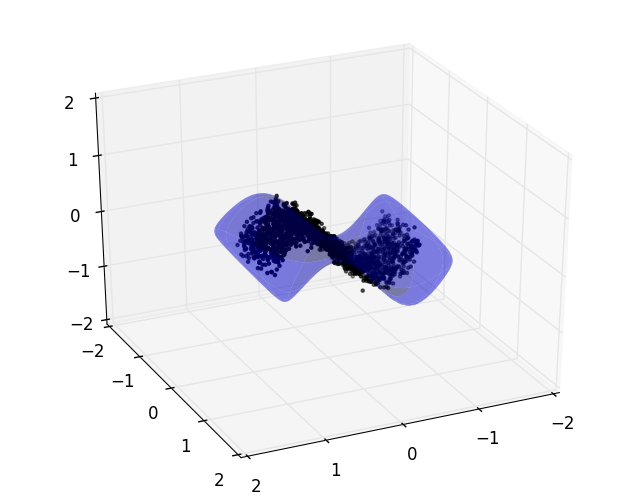}
	\end{subfigure}
	\caption{We trained an MLP with 2 hidden layers from $g:\R^2\rightarrow\R^3$, with hidden layer sizes 2, 3. From \emph{left} to \emph{right}: The extrapolation result with activation function $\texttt{softplus}(\cdot)$ and including the proposed linear map, without the linear map, with activation function $\texttt{tanh}(\cdot)$ and including the proposed linear map, without the linear map. We see that the linear map  improves the extrapolation of $g(\cdot)$, and the change is faster along the direction with the largest eigenvalue.}
	\label{appendix:fig:compare_extrapolations}
\end{figure}

\paragraph{Extrapolation.} The proposed meaningful extrapolation for a deterministic generator is one way to properly capture in $\Z$ the structure or topology of the data manifold, and thus, the geometry of the ambient space. Especially, this is necessary in the case where the behavior of the ambient metric is small only close to the data, which pulls the shortest paths towards the data manifold. Similarly, in the stochastic generator case meaningful uncertainty quantification is utilized in order to properly capture in $\Z$ the structure of the data manifold or in some sense its topology \citep{hauberg:only:2018}.

Thus, let us consider the deterministic generator case where $g(\b{z})$ is simply a neural network $f(\b{z})$ and let us pick a direction $\b{z}$ so that we move on the line $t\b{z}$ for $t\in\R$. When the $\texttt{tanh}(\cdot)$ activation function is used, as we move further from the support of the prior, the units of the first hidden layer will tend to output always a constant value $\rightarrow +1$ or $\rightarrow -1$. This means that the extrapolation will not be meaningful since it is gonna be always a constant. Similarly, for the $\texttt{softplus}(\cdot)$ as we move to the boundaries of the domain of $t$, the output of the activation will be either a constant $\rightarrow 0$ or a linear function. However, for each output dimension $f_j(\b{z}t)$ if the $t\rightarrow +\infty$ corresponds to a linear extrapolation the $t\rightarrow -\infty$ will extrapolate to zero. Therefore, in the $\texttt{softplus}(\cdot)$ case, even if the generator will potentially extrapolate meaningfully in some parts, in general, the behavior is arbitrary and hard to interpret $\forall~\b{z}\in\Z$. We show the behavior on a synthetic example in Fig.~\ref{appendix:fig:compare_extrapolations}.

So including the linear map $\b{A}, \b{b}$ could potentially fix the extrapolation issue, since the map $g(\cdot)$ after some threshold $t$ becomes solely linear. However, as regards the immersion condition, when $t\rightarrow \pm \infty$ if all the dimension $f_j(\b{z}t)$ cancel out the corresponding rows of $\b{A}\b{z}t + \b{b}$, then the $g(t\b{z})$ output will be a constant value. However, we argue again that this is quite unrealistic to happen on the same time for all the output $D$ dimensions.

Above, we only describe the theoretical conditions and the properties that a generator has to respect. Nevertheless, proper guarantees and analysis should be provided in the future.

\paragraph{Linear projection of the ambient space.}Additionally, we discuss the case where we linearly project the data manifold in $\X' = \R^{d'}$, a lower dimensional space $D>d'>d$, and we learn the ambient metric $\b{M}_{\X'}(\cdot)$ therein. Intuitively, instead of finding the shortest path $\gamma(t)$ on the $\M\subset \X$ we find the path on the projected manifold in $\X'$ and we expect that the actual structure of $\M$ is preserved. The reason of this step is to remove the non very informative extra dimensions from the data e.g. high frequency context, which do not provide any significant information regarding the structure of that data manifold or simply if they just correspond to noise. In other words, this step helps us to reduce the dimensionality, such that to construct the ``ambient'' metric using the projected data. Of course, this is only acceptable if the linear projection does not change the structure of the data manifold, for instance by introducing self intersections or contractions. Note that still the generator is trained between the space $\Z$ and $\X$, prior to the linear projection, so $g(\cdot)$ is still able to capture the high frequency context of the given data. 

The practical reason for this step is that for high dimensional data e.g. images, due the curse of dimensionality \citep{Bishop:2006:PRM},  we need to reduce the dimension of $\X$, especially when the learned ambient metric is based on pairwise Euclidean distances. Also, we know that, even locally, Euclidean distance makes not too much sense for images. Hence, the linear projection to a lower dimensional space $\X'$ helps us to ensure that at least locally straight lines will be more meaningful.

Therefore, the linear projection of the data, helps us to learn easier an ``ambient'' Riemannian metric that provides information regarding the structure of the actual data manifold. However, we note again that it is necessary this step to not change the structure of the data manifold. Thus, the Riemannian metric $\b{M}_{\X'}(\cdot)$ that is learned from the projected data is defined in $\R^{d'}$, and hence, the pull-back in the latent space takes the following form
\begin{align}
\inner{{\b{v}'}}{\b{M}_{\X'}({\b{x}'}){\b{v}'}} = \inner{\b{P} \overline{\b{v}}}{\b{M}_{\X'}(\b{P}(\overline{\b{x}} - \overline{\b{c}}))\b{P} \overline{\b{v}}} = \inner{{\b{v}}}{\b{J}_g(\b{z})^\T \b{P}^\T \b{M}_{\X'}(\b{P}(g(\b{z}) - \overline{\b{c}}))\b{P}\b{J}_g(\b{z}){\b{v}} },
\end{align}
where ${\b{v}}', {\b{x}}'\in\R^{d'}$ the point and the tangent vector in $\X'$, $\b{P}\in\R^{d' \times D}$ is the projection matrix derived from PCA with $\overline{\b{c}}\in\R^D$ the center of the data, $\overline{\b{x}},~\overline{\b{v}}\in\R^D$ the point and the tangent vector in $\X$ and $\b{z},{\b{v}}\in\R^d$ the latent space inputs with the Jacobian $\b{J}_g(\b{z})\in\R^{D\times d}$. Note that we can directly use the same setting when $g(\cdot)$ is a stochastic generator.

A simple constructive example is to consider the data in Fig.~\ref{appendix:fig:compare_extrapolations}, and expand the dimensions by concatenating 100 columns with noise sampled from $\varepsilon_i \sim \N(0,0.001^2)$ as $[\b{x}, \varepsilon_1,\dots, \varepsilon_{100}]$. Obviously, the structure of the actual data manifold will not be different in $\X = \R^{103}$, and also, we can ``project'' it in $\X' = \R^3$ by excluding the last 100 columns. Therefore, we can construct the ``ambient'' metric  $\b{M}_{\X'}(\cdot)$ in $\X'$, which will be induced on the 3-dimensional subspace in $\X$ where the actual data manifold lies. Thus, the shortest path computed in $\X'$ actually corresponds to the path on $\M$ in $\X$ that lies on a 3-dimensional subspace. As regards the real data, the extra dimensions might not be just noise, but high frequency context, which commonly does not affect the underlying structure of the manifold. Hence, the shortest paths computed in $\X'$ are able to approximate closely the true paths on the actual data manifold $\M\subset \X$, as long as the linear projection step does not change the structure of $\M$ in $\X'$.

%As a simple example consider the $1$-dimensional manifold in $\R^3$ parametrized as $\b{x}(t) = [t, \cos(2\pi t), \sin(2 pi t)]$. Then, if $\b{x} = [x_1, x_2, x_3, e_1, \dots, e_{D-3}]$ where $e_i\sim\N(0,1)$ the $\M$ lies in $\R^D$. Therefore, if we project the $\M$ in $\R^3$ we will simply remove the extra non informative dimensions, while we preserve the structure of $\M$, while if we project in $\R^2$ the structure will be not preserved.

\section{Details for the construction of ambient Riemannian metrics}
\label{appendix:metric_construction}

In this section we provide the details for constructing the metrics that have been used in the paper. As we discussed above, the ambient metrics can be either constructed in $\X$ or in lower dimensional space $\X'$ where we project linearly the given data manifold.

\subsection{Local linear discriminant analysis based Riemannian metric}
\label{appendix:sec:lda_metric}

To compute the ambient metric for a test point $\b{x}\in\X=\R^D$ using the local LDA we have first to learn the base metrics for a set of points $\mathcal{S}=\{\b{x}_s\}_{s=1}^S$ following the approach of \cite{Hastie94discriminantadaptive}, and then, compute the weighted average (see Sec 3.1, Eq.~4). Based on a given labeled set $\mathcal{D}=\{\b{x}_n, y_n\}_{n=1}^N$ the metric at each $\b{x}_s$ is defined as
\begin{align}
\b{M}_s & = \b{W}_s^{-1} \b{B}_s \b{W}_s^{-1} + \varepsilon \b{W}_s^{-1},
\end{align}
where $\epsilon>0$ a small scalar to avoid degenerate metrics, the $\b{W}_s\in\R^{D\times D}$ is called the within covariance matrix and $\b{B}_s\in\R^{D\times D}$ the in-between covariance matrix. Let $K$ be the number of the $k$-nearest neighbors denoted with the set $\text{knn}(\b{x}_s)$ computed under the initial $\b{M}_s=\Id_D$ and $d_s = \norm{\b{M}^{\sfrac{1}{2}}_s(\b{x} - \b{x}_s)}_2$. We use a  weighting function $w_s(\b{x}) = \left[1 - \left({d_s}/{\sigma_s}\right)^3\right]^3 \cdot {1}_{\{d_s<\sigma_s\}}$ where $\sigma_s = \max_{k \in \text{knn}(\b{x}_s)} d_k$. Then, from the labeled point set we consider only the ones that are within the $\text{knn}(\b{x}_s)$, and thus, the matrices within and in-between are defined as
\begin{align}
\b{W}_s &= \frac{1}{\sum_{k=1}^K w_s(\b{x}_k) }\sum_{c\in\mathcal{C}} \sum_{n:y_n = c} w_s(\b{x}_n) (\b{x}_n - \b{m}_c)(\b{x}_n - \b{m}_c)^\T,\\
\b{m}_c &= \frac{1}{\sum_{k:y_k = c} w_s(\b{x}_k)} \sum_{n:y_n = c} w_s(\b{x}_n) \b{x}_n,\\
\b{B} &=  \sum_{c\in\mathcal{C}} \pi_c (\b{m}_c - \b{m})(\b{m}_c - \b{m})^\T,\\
\pi_c &= \frac{\sum_{n:y_n = c} w_s(\b{x}_n) }{\sum_{k=1}^K w_s(\b{x}_k)}.
\end{align}
Using the updated metrics $\b{M}_s$ we iterate the procedure i.e. finding the $\text{knn}(\b{x}_s)$, computing $d_s$, etc, until either a fixed point is found i.e., the $\b{M}_s$ matrices do not change, or if we exceed a pre-specified number of iterations. Moreover, we use only the diagonal $\b{W}_s$ since in higher dimensions is easier to get degenerate metrics when this matrix is full.

As we have discussed in the main paper the construction of the base metrics $\b{M}_k$ is problem dependent. Hence, these can be constructed in any meaningful way, such that to provide the essential high-level information or domain knowledge for the problem we want to model. For further examples, we could construct these metrics based on ordinal information between points or triplet constraints. In general, this is a metric learning related problem \cite{surez2018tutorial}.

\subsection{Density and data support based Riemannian metric}
\label{appendix:sec:density_based_metric}

In order to construct a probability density function based ambient metric, essentially, we want to roughly estimate the density of the high dimensional data. A relatively simple, easy and robust model to learn such a density is the Gaussian Mixture Model (GMM). So in practice, we want to learn a $h(\b{x}) = \sum_{k=1}^K \pi_k \N(\bs{\mu}_k, ~\bs{\Sigma}_k)$, with $\sum_{k=1}^K \pi_k = 1$. However, we have to pay attention to some details. First we want to avoid centers $\bs{\mu}_k$ with huge covariance $\bs{\Sigma}_k$ that are placed outside of the data distribution. For that reason we chose to use the same covariance matrix for all the data $\bs{\Sigma}_k = \bs{\Sigma}$. Intuitively, we want this covariance to be roughly a spherical one, in order to cover the whole data manifold with balls or ellipsoids. So we chose $\bs{\Sigma} = \text{diag}(\sigma_1^2, \sigma_2^2, \dots, \sigma^2_D)$.

A second problem is that in higher dimensions the $|\bs{\Sigma}|$ can be tricky. In particular, the normalization constant will be an issue, since if many $\sigma_d<1$ the $|\bs{\Sigma}|\rightarrow 0$. For that reason we use the un-normalized Gaussian mixture model and this is not a problem, because all of the components share the same covariance. Of course, we are still able to set the parameters $\alpha,\varepsilon$ such that to lower and upper bound the metric. In some sense, these parameters define one aspect of the manifold's curvature, since they
define how big is the difference of the metric between the points where $h(\b{x}) \rightarrow 0$ and $h(\b{x}) \rightarrow 1$.

One drawback of this method, is that the metric will shrink the distances accordingly to the data density in the ambient space is high. Obviously, in some cases this might be a meaningful behavior. However, we might want to simply move near the data and not necessarily analogous to the corresponding density. So a close related approach is to utilize a positive function $h(\b{x}) = \sum_k w_k \phi_k(\b{x})$, with $w_k>0$ and $\phi_k(\b{x}) = \exp(- 0.5 \cdot \lambda_k \norm{\b{x}-\b{c}_k}_2^2)$, that is trained in such a way that the output near the given data is $h(\b{x})\rightarrow 1$, otherwise $h(\b{x})\rightarrow 0$. One way to train the parameters is to fix $\b{c}_k$ using $k$-means, setting the bandwidth $\lambda_k =\frac{1}{2}\left[ \kappa \frac{1}{|\mathcal{C}_k|}  \sum_{\b{x} \in \mathcal{C}_k }\norm{\b{x} - \b{c}_k}_2\right]^{-2}$ where $\kappa>0$ a scaling factor, $\mathcal{C}_k$ the points in the cluster of $\b{c}_k$, and the $w_k$ can be found using a closed form solution or gradient descent under the mean squared error $L(\b{w}) = \sum_{n=1}^N \norm{1 - h(\b{x}_n)}_2^2$. Obviously, this is a relatively simple model, however, it  models very well the desired behavior of the ambient metric.

\subsection{Cost based Riemannian metric}
\label{appendix:section:cost_based_metric}

The cost related ambient Riemannian metric essentially pulls the shortest paths towards regions of the ambient space $\X$ where the cost is low. For our experiments we used a relatively simple and interpretable cost function utilizing again the RBF network $h(x) = \sum_k y_k \phi_k(\b{x})$ with basis functions $\phi_k(\b{x}) = \exp\left(-\frac{1}{2\sigma^2} \norm{\b{x}- \b{c}_k}_2^2 \right)$ and $y_k>0$ some given values. Apart from the simplicity, this type of cost function has a very interpretable behavior, since it defines regions in $\X$ where the cost is high and the corresponding regions in $\Z$ will be avoided by the shortest paths. Intuitively, these can be neighborhoods of points in $\X$ that we want to avoid as we move on the data manifold.

\subsection{Can we construct the $\b{M}_{\X}(\cdot)$ in $\Z$?}

A logical question is, why we do not construct the informative metric directly in $\Z$ using the latent codes, and simply, combine it linearly with the pull-back metric that is induced by the generator? The answer is quite straight forward though. The metric $\b{M}_\X(\cdot)$ is mainly based on Euclidean distances. Therefore, the definition of this Riemannian metric in the latent space $\Z$ is impossible, since using the Euclidean distance in $\Z$ is fundamentally wrong and misleading.

\clearpage

\section{Approximation of the expected Riemannian metric in the latent space}
\label{appendix:section:stochastic_vs_deterministic}

Here we discuss the approximation to the true expected Riemannian metric, where we evaluate the ambient metric only on the expected generated $\M_\Z=\mu(\Z)$. In particular, the true stochastic Riemannian metric in the latent space is written as
\begin{align}
\label{appendix:eq:random_metric_latent_space}
\b{M}_\varepsilon(\b{z} ) = \left[\b{J}_\mu(\b{z}) +  \b{J}_\sigma(\b{z})\varepsilon\right]^\T \b{M}_\X\left(\mu(\b{z}) + \text{diag}(\varepsilon) \cdot \sigma(\b{z}) \right) \left[\b{J}_\mu(\b{z}) + \b{J}_\sigma(\b{z})\varepsilon\right],
\end{align}
for which we can approximate the expectation in the latent space as ${\b{M}}(\b{z}) = \E_{\varepsilon \sim p(\varepsilon)}[\b{M}_\varepsilon(\b{z})]$ with $\varepsilon\sim\N(\b{0}, \Id_D)$. Even if this is a doable computation, in practice, we need to estimate this metric, as well as, its derivative for all the computations on a Riemannian manifold. This directly means that the computational cost will be extremely high, and hence, prohibited. For this reason we provide the following relaxation
\begin{align}
\label{appendix:eq:expected_metric_latent_space}
{\b{M}}(\b{z} ) =\b{J}_\mu(\b{z})^\T \b{M}_\X\left(\mu(\b{z})\right) \b{J}_\mu(\b{z}) + \b{J}_\sigma(\b{z})^\T \b{M}_\X\left(\mu(\b{z})\right) \b{J}_\sigma(\b{z}).
\end{align}
Here we are based on the realistic assumption that the generator's uncertainty in the regions of the latent space with representations of the training data to be $\sigma(\b{z})\rightarrow \b{0}$. The reason is that $\mu(\b{z})$ is trained to reconstruct sufficiently well the training data $\b{x}$, and we are also based on the main assumption that the training data lie \emph{near} a manifold $\M\subset \X$. This essentially implies that the corresponding deviation of $\b{x}$ from $\M$ will be negligible and our $g(\Z)=\M_\Z\approx \M$. Therefore, the Eq.~\ref{appendix:eq:random_metric_latent_space} will become first
\begin{align}
\label{appendix:eq:relaxed_random_metric_latent_space}
\widetilde{\b{M}}_\varepsilon(\b{z} ) = \left[\b{J}_\mu(\b{z}) +  \b{J}_\sigma(\b{z})\varepsilon\right]^\T \b{M}_\X\left(\mu(\b{z})\right) \left[\b{J}_\mu(\b{z}) + \b{J}_\sigma(\b{z})\varepsilon\right],
\end{align}
and we will compute the expectation upon this to get the ${\b{M}}(\b{z}) = \E_\varepsilon[\widetilde{\b{M}}_\varepsilon(\b{z} )]$ that is shown in Eq.~\ref{appendix:eq:expected_metric_latent_space}. As regards the regions far from the latent codes where $\sigma(\b{z}) \gg 0$, the $\b{J}_\sigma(\cdot)$ will be the dominant term, and hence, the contribution of $\b{M}_\X(\mu(\b{z}))$ or even $\b{M}_\X(\mu(\b{z}) + \text{diag}(\varepsilon)\cdot\sigma(\b{z}))$ will be negligible there anyways.

In order to demonstrate this behavior, we generate a dataset near $\M$ as $\b{x}=[x_1, x_2, \sin(x_1)]$ and we add noise using $\N(0,\sigma^2)$ with two different $\sigma=0.1, 0.2$. For the ambient metric we use the cost based RBF approach by selecting 3 points and their 10 nearest neighbors in $\X$ with $y_k=10$. We train two VAEs and we show in the latent space the resulting Riemannian metric with and without the stochasticity of the generator for the evaluation of the ambient metric $\b{M}_{\X}(\cdot)$.

\begin{figure}[h]
	\centering
	\begin{subfigure}[b]{0.195\textwidth}
		\includegraphics[width=\textwidth]{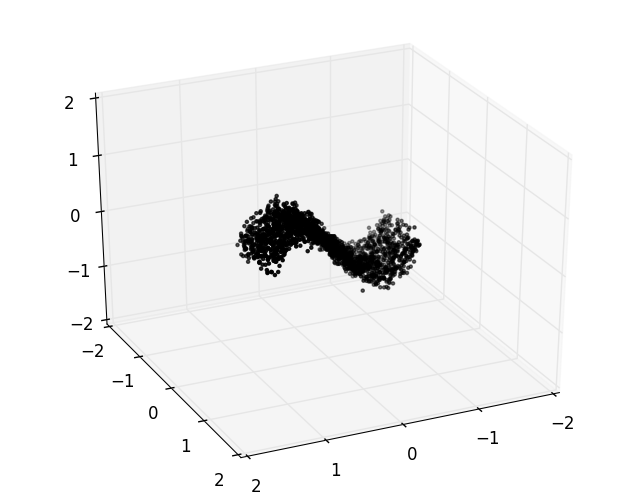}
	\end{subfigure}
	\begin{subfigure}[b]{0.195\textwidth}
		\includegraphics[width=\textwidth]{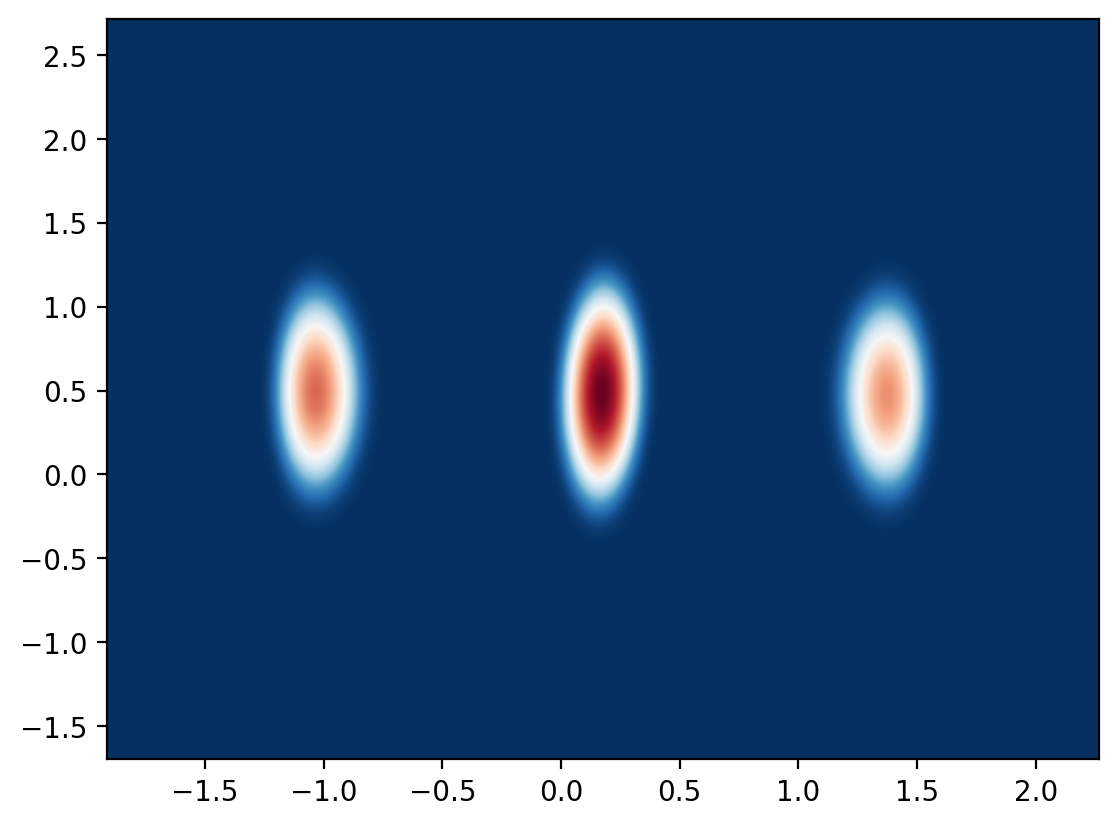}
	\end{subfigure}
	\begin{subfigure}[b]{0.195\textwidth}
		\includegraphics[width=\textwidth]{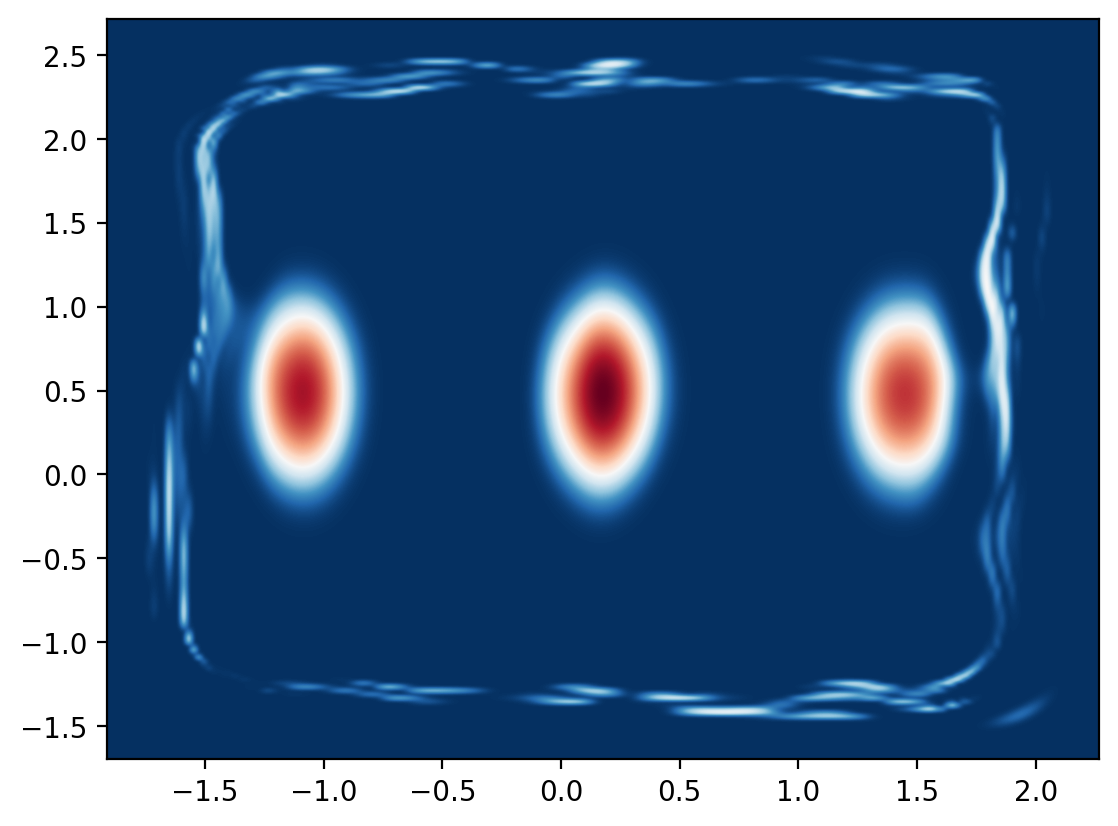}
	\end{subfigure}
	\begin{subfigure}[b]{0.195\textwidth}
		\includegraphics[width=\textwidth]{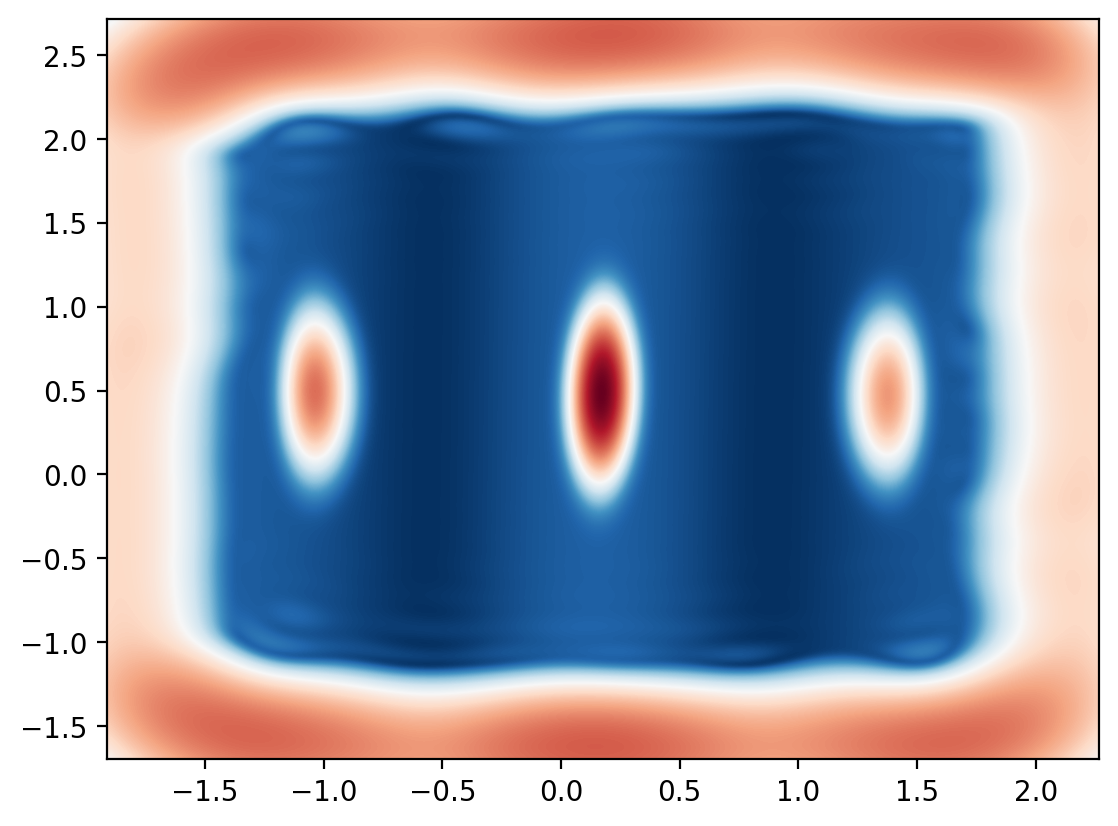}
	\end{subfigure}
	\begin{subfigure}[b]{0.195\textwidth}
		\includegraphics[width=\textwidth]{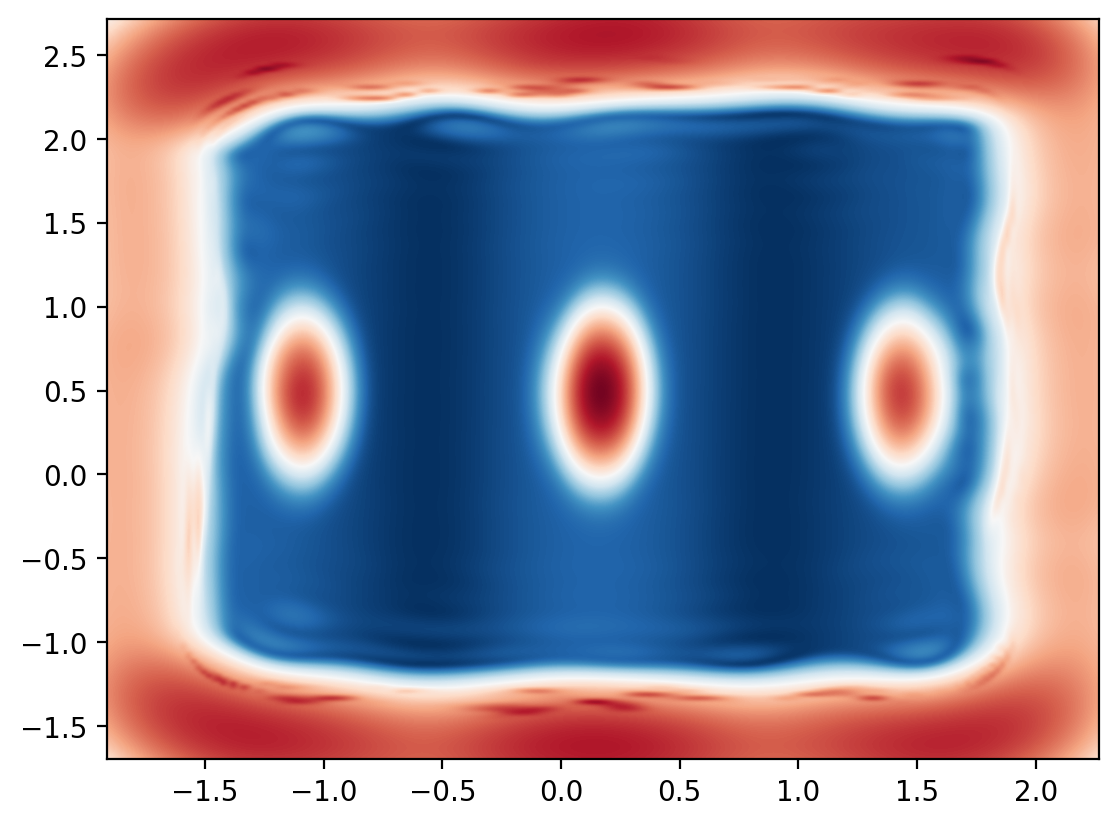}
	\end{subfigure}

	\begin{subfigure}[b]{0.195\textwidth}
		\includegraphics[width=\textwidth]{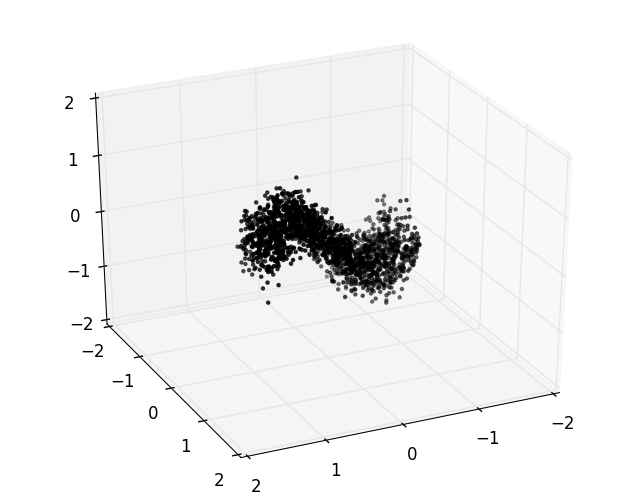}
		\caption*{Data}
	\end{subfigure}
	\begin{subfigure}[b]{0.195\textwidth}
		\includegraphics[width=\textwidth]{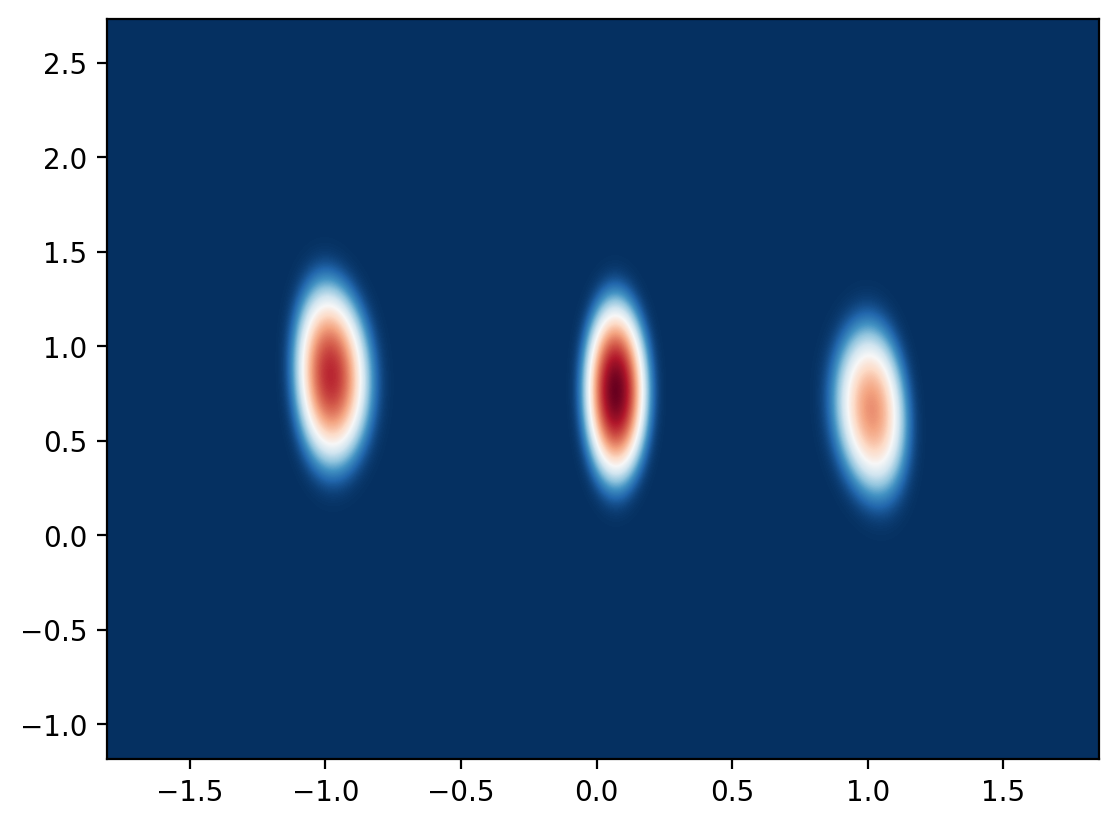}
		\caption*{$\sqrt{|\b{M}_\X(\E_{\varepsilon} [g_\varepsilon(\b{z})])|}$}
	\end{subfigure}
	\begin{subfigure}[b]{0.195\textwidth}
		\includegraphics[width=\textwidth]{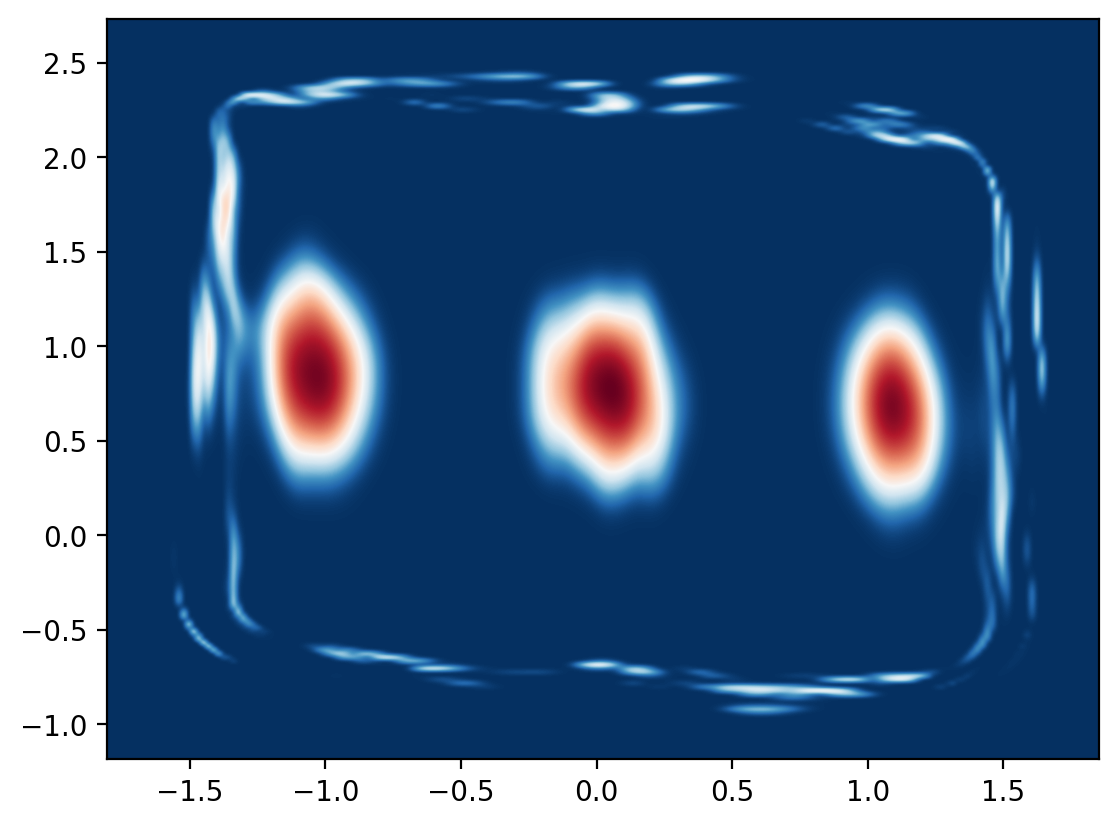}
		\caption*{$\E_{\varepsilon} [\sqrt{|\b{M}_\X(g_\varepsilon(\b{z}))|}]$}
	\end{subfigure}
	\begin{subfigure}[b]{0.195\textwidth}
		\includegraphics[width=\textwidth]{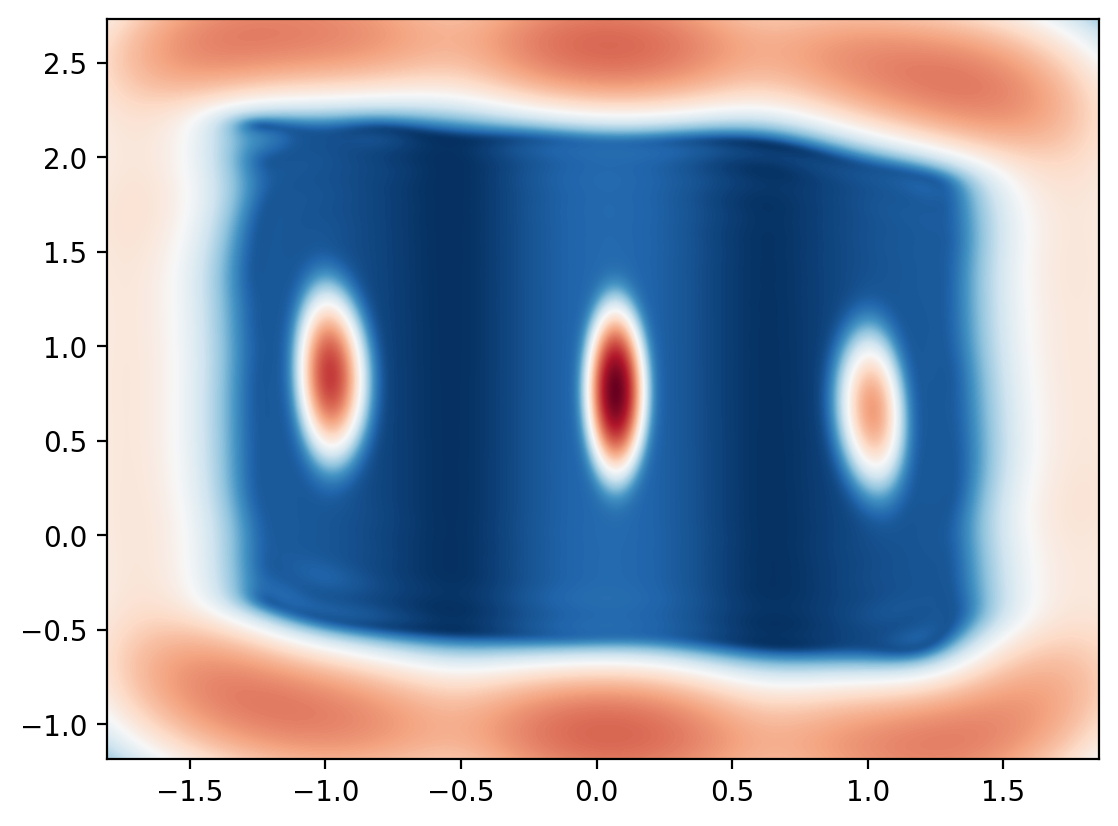}
		\caption*{$\sqrt{|\b{M}(\b{z})|}$}
	\end{subfigure}
	\begin{subfigure}[b]{0.195\textwidth}
		\includegraphics[width=\textwidth]{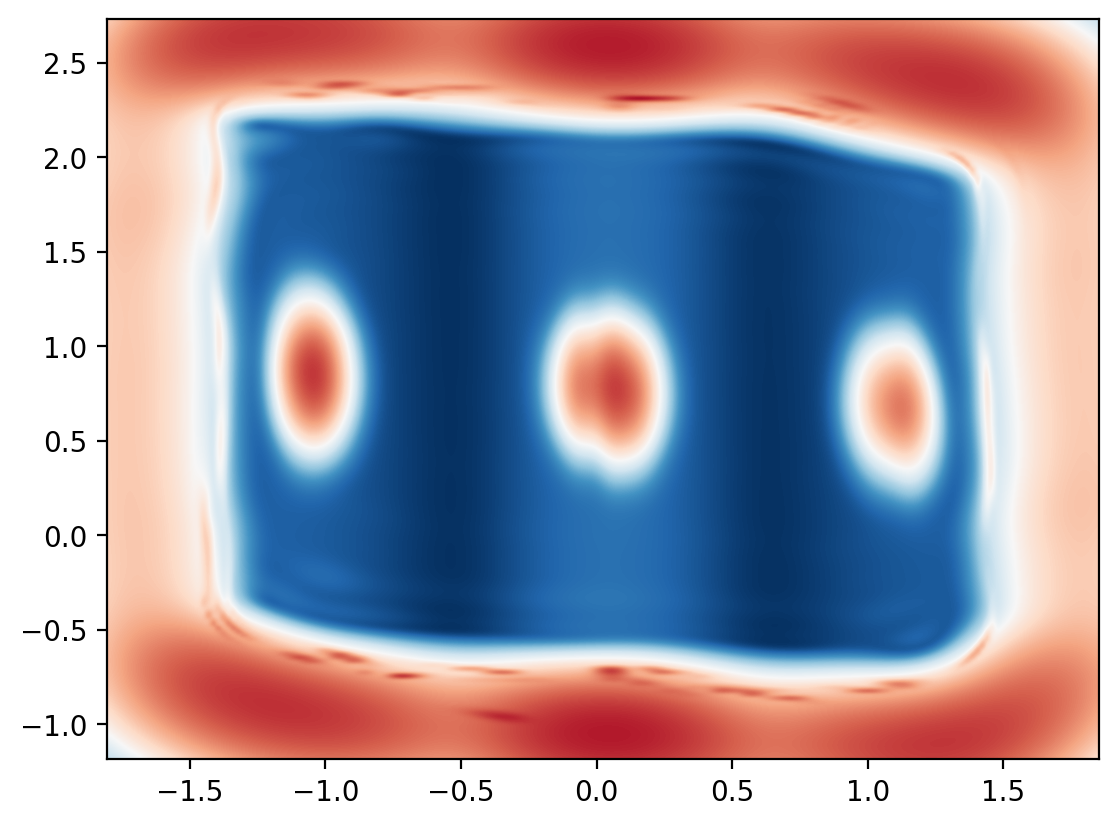}
		\caption*{$\E_\varepsilon[\sqrt{|\b{M}_\varepsilon(\b{z})|}]$}
	\end{subfigure}

	\caption{In the \emph{top} row we added noise with $\sigma=0.1$ and in the \emph{bottom} row $\sigma=0.2$. We see that in the two last columns and per row, that the structure does not change significantly when we use the proposed relaxation.}
	\label{appendix:experiment:compare_stochastic_approximation_deterministic}
\end{figure}

From the results in Fig.~\ref{appendix:experiment:compare_stochastic_approximation_deterministic} we observe that by considering the true expected Riemannian metric, the captured structure does not differ significantly from the one we get using the proposed relaxation, especially, near the latent codes. Therefore, by taking into account the trade off, we argue that it is sufficient to use the expected generated manifold $\M_\Z = \E_{\varepsilon \sim p(\varepsilon)}[\M_\varepsilon]$ such that to evaluate the ambient Riemannian metric $\b{M}_\X(\cdot)$ as it is shown in Eq.~\ref{appendix:eq:expected_metric_latent_space}.

\clearpage
\section{Experiments}
\label{appendix:section:experimental_details}

In this section we provide further details and discussion regarding the conducted experiments.

\subsection{Details for the Generative Adversarial Network demonstrations}

\label{appendix:gan_experiment}

\paragraph{Synthetic data.} The synthetic data are generated as follows. First, we pick the centers of 6 Gaussian distributions $\N(\mu, 0.2^2 \cdot \Id_2)$ uniformly on a circle with radius 3 and one in the center. Then, we generate 300 points from each Gaussian that can be seen as the actual latent representations, and thus, we construct the data $\b{x} = [z_1, z_2, 0.3 \cdot(z_1^2 + z_2^2) + \varepsilon]$, where $\varepsilon\sim\N(0,0.1^2)$. We used a Wassestein GAN with latent space  $\Z=\R^2$ and ambient space $\X=\R^3$, with functions

\begin{table}[h]
	\centering
	\begin{tabular}{c c c c}
		\toprule
		\emph{Function} & \emph{Layer 1} & \emph{Layer 2} & \emph{Output}\\\midrule
		$f(\b{z})$ & $\small\texttt{tanh(2)}$ & $\small\texttt{tanh(3)}$ & $\small\texttt{linear(3)}$\\
		$d(\b{x})$ & $\small\texttt{LeakyReLU(3) + Dropout(0.3)}$ & $\small\texttt{LeakyReLU(3) + Dropout(0.3)}$ & $\small{\texttt{linear(1)}}$\\
		\bottomrule
	\end{tabular}
\end{table}

We trained the model using Adam optimizer for 1000 epochs with stepsize $1e^{-2}$ and batch sizes of size 128, and also, we used $\ell_2$ regularization for the weights with parameter $1e^{-5}$. The discriminator is trained for 5 more steps within each epoch and the weights are clamped into the interval $(-0.01,0.01)$ to satisfy the Lipschitz constraint of the Wasserstein GAN. For the sampling of the latent codes we experimented both with standard Markon Chain Monte Carlo (MCMC), as well as rejection sampling \citep{Bishop:2006:PRM}. For the mixture of LAND we used the default training procedure with 10 epochs and full covariance matrices per component. In order to construct the RBF ambient metric we used 20 components and the scaling factor of the bandwidth was set to $\kappa=1$ as discussed in Appendix~\ref{appendix:sec:density_based_metric}.

\paragraph{MNIST data.} We used the digits 0,1,2, we scaled them in the interval $[-1,1]$ and we added point-wise noise $\varepsilon \sim\N(0,0.02^2)$, such that the data to not lie exactly on $\M$. Thus, is easier to train the generator without utilizing the bounded $\texttt{tanh}(\cdot)$ in the output layer to clip the values. Because, in such a case the meaningful extrapolation is not anymore useful, since the linear part will be also clipped. However, when we show or pass the images into the critic $d(\b{x})$ first we apply the $\texttt{tanh}(\cdot)$ function. Specifically, the latent space is $\Z=\R^5$ and $\X=\R^{784}$ and the functions are defined as

\begin{table}[h]
	\centering
	\begin{tabular}{c c c c}
		\toprule
		\emph{Function} & \emph{Layer 1} & \emph{Layer 2} & \emph{Output}\\\midrule
		$f(\b{z})$ & $\small\texttt{tanh(128)}$ & $\small\texttt{tanh(256)}$ & $\small{\texttt{linear(784)}}$\\
		$d(\b{x})$ & $\small\texttt{tanh+LeakyReLU(128)+Drop(0.3)}$ & $\small\texttt{LeakyReLU(128)+Drop(0.3)}$ & $\small{\texttt{linear(1)}}$\\
		\bottomrule
	\end{tabular}
\end{table}

The discriminator is trained for 5 more steps within each epoch and the weights are clamped into the interval $(-0.01,0.01)$ to satisfy the Lipschitz constraint of the Wasserstein GAN. The model is trained using Adam optimizer for 10000 epochs and batch size 64 with stepsize $1e^{-4}$ and $\ell_2 $ regularization of the weights with parameter $1e^{-7}$. For the sampling of the latent codes we experimented both with standard Markon Chain Monte Carlo, as well as rejection sampling. The ambient Riemannian metric is constructed with the RBF method discussed in Appendix~\ref{appendix:sec:density_based_metric} and we used 100 centers and $\kappa=0.33$ which decreases the bandwidth of the RBF kernels. Moreover, we projected linearly the data to a lower dimensional space $\X' = \R^{10}$ using principal components analysis (PCA), where we construct the metric $\b{M}_{\X'}(\cdot)$ (see  Appendix~\ref{appendix:manifold_theory}). Also, to stabilize training and to prevent mode collapse, we include a VAE loss when we train the generator with a regularization parameter $1e^{-5}$.

\begin{figure}[h]
	\centering
	\begin{subfigure}[b]{0.32\textwidth}
		{\includegraphics[width=\textwidth]{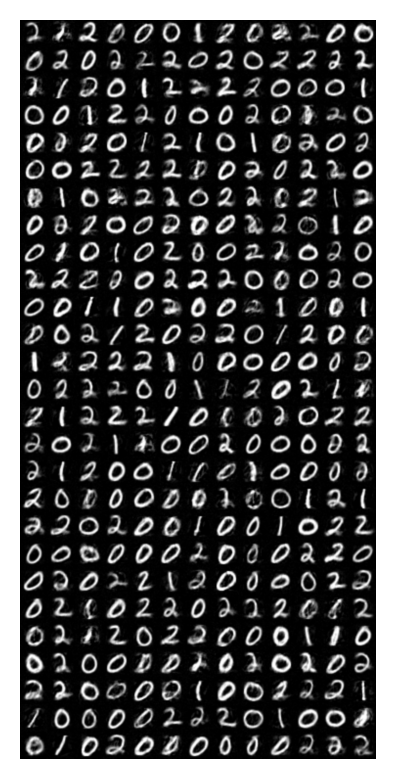}}
		\caption{From $q(\b{z})$ with $\b{M}_\X(\cdot)$.}
	\end{subfigure}
	~
	\begin{subfigure}[b]{0.32\textwidth}
		{\includegraphics[width=\textwidth]{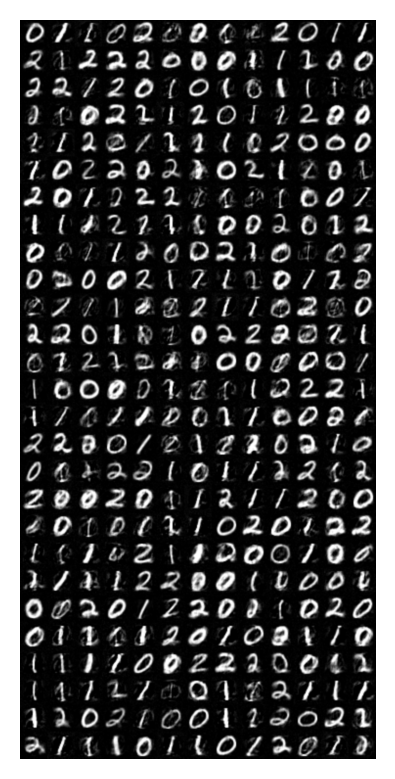}}
		\caption{From $q(\b{z})$ without $\b{M}_\X(\cdot)$.}
	\end{subfigure}
	~
	\begin{subfigure}[b]{0.32\textwidth}
		{\includegraphics[width=\textwidth]{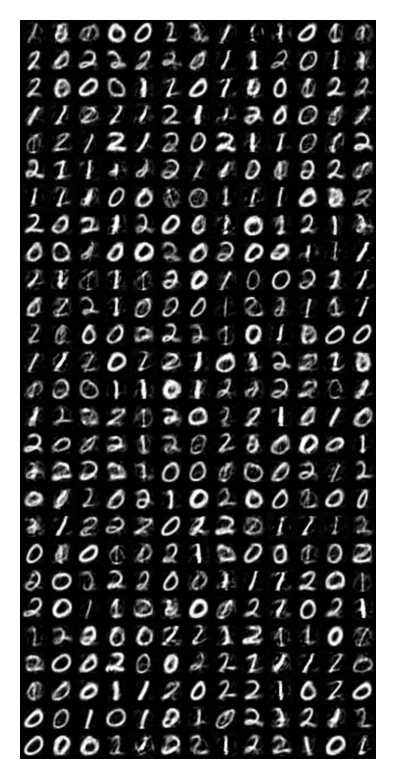}}
		\caption{From prior $p(\b{z})$.}
	\end{subfigure}
	\caption{Additional samples for the MNIST data, GAN experiment with $5$-dimensional latent space. Note that our proposed method does not generate a lot of ghostly samples, which fall on parts of the ambient space with no given data nearby.}
	\label{appendix:samples_gan_extra}
\end{figure}

\begin{figure}[h]
	\centering
	\begin{subfigure}[b]{1\textwidth}
		{\includegraphics[width=\textwidth]{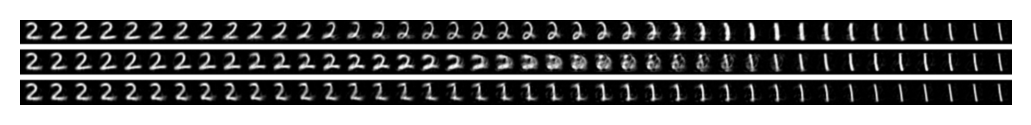}}
	\end{subfigure}
	
	\begin{subfigure}[b]{1\textwidth}
		{\includegraphics[width=\textwidth]{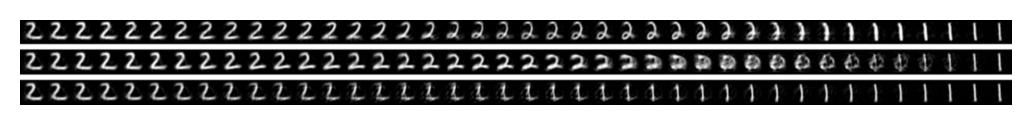}}
	\end{subfigure}
	
	\begin{subfigure}[b]{1\textwidth}
		{\includegraphics[width=\textwidth]{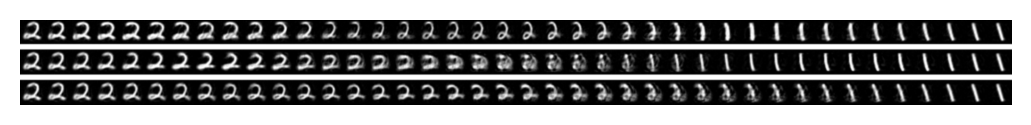}}
	\end{subfigure}
	
	\begin{subfigure}[b]{1\textwidth}
		{\includegraphics[width=\textwidth]{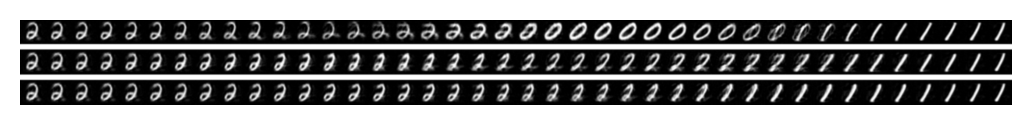}}
	\end{subfigure}
	
	\begin{subfigure}[b]{1\textwidth}
		{\includegraphics[width=\textwidth]{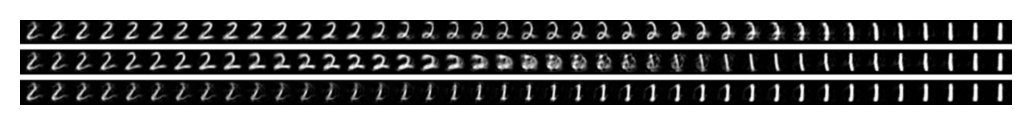}}
	\end{subfigure}
	
	\begin{subfigure}[b]{1\textwidth}
		{\includegraphics[width=\textwidth]{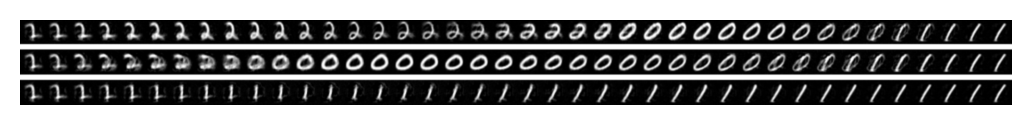}}
	\end{subfigure}
	
	\begin{subfigure}[b]{1\textwidth}
		{\includegraphics[width=\textwidth]{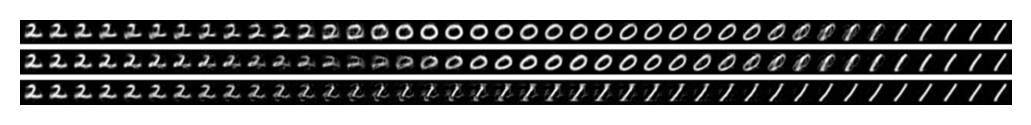}}
	\end{subfigure}
	
	\caption{Additional interpolation results. From \emph{top} to \emph{bottom}: our interpolation, interpolation without using the $\b{M}_\X(\cdot)$, linear interpolation. Note that our interpolation (top rows) avoids the ``shortcuts'' of the simple shortest path interpolant (middle rows), while the linear path is arbitrary.}
	\label{appendix:interpolation_gan_extra}
\end{figure}

We see that using the ambient metric $\b{M}_{\X'}(\cdot)$ improves the sampling, and some additional results are shown in Fig.~\ref{appendix:samples_gan_extra}. The resulting samples due to the $\b{M}_{\X'}(\cdot)$ lie closer to the support of the given data manifold, and also, we avoid samples in-between the disconnected components in $\X$. Moreover, we show some additional interpolations (see Fig.~\ref{appendix:interpolation_gan_extra}) where we again see that using the ambient metric improves the interpolations. In particular, the difference between our proposed approach and the standard shortest paths is that the ambient Riemannian metric pulls the paths towards the data manifold and avoids ``shortcuts''. Intuitively, shortcut means that the path moves optimally on the generated $\M_\Z$, but not necessarily always near the given data manifold. Note that $\M_\Z$ is a continuous smooth surface and some parts are not near the given data points/manifold, but without considering the $\b{M}_{\X'}(\cdot)$ it might be cheap to move there which is a misleading behavior.

\paragraph{Pre-trained model.} We used as generator a Progressive GAN (PGAN) \citep{karras2018progressive} which utilizes a latent space $\Z=\R^{d}$ with $d=512$ and has ambient space $\X=\R^D$ with $D={256\times 256 \times 3}$, while the labeled training dataset is not directly provided. Note that in this generator it is not included the linear map to ensure meaningful extrapolation, and also, due to $\texttt{ReLU}(\cdot)$ activation the $\b{M}(\cdot)$ is not sufficiently smooth. However, we tested how the additional consideration of an ambient metric can affect the shortest paths, and additionally, we use a heuristic that we describe below for computing approximate shortest paths where a smooth metric $\b{M}(\cdot)$ is not necessary. Moreover, we upscaled the standard CelebA labeled dataset of size $128\times128\times3$ to $D$ in order to be able to compute the linear projection matrix $\b{P}\in\R^{d' \times D}$ from $\X$ to $\X'$ with $d'=1000$ and the linear mean ${\b{c}}\in\R^D$. See discussion in Appendix~\ref{appendix:manifold_theory} regarding this linear projection step.

Obviously, the computation of the Jacobian matrix for this $g(\cdot)$ is prohibited due to the size of the latent space and the complexity of the model, even with finite differences. So we relied on some tricks that we explain below, in order to be able to compute relatively efficient shortest paths. First, we define  a new latent space $\widetilde{\Z}=\R^{\tilde{d}}$ of dimensionality $\tilde{d}<d$ with $\tilde{d}=10$ and we construct an ortho-normal  random projection matrix $\widetilde{\b{U}}\in\R^{d \times \tilde{d}}$. In such a way, we can compute shortest paths in $\widetilde{\Z}$ that correspond to shortest paths on a $\tilde{d}$-dimensional sub-space in $\Z$. Hence, in total we have
\begin{align}
\widetilde{\Z} \xrightarrow{\widetilde{\b{U}}\cdot} \Z \xrightarrow{g(\cdot)} \X \xrightarrow{\b{P}~( \cdot - \b{c})} \X'.
\end{align}
Clearly, this tactic constraints the shortest paths to lie on the linear subspace spanned by $\widetilde{\b{U}}$ in $\Z$, and hence, they are not be able to move freely in the whole $\Z$. However, this approximation allows us to compute shortest paths in reasonable time. Essentially, we induce the pull-back Riemannian metric in a lower dimensional latent space $\widetilde{\Z}$, while the $\widetilde{\b{U}}$ matrix does not introduce further distortions.

The main reason for using the $\widetilde{\b{U}}$ is that when $\tilde{d}$ is relatively small, we are able to compute the Jacobian matrix $\widetilde{\b{J}}_g(\tilde{\b{z}})\in\R^{D\times \tilde{d}}$ using finite differences. In particular, in the latent space $\tilde{d}$ we approximate the $j$-th column of the Jacobian from $\widetilde{\Z}\rightarrow \X$ with finite differences as
\begin{align}
\widetilde{\b{J}}^{j}_g(\tilde{\b{z}}) = \lim_{\lambda \rightarrow 0} \frac{g(\widetilde{\b{U}}\cdot (\tilde{\b{z}} + \lambda \b{e}_j)) - g(\widetilde{\b{U}} \cdot \tilde{\b{z}})}{\lambda},
\end{align}
where $\tilde{\b{z}}\in\R^{\tilde{d}}$ and $\b{e}_j = [0,\dots,1,\dots, 0]$ a $\tilde{d}$-dimensional vector of zeros with $1$ at the $j$-th location. Furthermore, we can exploit the forward pass to compute simultaneously all the columns of the Jacobian, by using a batch of inputs that we truncate using the identity matrix $\lambda \Id_{\tilde{d}}$. In such a way, we can compute the Jacobian at a point with only one forward pass with batch size $\tilde{d}+1$. Nevertheless, even in an approximate ODE solver this is still very computationally expensive. So we implemented one heuristic to compute the shortest path based on the idea of ISOMAP \citep{tenenbaum:global:2000}.

We start by sampling 10000 points in $\tilde{d}$ uniformly inside a hyper-sphere of radius $4$ and  using $k$-means we find $100$ prototypes. Note that we do not have access to latent codes, so we want to introduce some artificial codes in $\widetilde{\Z}$. Then, using these prototypes we construct the $K$-nearest neighbor graph with $K=7$ by using the Euclidean distance to find the neighbors. But, for the weight of the edges we use the straight line distance measured under pull-back Riemannian metric that we can evaluate using the finite differences based Jacobian as 
\begin{align}
\text{length}[l(t)] =\int_0^1\sqrt{\inner{\dot{l}(t_n)}{\b{M}_{\text{fd}}(l(t_n))\dot{l}(t_n)}} dt \approx  \sum_{t_n=1}^N \sqrt{\inner{\dot{l}(t_n)}{\b{M}_{\text{fd}}(l(t_n))\dot{l}(t_n)}} \Delta t_n,
\end{align}
where $l(t)$ is the line between two latent points in $\widetilde{\Z}$. For the metric $\b{M}_{\text{fd}}(\cdot)$ first we compute the Jacobian of the total map $\b{P}( g(\widetilde{\b{U}}\cdot l(t)) - \b{c})$ with respect to $l(t)$, which can be achieved by using the finite differences for the Jacobian computation of the map $g(\widetilde{\b{U}}\cdot l(t))$, and then, we use the $\b{M}_{\X'}(\b{P}( g(\widetilde{\b{U}}\cdot l(t)) - \b{c}))$. In particular the metric is equal to \begin{align}
\b{M}_{\text{fd}}(\tilde{\b{z}}) = \left[ \b{P}\parder{g(\widetilde{\b{U}}\cdot \tilde{\b{z}})}{\tilde{\b{z}}}\right]^\T \b{M}_{\X'}(\b{P}( g(\widetilde{\b{U}}\cdot \tilde{\b{z}}) - \b{c})) \left[ \b{P}\parder{g(\widetilde{\b{U}}\cdot \tilde{\b{z}})}{\tilde{\b{z}}}\right].
\end{align}
Essentially, the straight line in $\widetilde{\Z}$ measured under the Riemannian metric will inform us how far on the manifold in the space $\X'$ and under the metric $\b{M}_{\X'}(\cdot)$ are the decoded latent points that seem to be close in the $\tilde{d}$-dimensions.

For two test points in $\widetilde{\Z}$ that we want to compute the shortest path, first we find their closest $K$-neighbors from the points on the graph using the Euclidean metric, and then, we assign the corresponding edge weights using the Riemannian distances. Finally, we chose two auxiliary points, one per kNN set with the smallest Riemannian distance. Thus, we can find the discrete shortest path using Dijkstra's algorithm on the graph using the auxiliary nodes as the boundaries. Note that the path prefers edges with low weight i.e., the edge corresponds to a curve on $\M$ with small length. Ultimately, the continuous path is the a cubic spline interpolation through the points of the discrete path on the graph replacing the two auxiliary points with the test points. Obviously, this is a heuristic methodology to approximate the true shortest path which is inspired by ISOMAP, and also, a very similar heuristic approach that has been proposed in \cite{Chen_2019}.

\begin{figure}[h]
	\centering
	\begin{subfigure}[b]{1\textwidth}
		{\includegraphics[width=\textwidth]{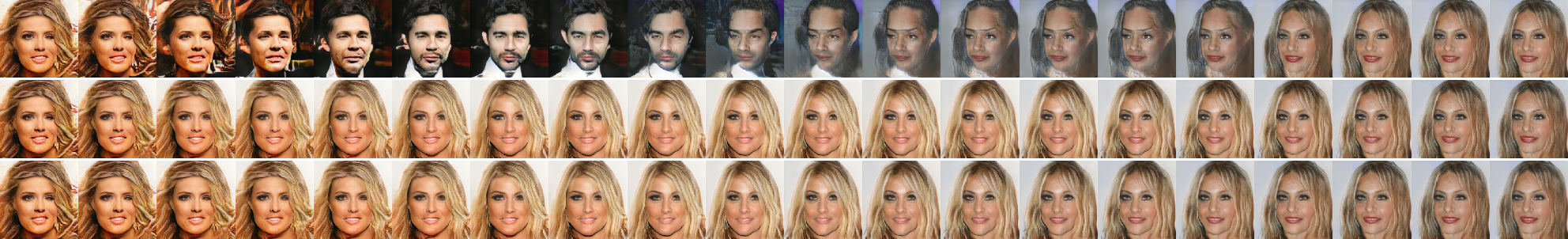}}
	\end{subfigure}

	\vspace{10pt}
	
	\begin{subfigure}[b]{1\textwidth}
		{\includegraphics[width=\textwidth]{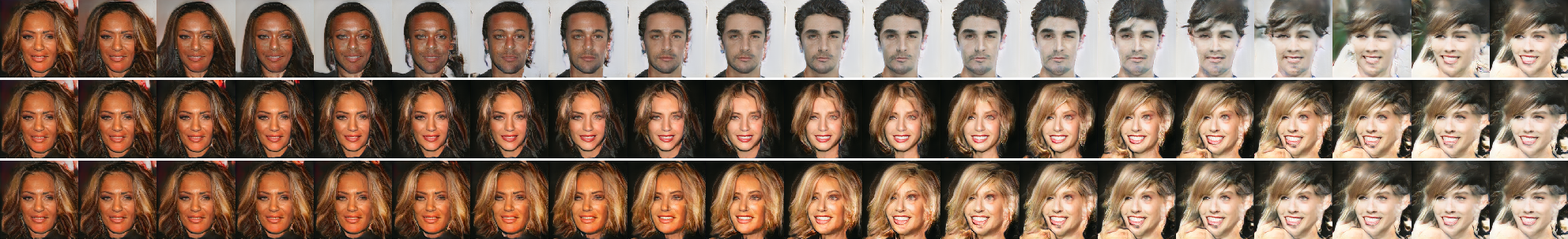}}
	\end{subfigure}
	
	\vspace{10pt}
	
	\begin{subfigure}[b]{1\textwidth}
		{\includegraphics[width=\textwidth]{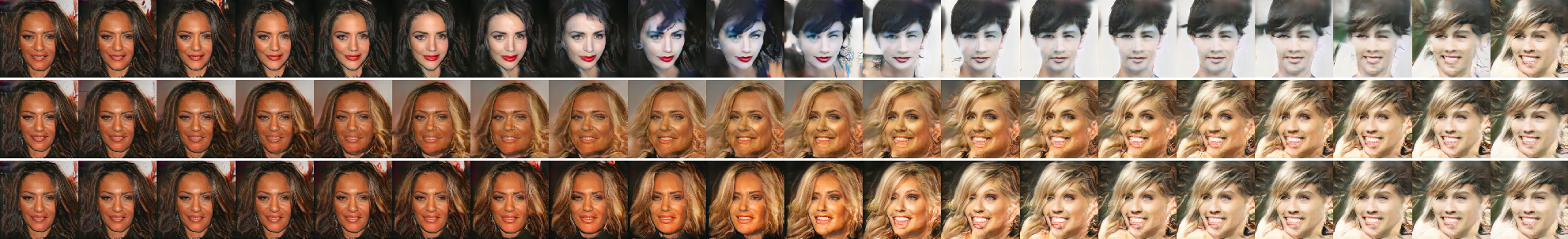}}
	\end{subfigure}
	
	\vspace{10pt}
	
	\begin{subfigure}[b]{1\textwidth}
		{\includegraphics[width=\textwidth]{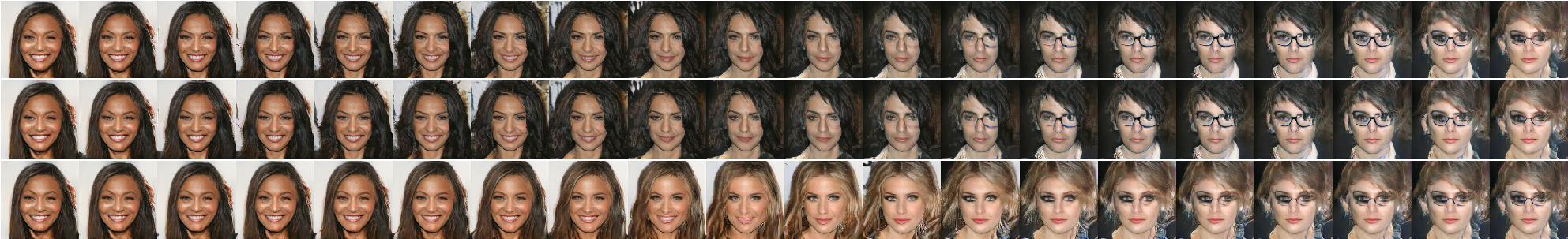}}
	\end{subfigure}

	\vspace{10pt}
	
	\begin{subfigure}[b]{1\textwidth}
		{\includegraphics[width=\textwidth]{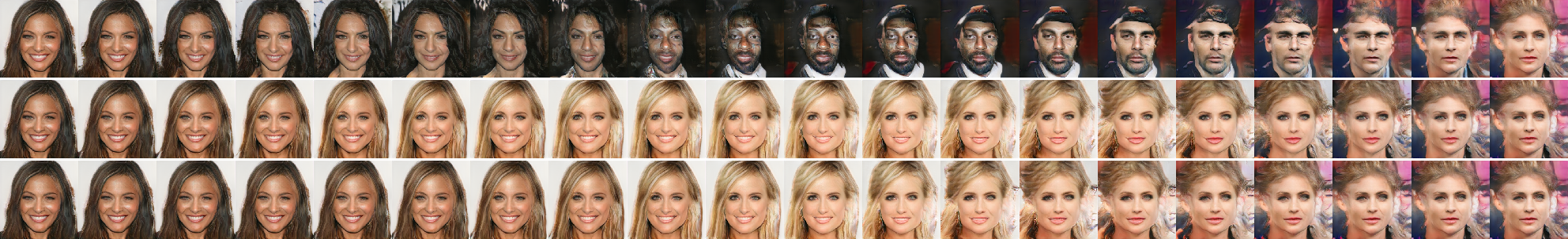}}
	\end{subfigure}

	\vspace{10pt}
	
	\begin{subfigure}[b]{1\textwidth}
		{\includegraphics[width=\textwidth]{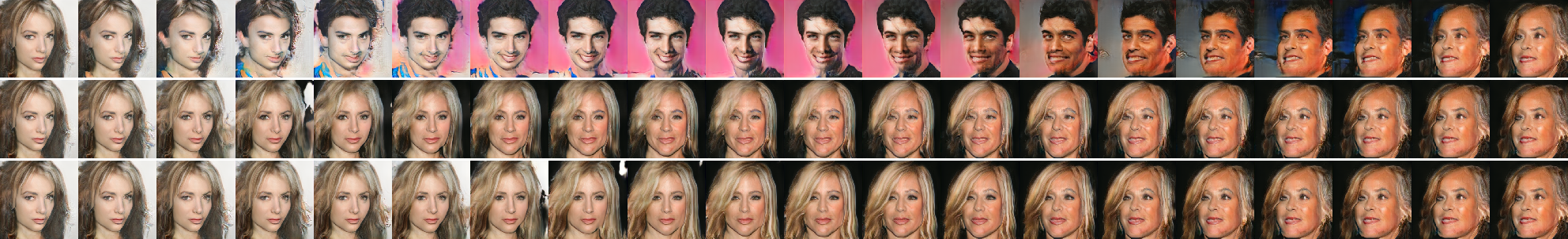}}
	\end{subfigure}

	\vspace{10pt}
	
	\begin{subfigure}[b]{1\textwidth}
		{\includegraphics[width=\textwidth]{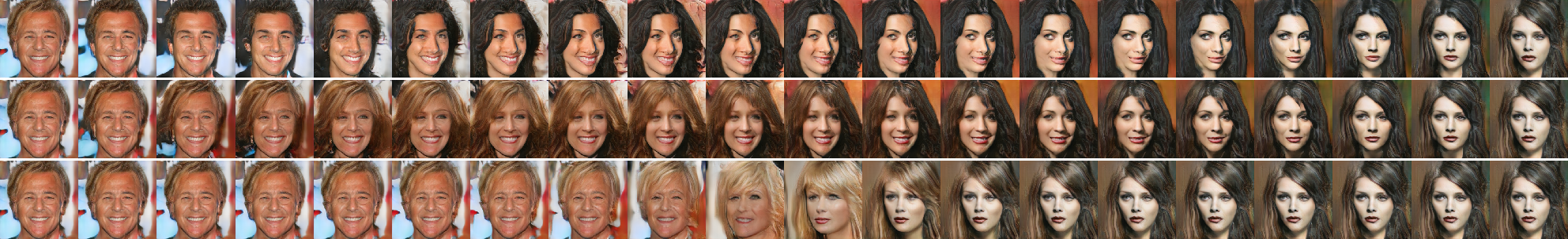}}
	\end{subfigure}

	\vspace{10pt}

	\begin{subfigure}[b]{1\textwidth}
		{\includegraphics[width=\textwidth]{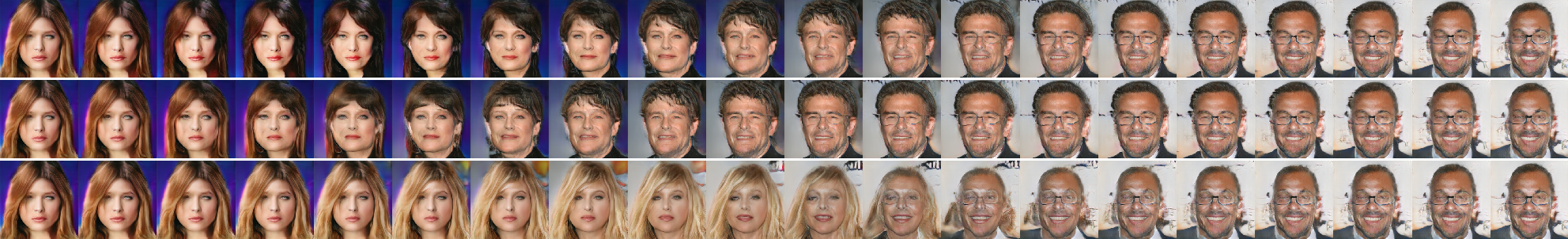}}
	\end{subfigure}

	\caption[none]{Additional interpolation results for the PGAN. From \emph{top} to \emph{bottom}: our interpolation, interpolation without using the $\b{M}_{\X'}(\cdot)$, linear interpolation. Note that our interpolation (top rows) provides a smooth transition between the images, while it avoids the high cost regions (people with blond hair). The shortest path without the ambient metric still provides smooth changes, however, it often crosses high cost regions. The relatively smooth behavior of the straight line is due to the nature of the generator and not due to the actual interpolant.} 
	\label{appendix:fig:interpolation_pgan_extra}
\end{figure}

The task that we want our ambient Riemannian metric to model, is to avoid regions with blond people when interpolating between two latent codes. As we described above we linearly project in $\X'$ the implicitly given data manifold $\M\subset \X=\R^D$, by using the standard labeled CelebA dataset. In $\X'$, we construct the $\b{M}_{\X'}(\cdot)$ which is based on a simple RBF cost based metric (see Appendix~\ref{appendix:section:cost_based_metric}) with $y_k=1e^9$ and $\sigma = 5$. Therefore, we have to define the centers $\b{c}_k \in\X'$. In order to do that, first we train on the labeled CelebA dataset of size $128\times 128\times 3$ a simple convolutional neural network classifier $c(\b{x})$ (see table below). Once the classifier is trained, we decode the nodes of the graph and samples from the prior, which we classify after resizing from $\R^{256\times 256 \times 3}$ to  $\R^{128\times 128 \times 3}$. With these steps, we are able to define the centers of the metric in $\X'$, by using the points that are classified as blond. Note that this is a very simple to implement metric, but rather informative, since the shortest path is penalized heavily when moves close to the high cost regions in $\X'$. Essentially, the (discrete) shortest path avoids the nodes which after decoding fall near the high cost regions in $\X'$. We show some further interpolation results in Fig.~\ref{appendix:fig:interpolation_pgan_extra} using different projection matrices $\widetilde{\b{U}}$, which means that we explore different subspaces in $\Z$, and consequently, on $\M_\Z$.

\begin{table}[h]
	\centering
	\begin{tabular}{c c c}
		\toprule
		\emph{Function} & \emph{Layer 1,2,3}& \emph{Output}\\\midrule
		$c(\b{x})$ & $3 \times [\small\texttt{conv(16,5,1)} + \small\texttt{MaxPool(2)} + \small\texttt{ReLU}]$ & $\small\texttt{Sigmoid(Linear(265, 1))}$\\
		
		\bottomrule
	\end{tabular}
\end{table}

%Brighter  {\protect 
%	\begin{tikzpicture} 
%	\draw [color=red, line width=0.25mm] rectangle (6pt, 6pt);
%	\end{tikzpicture}} denotes higher cost in $\X$.

\clearpage

\subsection{Details for the Variational Auto Encoder experiment}

We used the MNIST digits 0,1,2,3 scaled in the interval $[-1,1]$ and then we added point-wise noise $\varepsilon~ \sim~\N(0,0.02^2)$. As we explained before,  we add the noise such that the data to not lie exactly on $\M$, so that we can train the generator without utilizing the bounded $\texttt{tanh}(\cdot)$ in the output layer to clip the values. However, when we show the images first we apply the $\texttt{tanh}(\cdot)$ function. Note that in the stochastic generator case the meaningful extrapolation is not necessary, even if we use it in our experiments, since the uncertainty quantification helps to properly capture the data manifold structure. The ambient space is $\X=\R^{784}$ and the latent space $\Z=\R^5$. We used the following functions

\begin{table}[h]
	\centering
	\begin{tabular}{c c c c}
		\toprule
		\emph{Function} & \emph{Layer 1} & \emph{Layer 2} & \emph{Output}\\\midrule
		decoder: $f(\b{z})$ & $\small\texttt{softplus(128)}$ & $\small\texttt{softplus(256)}$ & $\small{\texttt{linear(784)}}$\\
		
		decoder: $\beta(\b{z})$ & $\small\texttt{RBF(100)}$ &  & $\small{\texttt{linear(784)}}$\\
		
		encoder: $\mu_\phi(\b{x})$ & $\small\texttt{softplus(256)}$ & $\small\texttt{softplus(128)}$ & $\small{\texttt{linear}_{>0}\texttt{(5)}}$\\
		
		encoder: ${\sigma}^2_\phi(\b{x})$ & $\small\texttt{softplus(256)}$ & $\small\texttt{softplus(128)}$ & $\small{\texttt{softplus(linear(5))}}$\\
		\bottomrule
	\end{tabular}
\end{table}

where the $\beta(\b{z}) + \zeta= \frac{1}{\sigma^2(\b{z})}$ with $\zeta=1e^{-6}$ and $\beta(\b{z})$ is an RBF with 100 centers and only positive weights. We trained the model using Adam optimizer for 1000 epochs and batch sizes of size 64 with stepsize $1e^{-4}$ and also $\ell_2 $ regularization of the weight with parameter $\lambda= 1e^{-5}$.

For the interpolation experiment, the LDA metric is constructed by considering the digtis 0,1,3 in the same class, while in the kernel density estimation experiment every class is separated. We used 2000 randomly chosen training points as the base points $\b{x}_s$, the $\varepsilon=1e^{-3}$, the number of nearest neighbors is $K=50$ and we used a fixed number of iterations 20. See Appendix~\ref{appendix:sec:lda_metric} for details.

For the cost function based ambient metrics we use the RBF cost discussed in Appendix~\ref{appendix:section:cost_based_metric}, and we start by picking 3 latent codes in $\Z$. Then, we decode these points and by using the closest 100 neighbors per decoded point in $\X'$ we constructed the metric with parameters $y_k = 100$ and $\sigma=0.2$. So in total we have 300 RBF basis functions in $\X'$. We used the same approach both in the interpolations and the KDE experiment.

For the linear combination of the ambient metrics we used the weights 1 for the LDA, 0.001 for the local diagonal inverse covariance and 0.1 for the cost metric. We used the same coefficients both in the interpolations and the KDE experiment. Also, the reason for so different coefficients is the scaling of each individual metric. Of course, choosing carefully the parameters of each ambient metric could regularize the scaling differences. However, a principled method to estimate the mixture coefficients is a future problem.

\begin{wrapfigure}{r}{0.35\textwidth}
	\vspace{-10pt}
	\begin{center}
		\begin{subfigure}[b]{0.35\textwidth}
			\includegraphics[width=\textwidth]{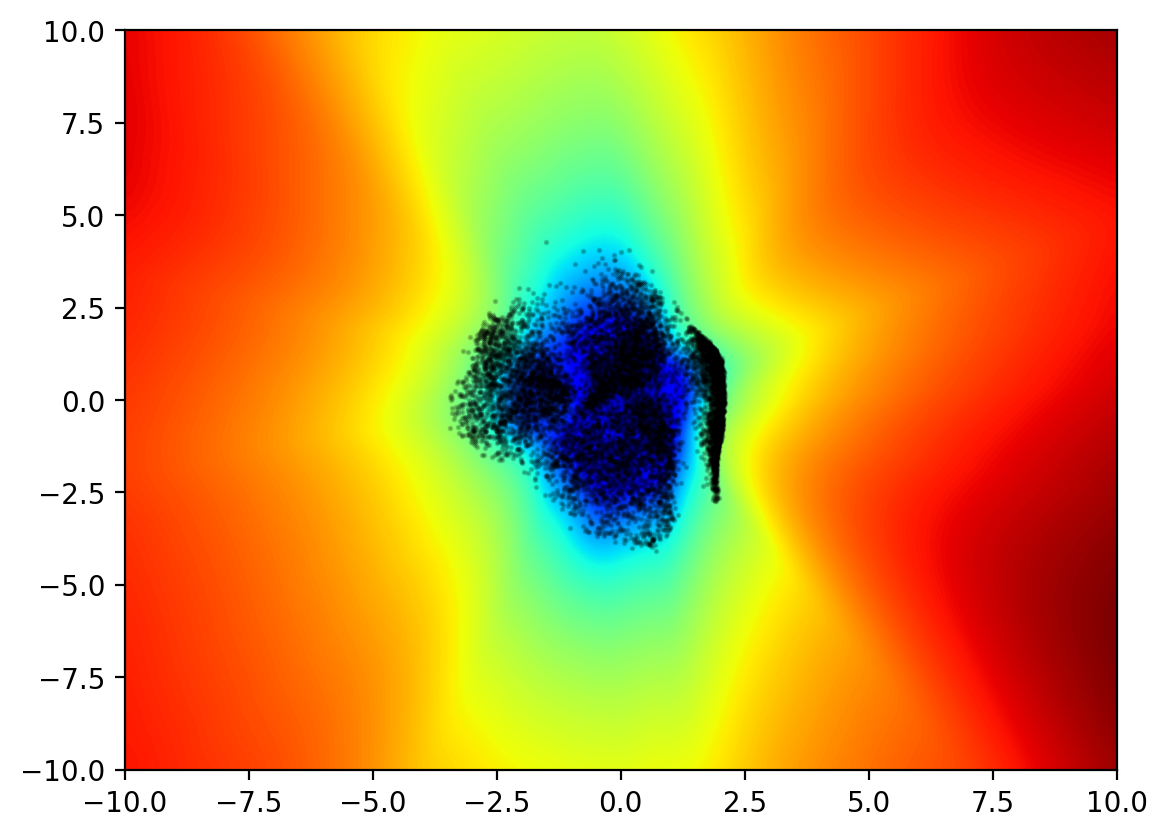}
			%		\caption{AE}
			%		\label{fig:AE}
		\end{subfigure}
		\caption[none]{Meaningful extrapolation. The distance $\norm{\b{b} - \mu(\b{z})}_2$ in $\Z$, where $\b{b}$ is the center of the data.}
		\label{appendix:experiment:vae:distanceFromCenter}
	\end{center}
	\vspace{-10pt}
\end{wrapfigure}As an additional experiment we examine if the proposed meaningful extrapolation technique is useful. Therefore, using a set of points $\b{z}_s$ on a uniform grid in the latent space, we generate the points in $\X$ on the expected manifold $\M_\Z$ as $\b{x}_s = \mu(\b{z}_s) = f(\b{z}_s) +\b{U}\cdot \text{diag}([\sqrt{\lambda_1}, \sqrt{\lambda_2}]) \cdot \b{z}_s + \b{b}$. Here, the $f:\Z\rightarrow\X$ is a DNN and the linear part is defined as explained in the main paper. In Fig.~\ref{appendix:experiment:vae:distanceFromCenter} we show for each $\b{z}_s$ the Euclidean distance measured in $\X$ between the center of the training data and the corresponding point $\b{x}_s$. Indeed, we see that as we move further from the prior $p(\b{z})$ support, the distance between the points on the generated surface and the center of the data increases. However, we observe that the distance on the $x_1$-axis increases faster than the $x_2$-axis. The reason is that the corresponding eigenvalue of the linear map is higher, so the generated $\M_{\Z}$ extrapolates linearly faster along this latent dimension. Note that in this example we used the $\texttt{softplus}(\cdot)$ activation function, for which the extrapolation behavior is more difficult to analyze than the $\texttt{tanh}(\cdot)$. Even so we get a meaningful extrapolation due to the linear part of the function $\mu(\cdot)$.

\begin{figure}[h]
	\centering
	
	\begin{subfigure}[b]{0.32\textwidth}
		{\includegraphics[width=\textwidth]{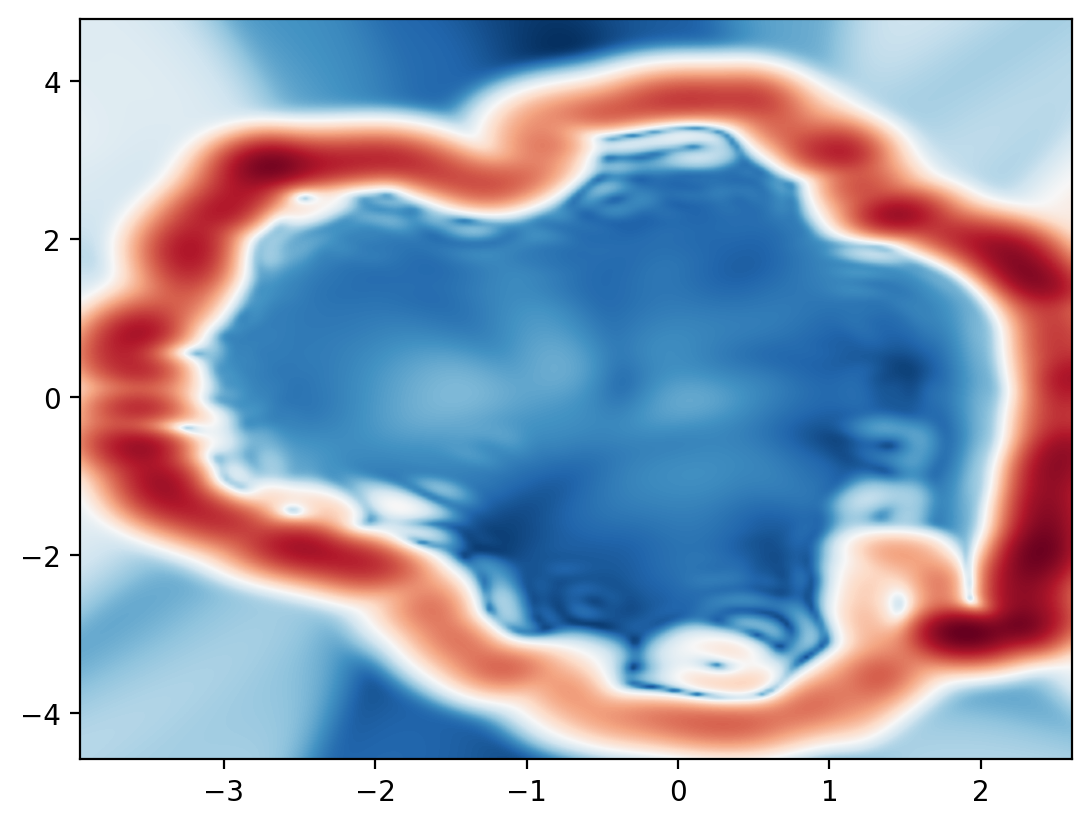}}
		\caption{Euclidean.}
	\end{subfigure}
	~
	\begin{subfigure}[b]{0.32\textwidth}
		{\includegraphics[width=\textwidth]{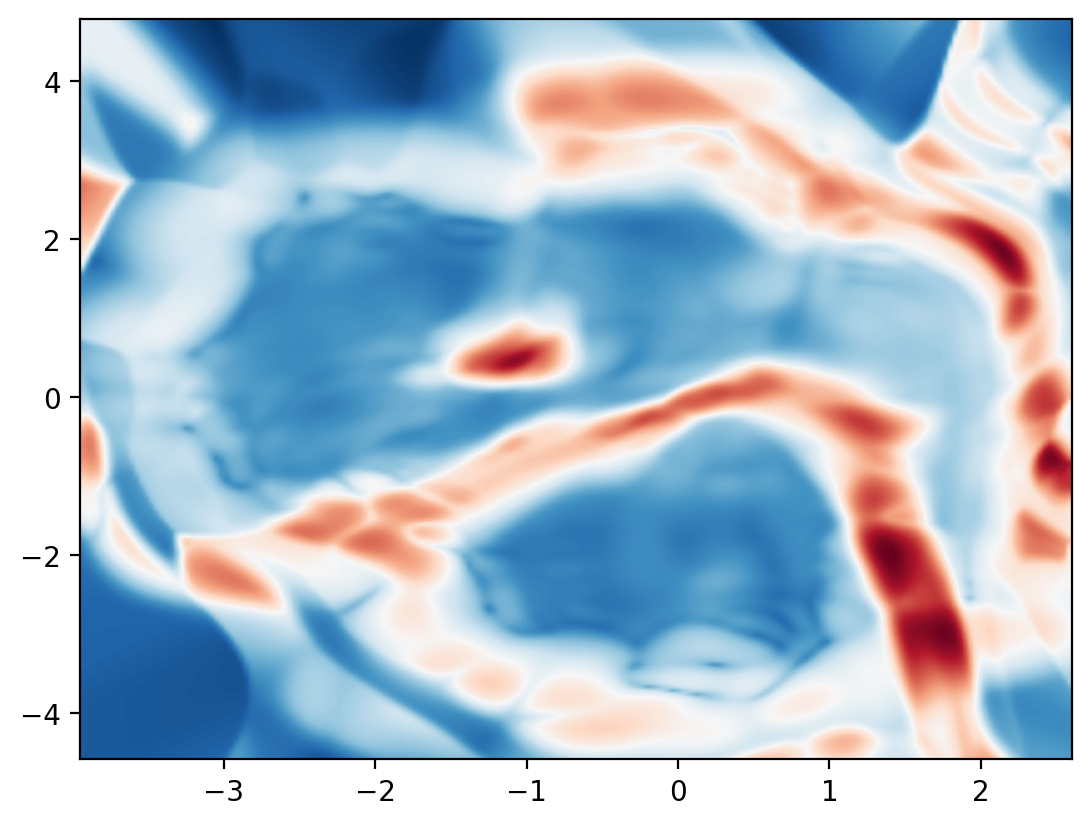}}
		\caption{LDA with 0, 1, 3 vs 2}
		\label{appendix:fig:lda_013_vs_2_measure}
	\end{subfigure}
	~
	\begin{subfigure}[b]{0.32\textwidth}
		{\includegraphics[width=\textwidth]{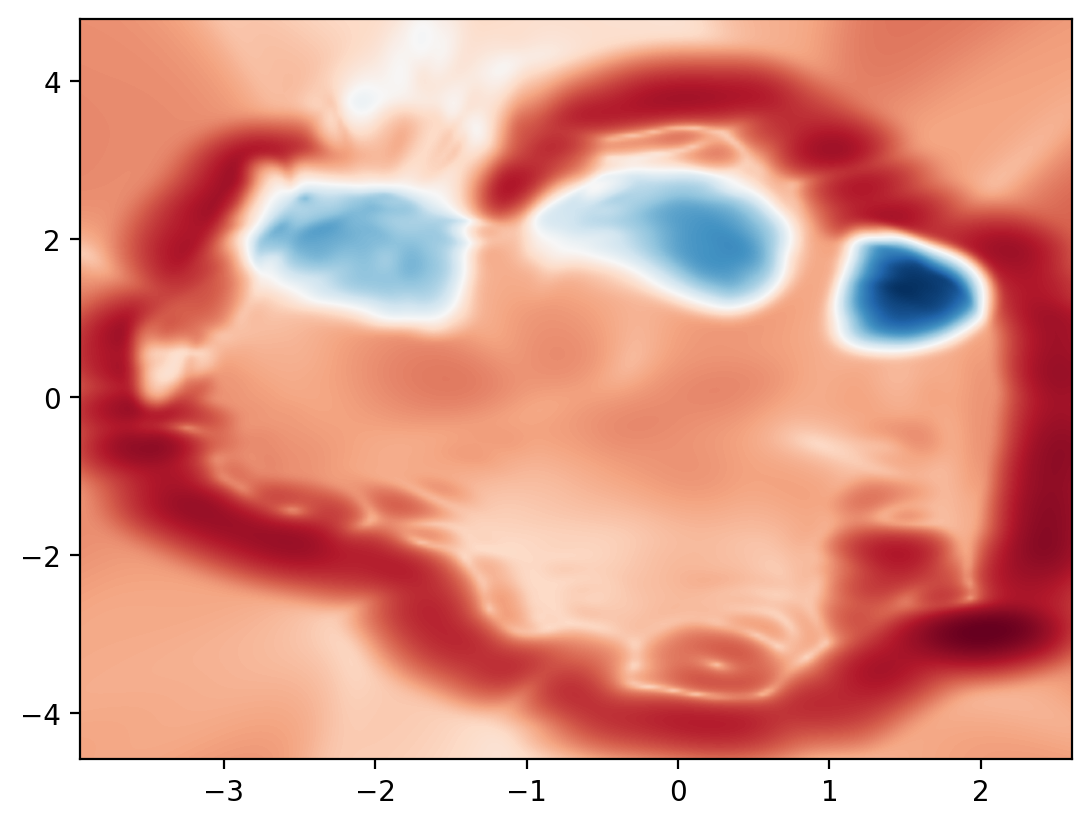}}
		\caption{Local diagonal PCA.}
		\label{appendix:fig:local_diag_pca_measure}
	\end{subfigure}
	
	\begin{subfigure}[b]{0.32\textwidth}
		{\includegraphics[width=\textwidth]{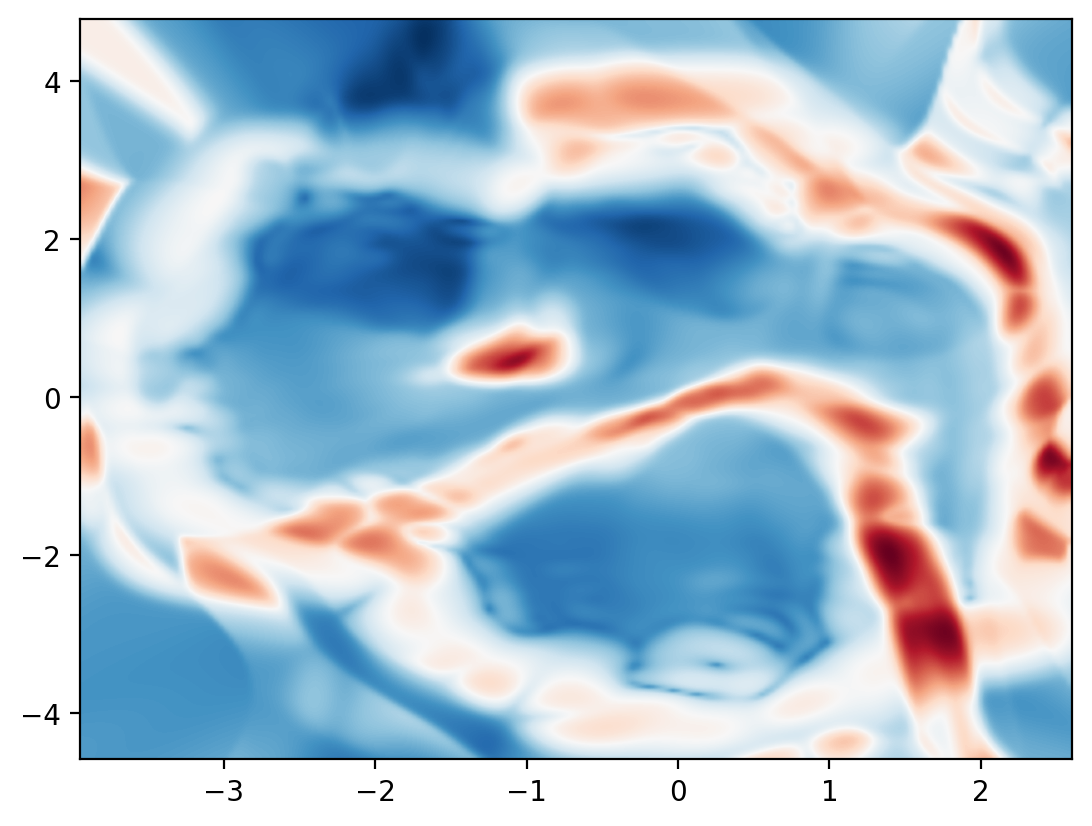}}
		\caption{Combination of \ref{appendix:fig:lda_013_vs_2_measure}, \ref{appendix:fig:local_diag_pca_measure}.}
	\end{subfigure}
	~
	\begin{subfigure}[b]{0.32\textwidth}
		{\includegraphics[width=\textwidth]{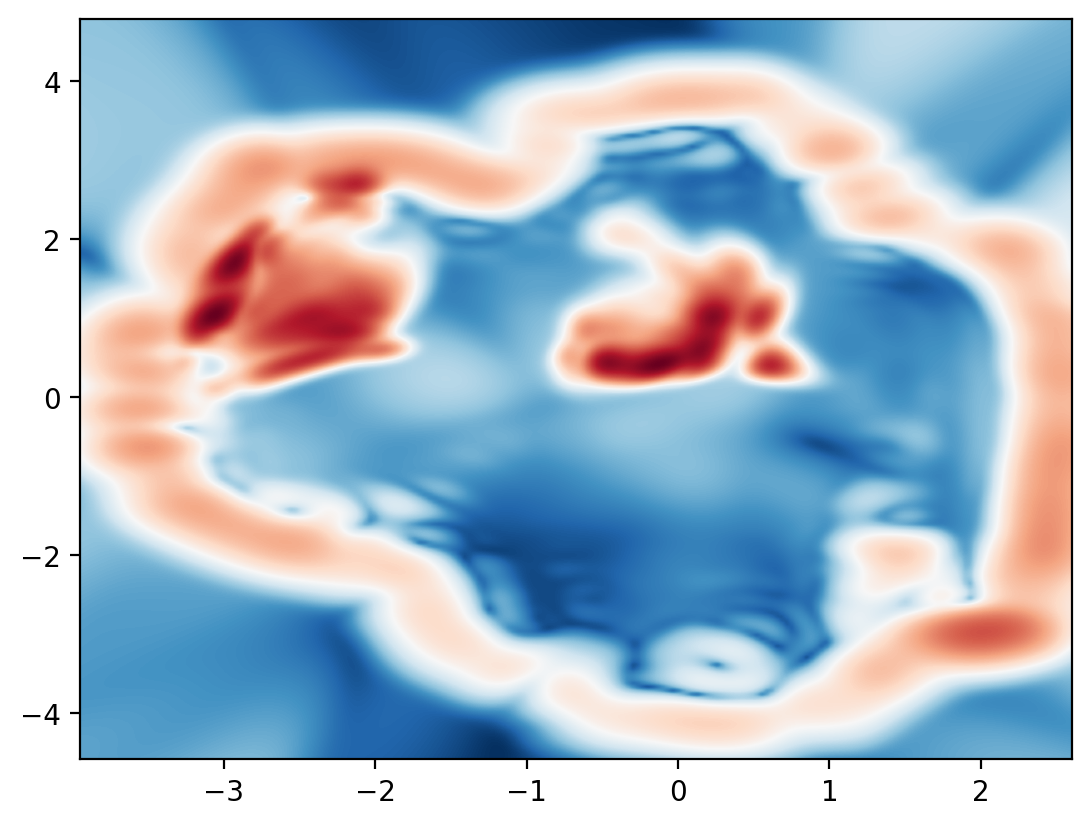}}
		\caption{RBF cost (interpolations).}
		\label{appendix:fig:rbf_cost_inter}
	\end{subfigure}
	~
	\begin{subfigure}[b]{0.32\textwidth}
		{\includegraphics[width=\textwidth]{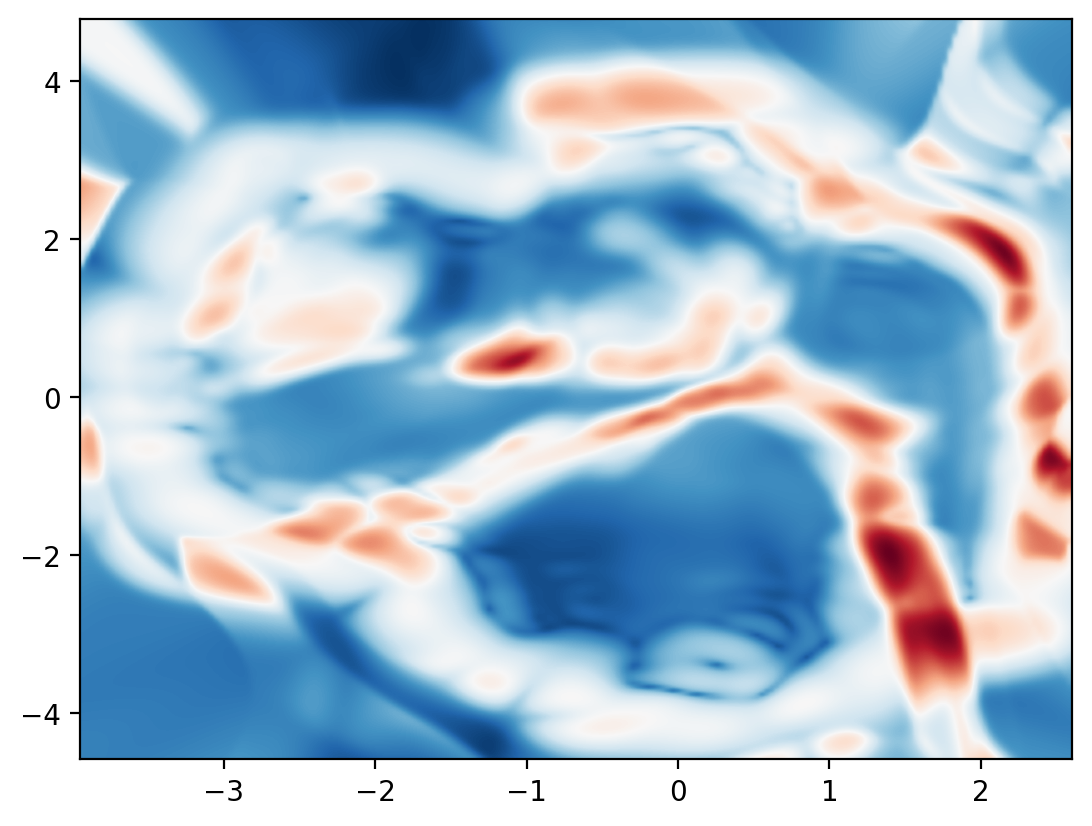}}
		\caption{Combination of \ref{appendix:fig:lda_013_vs_2_measure}, \ref{appendix:fig:local_diag_pca_measure}, \ref{appendix:fig:rbf_cost_inter}.}
	\end{subfigure}	

	\begin{subfigure}[b]{0.32\textwidth}
		{\includegraphics[width=\textwidth]{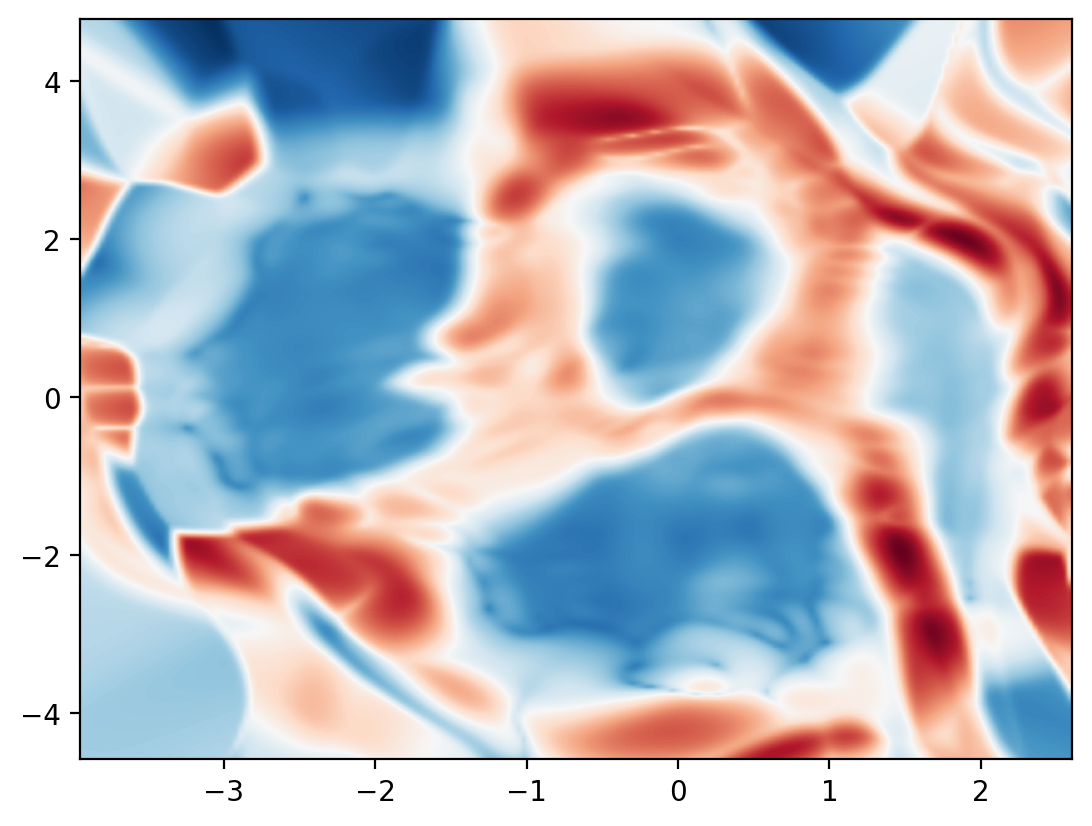}}
		\caption{LDA separate 0,1,2,3.}
		\label{appendix:fig:lda_0123}
	\end{subfigure}
	~
	\begin{subfigure}[b]{0.32\textwidth}
		{\includegraphics[width=\textwidth]{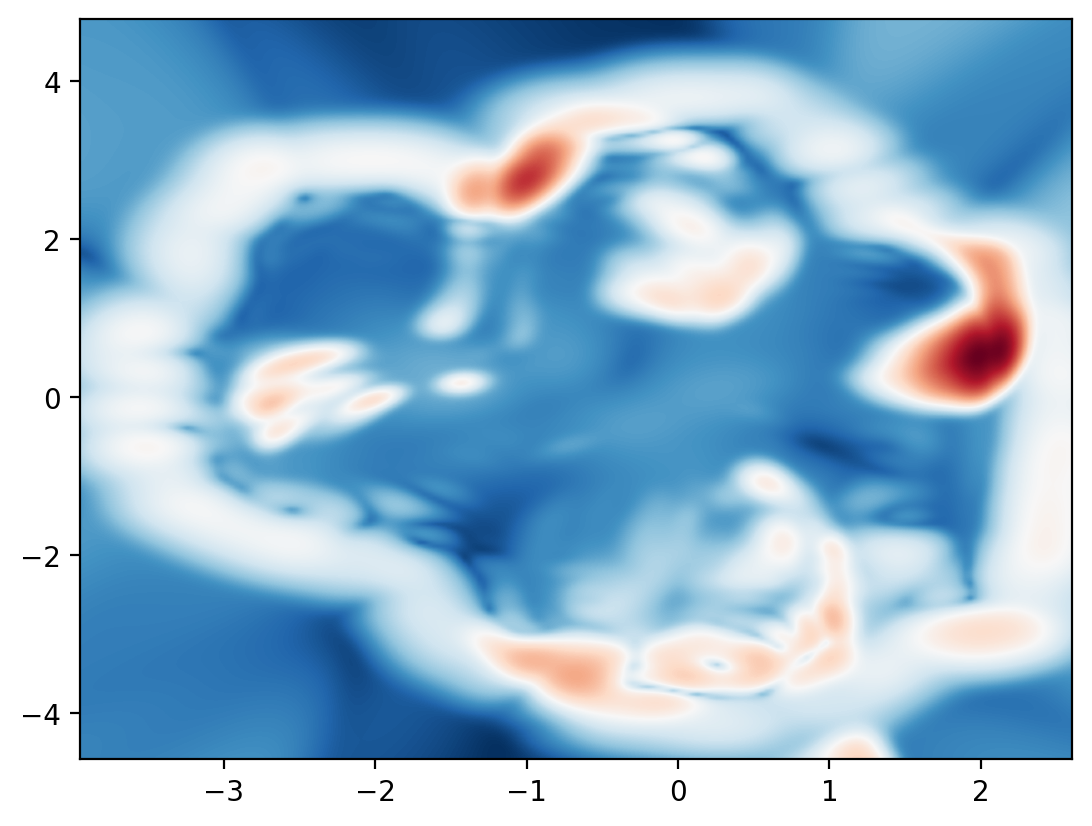}}
		\caption{RBF cost (KDE experiment).}
		\label{appendix:fig:rbf_cost_kde}
	\end{subfigure}
	~
	\begin{subfigure}[b]{0.32\textwidth}
		{\includegraphics[width=\textwidth]{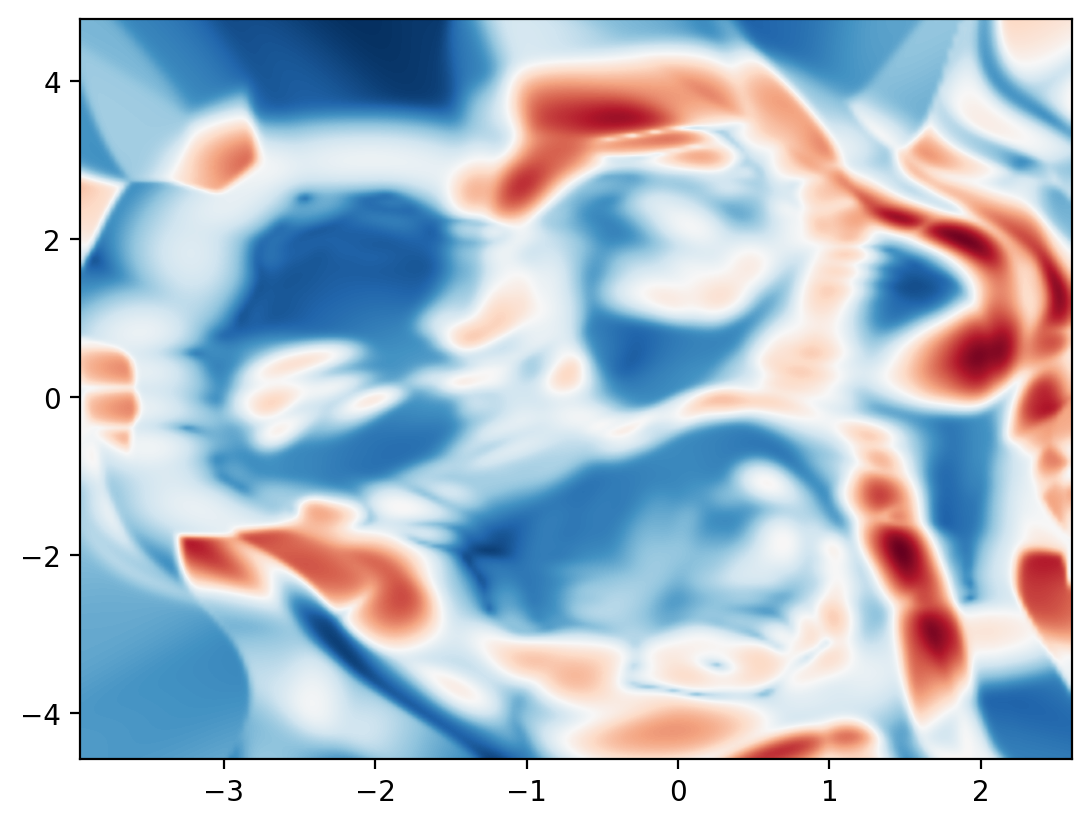}}
		\caption{Combination of \ref{appendix:fig:lda_0123}, \ref{appendix:fig:rbf_cost_kde}.}
	\end{subfigure}
	
	\caption[none]{The Riemannian measure for the VAE experiments and several ambient metrics $\b{M}_\X(\cdot)$. In each caption we mention briefly the form and the details of the ambient metric.}
	\label{appendix:lda_pca}
\end{figure}